\documentclass[conference]{IEEEtran}
\IEEEoverridecommandlockouts
\usepackage{cite}
\usepackage{amsfonts,amsmath,amsthm,amssymb, graphicx, float, nameref, booktabs, enumitem}
\usepackage{algorithmicx}
\usepackage{textcomp}
\usepackage{xcolor}
\usepackage{array}
\usepackage[caption=false,font=normalsize,labelfont=sf,textfont=sf]{subfig}
\usepackage{textcomp}
\usepackage{stfloats}
\usepackage{pifont}
\usepackage{url}
\usepackage{verbatim}
\usepackage{graphicx}
\usepackage{cite}
\usepackage{times}
\usepackage{epsfig}
\usepackage{multirow}
\usepackage{todonotes}
\usepackage{csquotes}

\usepackage{amsfonts,bbm}
\usepackage{siunitx}
\usepackage{subcaption}
\usepackage{algorithm}
\usepackage{algpseudocode}
\usepackage{xspace}
\def\BibTeX{{\rm B\kern-.05em{\sc i\kern-.025em b}\kern-.08em
    T\kern-.1667em\lower.7ex\hbox{E}\kern-.125emX}}

\newcommand*{\model}{\ensuremath{h}\xspace}
\newcommand*{\transformation}{\ensuremath{\mathcal{A}}\xspace}
\newcommand*{\dataset}{\ensuremath{\mathcal{D}}\xspace}
\newcommand*{\datasetin}{\ensuremath{\mathcal{X}}\xspace}
\newcommand*{\datasetout}{\ensuremath{\mathcal{Y}}\xspace}
\newcommand*{\datasetlab}{\ensuremath{\{0,1\}}\xspace}
\newcommand*{\singlein}{\ensuremath{\mathbf{x}_i}\xspace}

\newcommand*{\singleout}{\ensuremath{\mathbf{y}_i}\xspace}

\newcommand*{\singlelab}{\ensuremath{{z}_i}\xspace}
\newcommand*{\perturbation}{\ensuremath{p}\xspace}
\newcommand*{\perturbationSet}{\ensuremath{\mathcal{P}}\xspace}

\newcommand*{\targetFun}{\ensuremath{t}\xspace}
\newcommand*{\method}{\emph{SpaNN}\xspace}

\usepackage{fancyhdr}
\begin{document}
\title{SpaNN: Detecting Multiple Adversarial Patches\\ on CNNs by Spanning Saliency Thresholds
}

\author{
\IEEEauthorblockN{Mauricio Byrd Victorica}
\IEEEauthorblockA{
\textit{KTH Royal Institute of Technology}\\
Stockholm, Sweden \\
mbv@kth.se
}
\and
\IEEEauthorblockN{György Dán}
\IEEEauthorblockA{
\textit{KTH Royal Institute of Technology}\\
Stockholm, Sweden \\
gyuri@kth.se}
\and
\IEEEauthorblockN{Henrik Sandberg}
\IEEEauthorblockA{
\textit{KTH Royal Institute of Technology}\\
Stockholm, Sweden \\
hsan@kth.se}
}

\maketitle
\fancypagestyle{firstpage}
{
    \fancyhead[L]{This work
has been accepted for publication in the IEEE Conference on Secure and
Trustworthy Machine Learning (SaTML). The final version will be
available on IEEE Xplore.}    
}
\thispagestyle{firstpage}
\begin{abstract}
State-of-the-art convolutional neural network models for object detection and image classification are vulnerable to physically realizable adversarial perturbations, such as patch attacks. Existing defenses have focused, implicitly or explicitly, on single-patch attacks, leaving their sensitivity to the number of patches as an open question or rendering them computationally infeasible or inefficient against attacks consisting of multiple patches in the worst cases.
In this work, we propose SpaNN, an attack detector
whose computational complexity is independent of the expected number of adversarial patches. The key novelty of the proposed detector is that it builds an ensemble of binarized feature maps by applying a set of saliency thresholds to the neural activations of the first convolutional layer of the victim model. It then performs clustering on the ensemble and uses the cluster features as the input to a classifier for attack detection.
Contrary to existing detectors, SpaNN does not rely on a fixed saliency threshold for identifying adversarial regions, which makes it robust against white box adversarial attacks. 
We evaluate SpaNN on four widely used data sets for object detection and classification, and our results show that SpaNN outperforms state-of-the-art defenses by up to 11 and 27 percentage points in the case of object detection and the case of image classification, respectively. Our code is available at~\url{https://github.com/gerkbyrd/SpaNN} .
\end{abstract}


\section{Introduction}
\label{sec:intro}
Deep learning models achieve state-of-the-art performance on computer vision tasks,  but they are vulnerable to adversarial attacks, i.e., input perturbations crafted to change the model's output~\cite{ExpHarAdv, AdvPatch, advTshirt}. One class of adversarial attacks is digital attacks, which involve imperceptible perturbations of the input image, often bounded in some $\ell_p$ norm. Several digital attack generation methods have been proposed in the past decade~\cite{ExpHarAdv, CW, deepfool, bypassing-det}, followed by corresponding defense schemes~\cite{det-art, aex-bugs, rand-smooth, LNG}.  These attacks assume the adversary has direct access to the pixels of the input image provided to the model.

In more recent years, focus has shifted towards physically realizable attacks~\cite{synthesizing-rob}. They differ from digital attacks in that they are spatially constrained, and they typically involve applying a printable patch containing an adversarial pattern to an object in the physical scene. For instance, an adversarial patch can be applied in the form of a sticker~\cite{AdvPatch,tnt}, a printed pattern on clothing~\cite{advTshirt,tnt}, or a projected image~\cite{phantom}. Unlike digital attacks, patch attacks do not assume access to the digital images in the deep learning model's processing pipeline, and instead manipulate physical objects in the scene, which makes their implementation more feasible and eliminates the need to access the victim model's input directly.

Existing defenses against patch attacks either aim at detecting adversarial patches~\cite{patchguard++, z-mask, MRD,scale-cert, uap-realtime, vip,lancex, napguard} or at recovering from patch attacks by localizing the patches and removing them~\cite{themis, jedi,ObjSeek, lightweightSSdet,lancex,adv-pf-energy, patchzero, wcacy, PAD, FNS}. The approach they follow for  detecting patches is based on the patches' impact on statistical properties of the input data, e.g., by computing gradients in the pixel domain~\cite{patchzero},  by detecting unusually high activations in feature maps~\cite{themis, z-mask}, or by detecting high entropy regions in pixel space~\cite{jedi}.    
As a result, existing approaches for detecting patch attacks against convolutional neural networks (CNNs) suffer from two main limitations. First, most methods, explicitly or implicitly, assume a single patch per object~\cite{napguard, jedi, PAD}, or even a single patch per image~\cite{themis, ObjSeek}, making them vulnerable to attacks deviating from such assumptions. Second, they are based on one or more detection thresholds that are compared to image statistics in one or more feature spaces (e.g., thresholds on image entropy~\cite{jedi} or internal layers' neural activations~\cite{themis, scale-cert, z-mask}), and hence they require parameter tuning and adjustments to changes in the defended model or the input data distribution.

In this paper, we propose \method, a novel patch attack detection method that overcomes limitations of existing defenses. \method achieves superior detection performance owing to two key ideas. First,  detection in \method is based on how the spatial patterns of important neurons in a shallow feature map \emph{change} as the  definition of \emph{important neurons} changes. Second, the pattern changes used to distinguish attacked images from clean images are independent of the number of patches in the image.
These two design choices make it possible for \method to detect attacks regardless of the number of patches, while making it robust to adaptive attacks that maximize impact subject to remaining undetected. Our main contributions are as follows:

\vspace{2mm}
\noindent
i) We propose \method, an approach for detecting multiple adversarial patches based on the clustering analysis of an ensemble of  binarized saliency maps in feature space.\\
ii) We evaluate the proposed detection method on various object detection and image classification tasks and show that \method achieves an effective attack detection accuracy of at least 86.13\% for object detection and 96.64\% for image classification, for any number of patches.\\
iii) We compare \method to various baselines and show that it achieves state-of-the-art performance on single-patch detection, and establishes the new state-of-the-art for multiple-patch attacks.\\
iv) In further experiments, we show that the computational cost of \method is independent of the number of patches, and evaluate its effectiveness against an adaptive attacker.

\vspace{2mm}
The rest of the paper is organized as follows. We discuss related works in Section~\ref{sec:rel-work}. We introduce the relevant background in Section~\ref{sec:preliminaries}, and present  \method in Section~\ref{sec:method}. We present numerical results in Section~\ref{sec:numres} and we conclude the paper in Section~\ref{sec:conc}.

\section{Related Work}
\label{sec:rel-work}
Mechanisms to detect patch attacks against image classification and object detection CNN models have been proposed in recent works~\cite{themis,jedi,ObjSeek, z-mask, napguard}. In \emph{Themis}~\cite{themis}, a sliding window is applied on the feature map produced by the first convolutional layer of a CNN model, to identify ``patch candidates'', i.e., relatively dense areas in terms of neural activity. Attacks are then detected based on the effect that occluding the patch candidates has on the model's output. \textit{Jedi}~\cite{jedi} computes an entropy heat map for the input image using a threshold that is adjusted for each input image and then uses filtering and post-processing to keep only non-sparse high-entropy areas in the heat map. An autoencoder is used to construct patch masks corresponding to high-entropy areas, and the masks are applied to the input image before feeding it to the CNN model. \emph{Z-Mask}~\cite{z-mask} is a similar method to \emph{Jedi}, where two over-activation heat maps are computed using \emph{Spatial Pooling Refinement} (one focusing on the defended model's shallow layers and the other on its deep layers). The heat maps are processed using two MLPs and simple aggregations, resulting in a mask for over-activated input areas and a scalar measure of over-activation. If the scalar measure is above a given threshold, then the mask is applied to the input image. 

NAPGuard~\cite{napguard} trains a modified YOLOv5 object detector to detect only the \enquote{patch} class; to achieve good performance, the loss function used during training encourages the detector to accurately detect the high-frequency aggressive features of adversarial patches, and a low-pass filter is used at inference time to suppress natural features and facilitate the detection of patches. \emph{PAD}\cite{PAD} analyses images through a sliding window to generate semantic independence and spatial heterogeneity heatmaps. After fusing the two heatmaps, the regions which may contain adversarial patches are determined with respect to a threshold that depends on the statistics across the image; \emph{PAD} then relies on the Segment Anything (SAM)~\cite{SAM} image segmentation model to produce adequate patch masks.

Slightly different from the above approaches, certifiable methods aim at providing formal guarantees for a given attack model~\cite{patchguard++,ObjSeek,vip}. \textit{Object Seeker}~\cite{ObjSeek} is a certifiable recovery method specific to object detection, and relies on a two-step process consisting of (i) patch-agnostic masking, where horizontal and vertical lines are used to split the image into two parts at $k$ interpolations on each axis, and (ii) pruning, where the objects detected in masked images are filtered, merged, and subsequently pruned to obtain a robust final inference. In short, a filtered set of masked bounding boxes containing the ones dissimilar enough from the originals is clustered, and a representative from each cluster is selected. The final output consists of the pruned new boxes and those detected on the original image. 

Closely related to our work is \emph{ViP}~\cite{vip}, which is a certifiable detection and recovery method for patch attacks on Vision Transformer models (ViTs) for image classification, explicitly addressing the double-patch attack scenario. As with other methods, \emph{ViP} relies on applying a set of masks to the input image and analyzing the corresponding set of predictions, assuming that at least one mask occludes the patch attack. An attack is detected if any two predictions are inconsistent, and clean predictions can be recovered by majority voting. To mask double patches, \emph{ViP} uses \emph{generalized windows}, which cover disjoint input regions, guaranteeing the occlusion of any two patches of known size. ViT models are leveraged to implement adequate \emph{base classifiers}, which predict labels using a subset of the tokenized input image. Each mask thus corresponds to the unused input regions of a base classifier. Since it focuses on ViT models instead of CNNs, we do not consider \emph{ViP} as a relevant baseline for our work. 
We note that there is no need to restrict ourselves to attack detection methods, since attack recovery methods also perform detection internally. Moreover, most state-of-the-art adversarial patch defenses focus on recovery, and usually, some form of attack detection is at their core~\cite{jedi, themis, ObjSeek, PAD}. 

\noindent
\textbf{Fixed saliency thresholds.} In~\emph{Themis}, a neural activation threshold $\beta$ determines what neurons should be considered important. Important neurons are used to construct a binarized feature map, and if the number of important neurons in a given area exceeds a second threshold $\theta$, then the area becomes a \emph{patch candidate}~\cite{themis}. In \emph{Jedi}, entropy heat maps are constructed using a dynamic threshold, which is computed based partly on the input image, partly on pre-computed statistics for clean images, and partly on hyper-parameters chosen empirically~\cite{jedi}. Similarly, the over-activation heat maps in \emph{Z-Mask} are constructed using pre-computed statistics for activation values of clean images~\cite{z-mask}. Even \emph{Object Seeker}, which aims to be agnostic to the attack model, tunes the victim model's confidence threshold to detect bounding boxes on masked images adequately~\cite{ObjSeek}.

A main shortcoming of these methods is that the optimal threshold values depend on either the data set, the model under attack, or the attack formulation.
\begin{figure*}[!t]
\centering
\subfloat[\footnotesize Number of important neurons.]{\includegraphics[width=0.19\linewidth, height=0.2\linewidth]{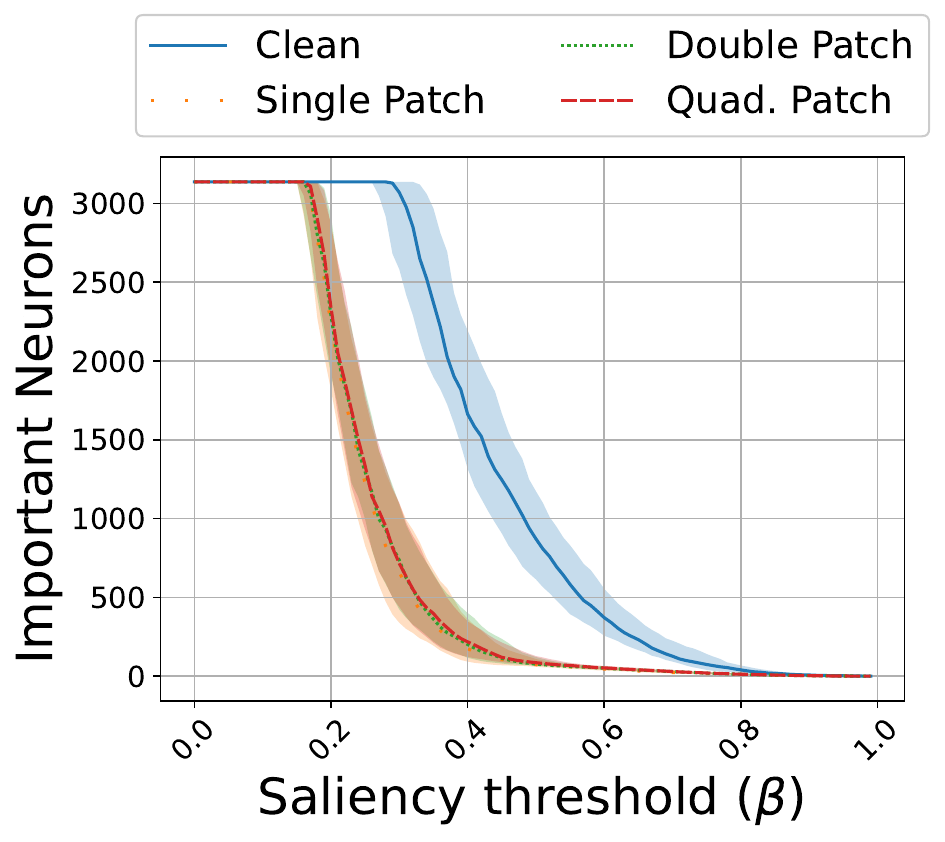}%
}
\hfil
\subfloat[\footnotesize Number of high entropy regions.]{\includegraphics[width=0.19\linewidth, height=0.2\linewidth]{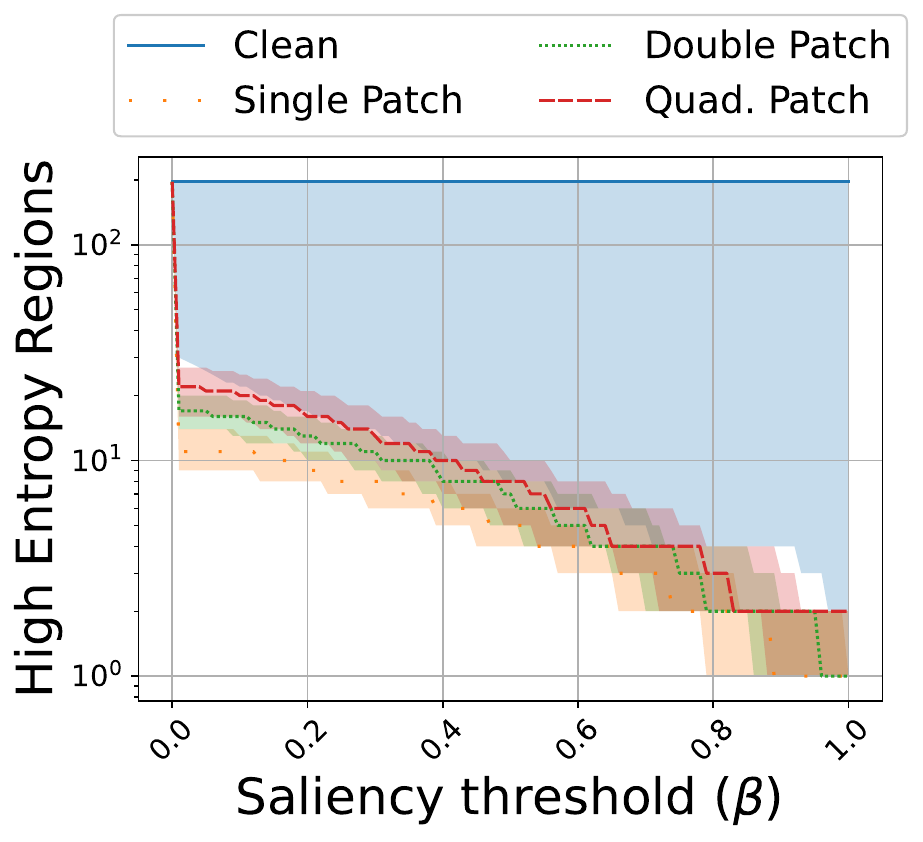}%
}
\hfil
\subfloat[\centering \footnotesize Number of clusters.]{\includegraphics[width=0.19\linewidth, height=0.2\linewidth]{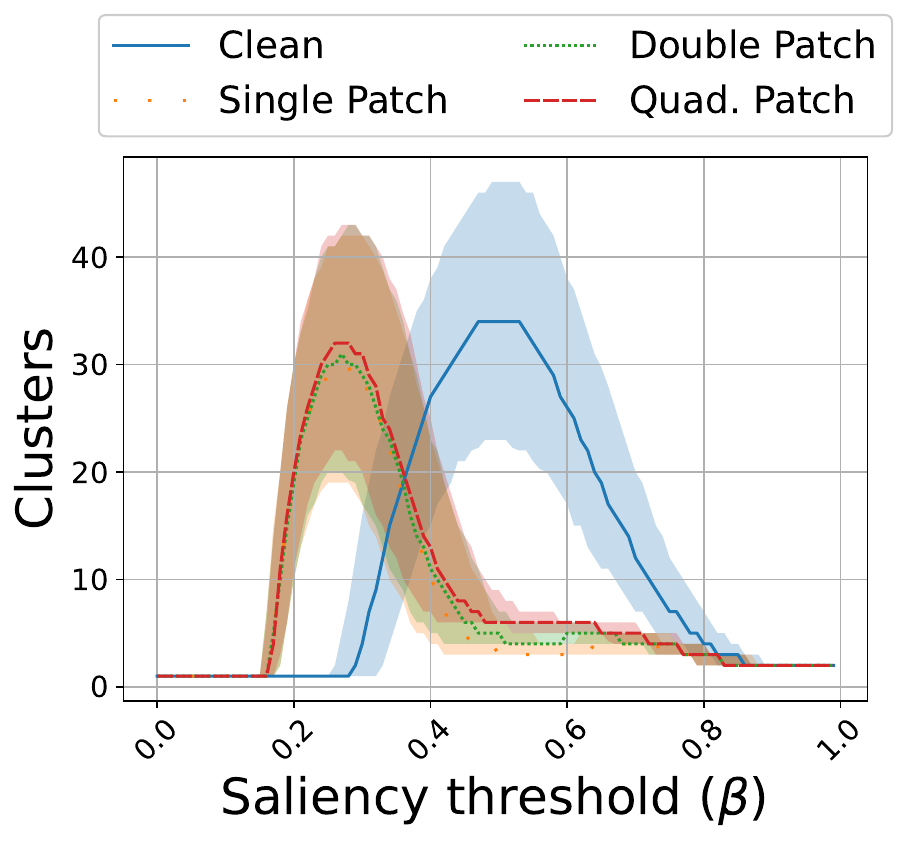}%
}
\hfil
\subfloat[\footnotesize Average mean intra-cluster distance.]{\includegraphics[width=0.19\linewidth, height=0.2\linewidth]{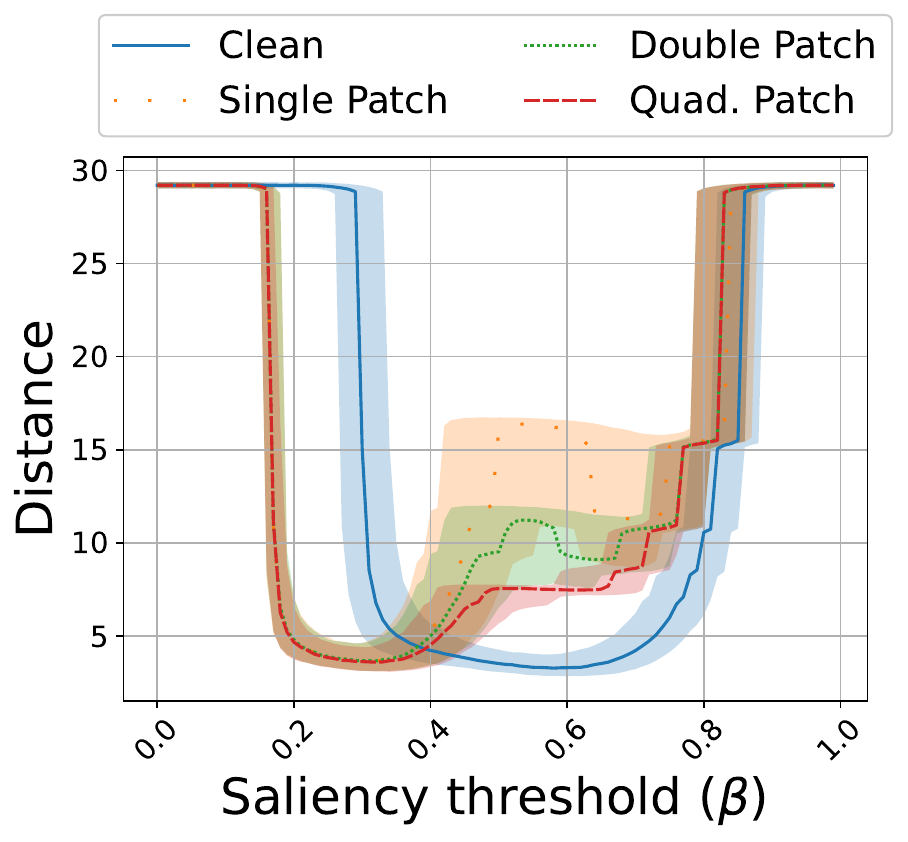}%
}
\hfil
\subfloat[\footnotesize Standard deviation of mean intra-cluster distance.]{\includegraphics[width=0.19\linewidth, height=0.2\linewidth]{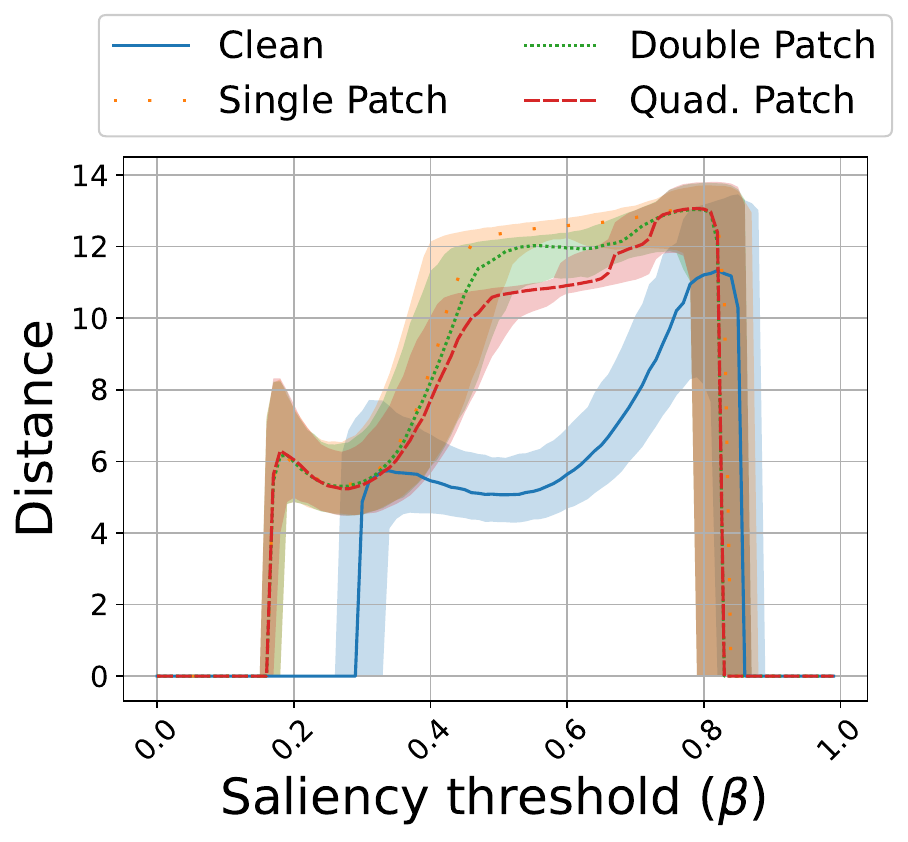}%
}
\caption{Input characteristics vs. saliency threshold $\beta$. Lines represent the median for each quantity, and shaded regions show the first and third quartiles.}
\label{preliminary}
\end{figure*}

\noindent
\textbf{Dealing with multiple patches.} \emph{Themis} operates under the assumption that at most a single patch can be present in any given image~\cite{themis}. \emph{Object Seeker} is also formulated for a single-patch attack, and while a proof of concept for two patches is presented, it requires expensive computations and, unlike the regular method, is not patch-agnostic~\cite{ObjSeek}. While in principle most recent works apply to attacks that place multiple-patches \emph{per object}, their evaluation does not address this scenario~\cite{jedi, z-mask,PAD, napguard}. Focusing on single-patch attacks is a common limitation among defense methods for CNN models and for object detection models in general.

 
\section{Preliminaries}
\label{sec:preliminaries}
In what follows we define the attack model and formulate the attack detection problem. We then motivate our approach by illustrating the relationship between saliency thresholds and input characteristics induced by adversarial patches.
\subsection{Adversarial Patches and Detection Problem}
For a machine learning model $\model$ (e.g., used for image classification, object detection, etc.), and a set $\dataset~=~{\{(\singlein, \singleout): \singlein\in\datasetin, \singleout\in\datasetout\}}$ of input-output pairs, we define the attacker's target output $\targetFun(\singlein)$ as the inference the model should output given input $\singlein$. 
The adversarial attack model is defined by the set of perturbations $\perturbationSet$ that the attacker can choose from and by the transformation function $\transformation$, which is used to apply a perturbation $\perturbation \in \perturbationSet$ to input $\singlein$. Hence, for a model $\model$ the attacker aims to find a perturbation that minimizes the loss function 
\begin{align*}
    \mathcal{L}(\perturbation) =-\mathbb{E}_{\dataset}&[\log \text{Pr}(\model(\transformation(\singlein, \perturbation)) \in \targetFun(\singlein))],
\end{align*}
i.e., it aims to find $ \hat{\perturbation} \in {\arg\min}_{\perturbation \in \perturbationSet}\hspace{2mm}\mathcal{L}(\perturbation)$.
For a targeted attack, the attacker's target output $\targetFun(\singlein)$ is a particular $\singleout'\neq~\singleout$, i.e., $\targetFun(\singlein)~=~\singleout'$. For an untargeted attack, $\targetFun(\singlein)$ is any output different from the clean output $\singleout$, i.e., $\targetFun(\singlein)~\in~\datasetout\backslash\{\singleout\}$.
Adversarial patches are the most explored physically realizable attack model in the literature~\cite{AdvPatch, advTshirt, foolAutoSurveillance, tnt, diffpatch}. In the case of adversarial patches, $\perturbationSet$ defines the number, size, shape, location, and pixel value range of adversarial patches, while $\transformation$ is the replacement operation where the corresponding pixels in the input $\singlein$ are replaced by the patch~$\perturbation$.

Now consider a set $\dataset=\{(\singlein, \singlelab): \singlein\in\datasetin, \singlelab\in\datasetlab\}$ of input-label pairs, where the label indicates whether or not the input has been subject to an adversarial attack. An attack detector $\mathcal{F}_\phi$ parametrized by $\phi\in\Phi$ should thus predict the label for each input $\singlein\in\datasetin$, and the objective is to find detector parameters that minimize the loss function
\begin{align*}
    \mathcal{L}(\phi) =\mathbb{E}_{\dataset}&[\log \text{Pr}(\mathcal{F}_\phi(\singlein) \neq \singlelab)],
\end{align*}
i.e., the goal is to find $\hat{\phi} \in {\arg\min}_{\phi \in \Phi}\hspace{2mm}\mathcal{L}(\phi)$. $\mathcal{F}_\phi$ and $\Phi$ vary widely between attack detection methods proposed in the literature, but most mechanisms proposed to detect patch attacks include a saliency threshold $\beta\in\mathbb{R}$ among their parameters $\phi$~\cite{themis,jedi,PAD}. The saliency threshold is compared to features computed from $\singlein$, e.g., entropy~\cite{jedi} or the neural activations in a hidden layer~\cite{themis}, and as such its choice has a significant impact on $\mathcal{F}_\phi$, as we show next.

\subsection{Input Characteristics Across Thresholds} 
To illustrate the dependence of the attack detection results on the choice of the threshold that is used to detect regions containing patch attacks, we selected a random subset of 3,334 images from the ImageNet validation set and created three attacked versions for each image, using one, two, and four adversarial patches. We also extracted feature maps from all clean and attacked images using the ResNet-50 CNN model~\cite{resnet}.

We then computed the summary statistics on neural activation and entropy, used for detection by \emph{Themis} and \emph{Jedi}, respectively. For neural activation, for each feature map $M$ we extracted, we compute its maximum neural activation $\max(M)$, and for a threshold $\beta\in[0,1]$ we calculate the number of neurons with activation above (or equal to) $\beta\cdot\max(M)$, i.e., the number of important neurons~\cite{themis}. For entropy, we split each image into multiple regions using a sliding window and calculated the entropy $H$ of each region. We then compute the maximum entropy $H_{max}$ among all regions in the image, and for a threshold $\beta\in[0,1]$ we calculate the number of regions with activation above (or equal to) $\beta\cdot H_{max}$, i.e., the number of high entropy regions~\cite{jedi}.

In Figures~\ref{preliminary}(a)-(b) we present summary statistics for the number of important neurons and high entropy regions, across all clean and attacked inputs. The results show that the choice of $\beta$ determines the ability to discriminate between attacked and clean images based on these features, i.e., a detector based on one of these summary statistics must set $\beta$ to a value where the curve for clean images does not overlap with curves for patched images. Once a specific $\beta$ is chosen, an attacker might adapt its attack to generate inputs that are similar to clean inputs for a particular choice of $\beta$. Hence, choosing a single value of $\beta$ makes an attack detector brittle. 

Thus, instead of computing features for a particular  saliency threshold $\beta$, we propose to base detection on how a carefully selected set of features changes as a function of $\beta$.
A simple choice would be to use the previously considered features, as Figures~\ref{preliminary}(a)-(b) show that the shapes of the curves as a function of the saliency threshold $\beta$ are substantially different for attacked and clean images regardless of the number of patches. Nonetheless, features better than these can be constructed by considering the spatial distribution of important neurons in the feature map, inspired by the observation that adversarial patches affect localized areas of the feature map~\cite{themis,patchguard++}. For designing new features, we binarized each feature map $M$ using different values of the importance threshold $\beta$, i.e., for each $M$ and $\beta$ we computed a binary version of $M$, replacing with zeros the elements with values below $\beta\cdot\max(M)$ and replacing all other elements with ones. We performed clustering on the binarized feature maps using DBSCAN, and for each $\beta$ and each $M$, we computed the number of clusters, the mean average intra-cluster distance, and the mean standard deviation of intra-cluster distances. We show summary statistics of these quantities as a function of the threshold $\beta$ in Figures~\ref{preliminary}~{(c)~-~(e)}. We observe that for each quantity, there is a notable difference between the curves corresponding to clean and patched images for any number of patches. This observation is the basis of the detection scheme we propose next. 
Importantly, our approach does not rely on any single saliency threshold to distinguish attacked images from clean ones, which makes it less vulnerable to evasion attacks.

\section{Clustering-based Attack Detection from Binarized Feature Map Ensembles}
\label{sec:method}
%



\begin{figure*}
    \centering
    \includegraphics[width=\textwidth, height=0.3\textwidth]{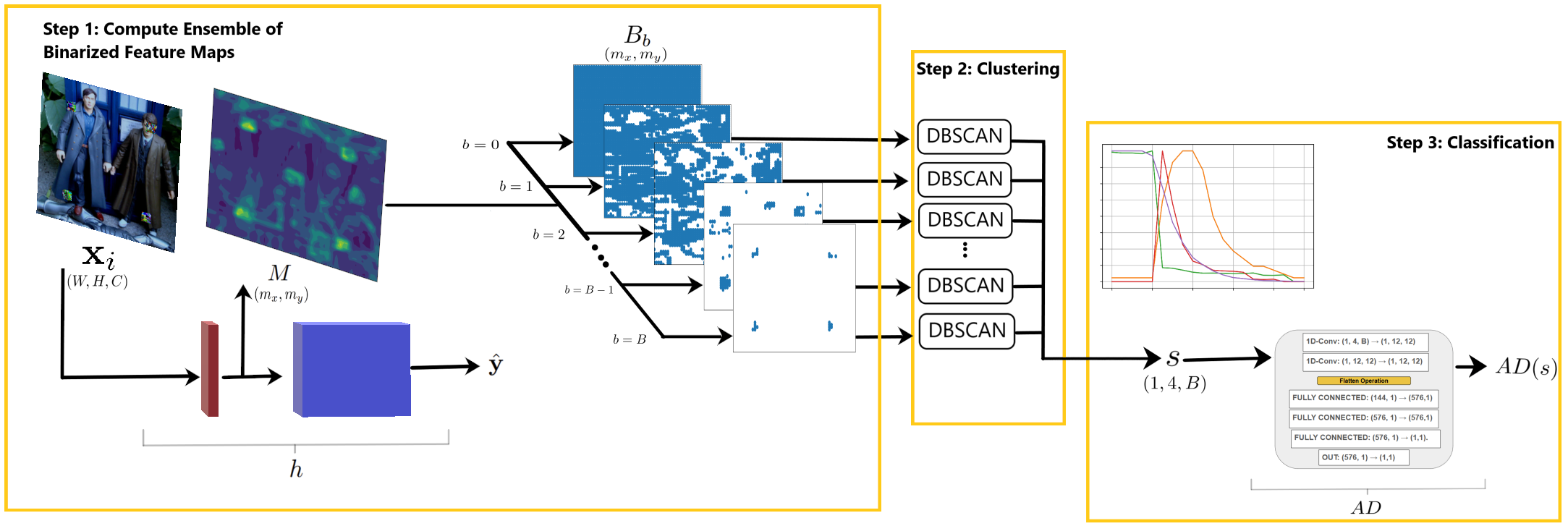}
    \caption{\method: For any input $\singlein$, after extracting a feature map $M$ from a shallow layer of the victim model $h$, a binarized feature map $B_b$ is obtained for each threshold $\beta_b$ in the set $\mathcal{B}$. DBSCAN is applied to each element in the ensemble, and the resulting clustering feature vector $s$ is fed to the neural network $AD$, which outputs an attack detection score~$AD(s)$.}
    \label{diagram}
\end{figure*}

Our proposed approach consists of three steps, performed on input $\singlein$ to a given CNN model $\model$: (i) computing an ensemble of binarized feature maps, (ii) executing a clustering algorithm for each element in the ensemble, and (iii) classifying $\singlein$ as benign or adversarial based on the clustering results across the ensemble. Our method is illustrated in Figure~\ref{diagram}.

\subsection{Computing Ensembles of Binarized Feature Maps} 
Given a CNN model $\model$ and an input $\singlein$, we sum the output of intermediate layer $\ell$ across channels to obtain the feature map $M = \model_{\ell}(\singlein)$; hence $M$ is two-dimensional. 
For a set ${\cal B}=(\beta_1,\ldots,\beta_B)$ of threshold values, the key tenet of the proposed attack detector is to compute a binarized feature map $B_b$ from $M$ for each threshold $\beta_b\in {\cal B}$. 
For threshold $\beta_b$ the binarized feature map $B_b$ has the same dimensions as $M$, and binary entries ${B_b}_{ij}  = \mathbbm{1}_{M_{ij} \geq \beta_b\cdot\max(M)}$. We thus obtain a total of $|\mathcal{B}|$ binarized feature maps. Note that any threshold $\beta_b$ must be between 0 and 1.

\subsection{Clustering of Binarized Feature Maps} 
The second step is to characterize the spatial distribution of nonzero entries in each binarized feature map. We do so by clustering each binarized feature map $B_b$ using DBSCAN~\cite{DBSCAN}. DBSCAN aims to find areas of high density in data space, in terms of the Euclidean distance $\epsilon$ between data points. Clusters are formed according to core samples, data points with at least $w_{\min}$ neighbors at a distance lower than $\epsilon$. Any data points not neighboring or being a core sample are discarded as outliers. 

There are three main reasons for choosing DBSCAN. First, it is a density-based approach, aligning with the notion that adversarial patches result in dense localized areas of important neurons. Second, DBSCAN does not require tuning hyper-parameters such as the number of clusters. Third, it is widely available, easy to implement, and computationally efficient.

\noindent
\subsection{Construction of Clustering Features and Classification.} Given the clustering results, we compute for each threshold $\beta_b$ the number $n_c$ of clusters, the mean average intra-cluster distance $\overline{d_{ic}}$ (i.e., the mean distance between points in a cluster, averaged over all clusters obtained for $B_b$), the standard deviation $\sigma(d_{ic})$ of the average intra-cluster distance, and the number of important neurons $n_{imp}$, i.e., the number of non-zero elements. 
We thus obtain $4B$ quantities, which we use to construct a clustering feature vector $s\in\mathbb{R}^{4\times B}$, where the rows of $s$ are one-dimensional curves of length $B$ corresponding to each clustering metric. We preprocess $s$ by normalizing it over its second dimension, and then re-scaling and centering it around zero so that each row of $s$ has its values between -1 and 1. $s$ is used as input to \emph{AD}, a 4-channel one-dimensional CNN, taking each row of $s$ as a separate channel;  \emph{AD} has an intentionally simple architecture, as it can be trained quickly, is less prone to overfitting, and allows fast inference. The parameters of \emph{AD} are as follows.
\begin{itemize}[align=parleft, nosep, left=0pt]
    \item 1D convolutional layer: 4 input channels, kernel size~=~2, stride~=~1, 12 output channels. Followed by 1D average pooling, 1D batch-norm, and ReLU activation function.
    \item 1D convolutional layer: 12 input channels, kernel size~=~2, stride~=~1, 12 output channels. Followed by 1D average pooling, 1D batch-norm, ReLU activation function, and a flattening operation to pass a single-dimensional input to the next layer.
    \item Fully connected layer: 144-dimensional input, 576 units, followed by ReLU activation function.
    \item Fully connected layer: 576-dimensional input, 576 units, followed by ReLU activation function.
    \item Output layer: 576-dimensional input, 1 unit, followed by a sigmoid activation function.
\end{itemize}

The output of \emph{AD} is the detection score (between 0 and 1), which is used for identifying whether or not there is an adversarial patch in input $\singlein$. The pseudocode of the proposed attack detector is shown in Algorithm~\ref{ditad}. 
\begin{algorithm}[!b]
\small
\caption{\method.}\label{ditad}
\begin{algorithmic}
\Require Model~$\model$, Attack detector~$AD$, set of thresholds~${\cal B}~\in~[0,1]^{B}$, input data~$\mathcal{X}$
\For{$\singlein \in \mathcal{X}$}
\State $M, \hat{\mathbf{y}} \gets \model(\singlein)$ \Comment{$M \in \mathbb{R}^{m_x\times m_y}$ is a feature map}
\State $s := \{\}$ \Comment{Empty sequence to be filled}
\For{$\beta_b \in \mathcal{B}$}
\State $t \gets \beta_b\cdot\max(M)$ \Comment{Importance threshold}
\State$B_b := M\geq t$\Comment{${B_b}_{ij} := \mathbbm{1}({M_{ij} \geq t})$}\vspace{2mm}
\State ${n_{imp}}_b \gets \sum_{i,j} {B_b}_{ij}$\vspace{2mm}
\State ${n_c}_b, {\overline{d_{ic}}}_b, {\sigma(d_{ic})}_b \gets \text{Clustering}(B_b)$\vspace{2mm}
\State $s \gets s\cup\{{n_c}_b, {\overline{d_{ic}}}_b, {\sigma(d_{ic})}_b, {n_{imp}}_b \}$
\EndFor
\State $s \gets \text{Preprocess}(s)$
\State \Return $AD(s)$ \Comment{Attack detection model output}
\EndFor
\end{algorithmic}
\end{algorithm}

\section{Numerical Results}
\label{sec:numres}
We use our clustering-based approach to detect single and multiple patch attacks against commonly used models performing object detection and image classification. We evaluate multiple datasets widely used in the literature on patch attacks, and compare against relevant state-of-the-art baselines.

\begin{figure*}[!t]
\centering

\subfloat[\footnotesize Single patch.]{\includegraphics[width = 0.16\textwidth]{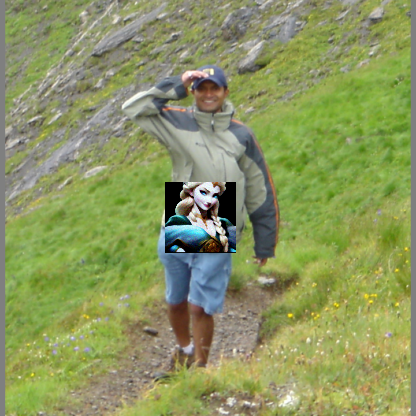}}\hfill
\subfloat[\footnotesize Double patch.]{\includegraphics[width = 0.16\textwidth]{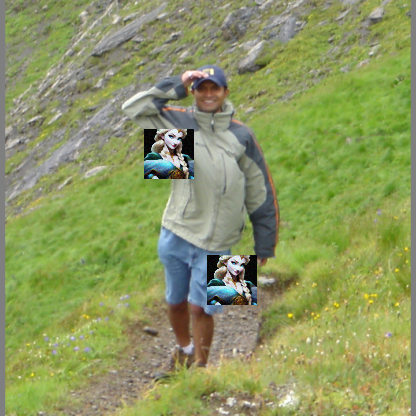}}\hfill
\subfloat[\footnotesize Single patch.]{\includegraphics[width = 0.16\textwidth]{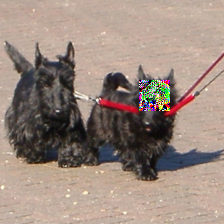}}\hfill
\subfloat[\footnotesize Double patch.]{\includegraphics[width = 0.16\textwidth]{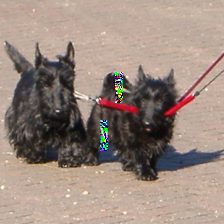}}\hfill
\subfloat[\footnotesize Quad. patch.]{\includegraphics[width = 0.16\textwidth]{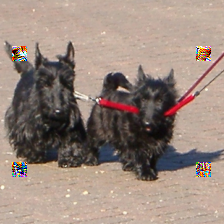}}
\caption{Single and multiple patches for object detection (a-b) and for image classification (c-e).} \label{patch-examples}
\end{figure*}
\subsection{Experimental Setup}\label{sec:setup}
\noindent
\textbf{Models.} We use YOLOv2~\cite{yolo2} to perform object detection for three reasons. First, YOLOv2 is widely available and is computationally efficient. Second, the attack model we considered to train \method was developed for and evaluated on YOLOv2~\cite{foolAutoSurveillance}. Third, the baseline attack detection schemes were initially evaluated on YOLOv2~\cite{themis} or on later versions of YOLO~\cite{ObjSeek, jedi}; YOLOv2 is a representative architecture of later versions, and in general, of state-of-the-art one-stage CNN-based object detectors. For image classification, we use the widely used ResNet-50~\cite{resnet} model, representative of state-of-the-art CNN models for image classification.

\noindent
\textbf{Data.} We use the INRIA Person~\cite{inria} (614 training and 288 test images) and Pascal VOC 2007~\cite{pascal} (4947 training and 4953 test images) datasets for object detection. For image classification, we use the ImageNet~\cite{imagenet} validation set (50,000 images) and the CIFAR-10~\cite{cifar} test set (10,000 images). We focus mainly on INRIA and ImageNet, and report additional results for Pascal VOC and CIFAR-10 in the appendix. 

\noindent
\textbf{Patch Attack Models.} We use state of the art patch attacks against object detection and image classification, as follows.\\
{\it For object detection} adversarial patches are \emph{created} following the attack model presented by Thys et. al~\cite{foolAutoSurveillance} during training and when optimizing defense-aware adaptive patches. A 300~$\times$~300 pixel patch is optimized to minimize the objectness score of the model under attack for a given object (i.e., the patch attack aims to make objects \enquote{disappear}). For the evaluation we use a patch not used during training: the diffusion-based naturalistic \emph{DM-NAP-Princess} patch~\cite{diffpatch}. This patch is readily available and is more challenging to detect than most other patches available in the GAP dataset~\cite{napguard}, which was constructed to evaluate patch attack detection methods. We refer to the appendix for further evaluations on other attacks from the GAP dataset. We \emph{apply} all adversarial patches following Thys et al.~\cite{foolAutoSurveillance}. For a single-patch attack on an object in an image, the square patch is re-scaled to occupy 20\% of the total area of the bounding box the model outputs for the given object, and it is placed in the center of said bounding box. For an attack with two patches on an object, we re-scale each patch to occupy 10\% of the attacked object's bounding box, and place the patches diagonally reflected from each other w.r.t. the center of the bounding box, as illustrated in Figures~\ref{patch-examples}~(a)-(b). The patches differ in location but have the same pixel content, shape, and size. Note that unlike previous works, we consider multiple patches on the same object. 

We say that a patch attack is \emph{effective} 
if at least one of the detected objects in the clean inference $\model(\singlein)$ has an overlap of no more than 50\% with each detected object in the model's inference on the perturbed input $\model(\transformation(\singlein, \perturbation))$. A true positive occurs when a patch attack $\transformation(\singlein, \perturbation)$ is detected by the detector. 
A false alarm (false positive) occurs when an attack is detected for a clean image $\singlein$.

{\it For image classification}, adversarial attacks are created following the open source implementation of the \emph{PatchGuard++} defense~\cite{patchguard++}. For a region in pixel space corresponding to a single patch, with a fixed size of $32\times 32$ pixels and a randomly chosen location, the pixels within the region are optimized to maximize the cross-entropy loss corresponding to the attacked model's prediction of the correct label. For multiple patches, besides the single-patch region, new regions are added symmetrically reflected within the complete image area, as illustrated in Figures~\ref{patch-examples}(c)-(e). Note that in contrast to the patches used for object detection, each patch attack on image classification is optimized for a specific image, i.e., patches in the test set are not used during training.

We say that a patch attack is \emph{effective} if the classifier model inference $\model(\transformation(\singlein, \perturbation))$ is different from the ground truth $\singleout$. A true positive occurs when a perturbed image $\transformation(\singlein, \perturbation)$ is detected as attacked.  A false alarm (false positive) occurs when an attack is detected for a clean image $\singlein$. 
Given a dataset \datasetin of clean and attacked images, we define the attack detection accuracy as the fraction of correct inferences over \datasetin, considering both true positives (\emph{TP}) and true negatives (\emph{TN}). Note that these quantities can be computed over effective attacks,  non-effective attacks, or both; in the sequel, we always state the type of attacks considered. We define the attack detection rate as the fraction of detected attacks, i.e., the recall over \datasetin including both effective and non-effective attacks.

\subsection{Attack Detector Parameters}\label{sec:adparams}
\noindent
\textbf{DBSCAN parameters.} For evaluation, we used DBSCAN parameters $\epsilon=1$ and $w_{\min}=4$ (i.e., points with a Euclidean distance of $\leq 1$ are clustered together, and at least $4$ points must be grouped together to be considered a cluster). The parameter $\epsilon$ is set considering that unimportant and important neurons should not be clustered, and only neurons directly adjacent to each other in the feature map should be clustered. The choice of $w_{\min}$ captures the consideration that less than $4$ adjacent neurons, important or not, are too few to be considered a cluster.

\noindent
\label{nn-arch}
\textbf{\emph{AD} training setup} 
For INRIA and Pascal VOC, $20\%$ of the training set is used for training, and the complete test set is used 
for evaluation. For ImageNet and CIFAR-10, $2\%$ and 2.5\%  of the validation sets are used for training, respectively, and the rest of each validation set is used for evaluation. For any dataset, single-patch attacks are applied on the training samples, yielding a labeled training set of clean and attacked samples; $20\%$ of this training set is randomly selected and set aside as a validation set during training. To train \emph{AD}, we minimize the binary cross-entropy loss using Adam with $\beta_1~=~0.9$, $\beta_2~=~0.999$, batch size of 1, and learning rate of $0.0001$. We implement our model using PyTorch 1.13.1 and use the default parameters and initializations for all layers~\cite{torch}. We shuffle the training set before each epoch, and stop training after 200 epochs without improvement on the validation loss. Unless otherwise noted, the default $\mathcal{B}$ used throughout our evaluation consists of 20 equidistant thresholds starting from (and including) 0, i.e. ${ \cal B}=\{0,0.05,...,0.95\}$. Note that for image classification, we consider only clean images that are correctly classified by the ResNet-50 victim model, and their corresponding attacked versions. 

\subsection{Attack Detection Baseline Methods}
\noindent
\textbf{\emph{Themis}-detect.} In \emph{Themis}~\cite{themis},  the feature map $M$ is binarized using a threshold $\beta$. A sliding window is then applied on the resulting binarized feature map $B$, and patch candidates are the image regions associated with the windows in which the fraction of non-zero entries is above a threshold $\theta$. \emph{Themis}-detect issues an alert whenever it finds at least one patch candidate whose occlusion would modify the output $\model(\singlein)$ (it does not cover patch candidates to recover from patch attacks).

\noindent
\textbf{\emph{ObjectSeeker}-detect.} \emph{Object Seeker}~\cite{ObjSeek} uses $k_x$ horizontal and $k_y$ vertical lines, and it splits the input $\singlein$ into two halves in pixel space using each line, one at a time. It then occludes each of the resulting halves separately and feeds the object detector $\model$ with each of the $2\cdot (k_x+ k_y)$ masked inputs. Given \emph{Object Seeker} considers objects detected in masked inputs as distinct from those in $\model(\singlein)$  when their intersection over area (IoA) is below some threshold $\tau$ for any object in $\model(\singlein)$, \emph{ObjectSeeker}-detect computes the lowest IoA across masked input detections, denoted $\alpha$, and outputs the attack detection score $1 - \alpha$. Note that \emph{ObjectSeeker}-detect cannot be used to detect attacks on image classification.

\noindent
\textbf{\emph{Jedi}-detect.} \emph{Jedi} computes the entropy over a sliding window in pixel space to obtain a heat map. Entries of the heat map that exceed the entropy threshold, determined based on the current input $\singlein$ and on pre-computed statistics for clean images, are retained. It then removes scattered clusters from the truncated heat-map, and feeds the truncated heat-map in an autoencoder trained to reconstruct patch masks. The reconstructed heat map is applied as a mask to the original input $\singlein$, and the masked input is fed into $\model$. In \emph{Jedi}-detect, an attack is detected if the output for the masked input differs from $\model(\singlein)$. 

\noindent
\textbf{\emph{NAPGuard}.} \emph{NAPGuard} uses a one-class object detector based on the YOLOv5 model to detect adversarial patches. The model is trained using an aggressive feature aligned loss (AFAL) and images are pre-processed to remove natural features (those below a frequency threshold) in order to facilitate the detection of adversarial patches. We use the maximum objectness score among patches detected by NAPGuard in a given image as the detection score for that image.

\subsection{Results}
\label{subsec:results}
\begin{figure*}[!t]
\centering
\includegraphics[width = 0.4\textwidth]{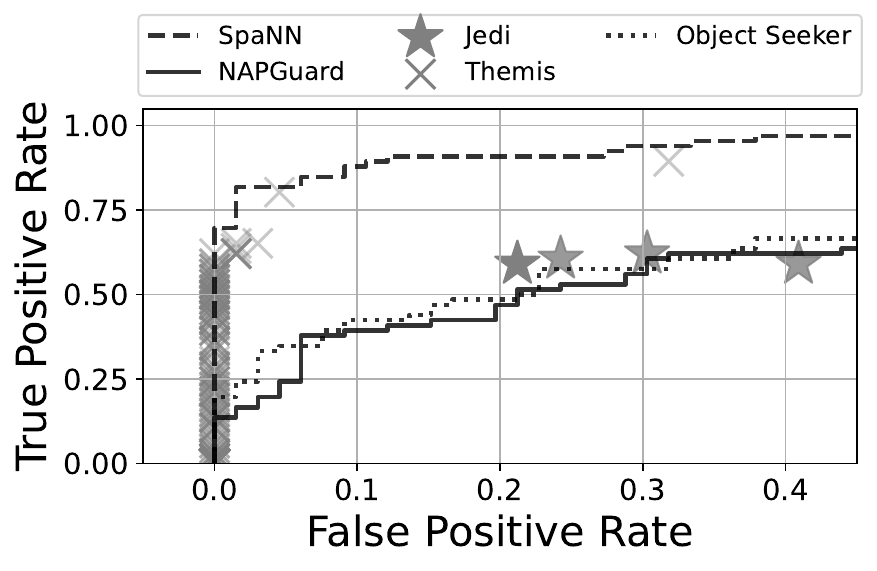}\hfil
\includegraphics[width=0.4\textwidth]{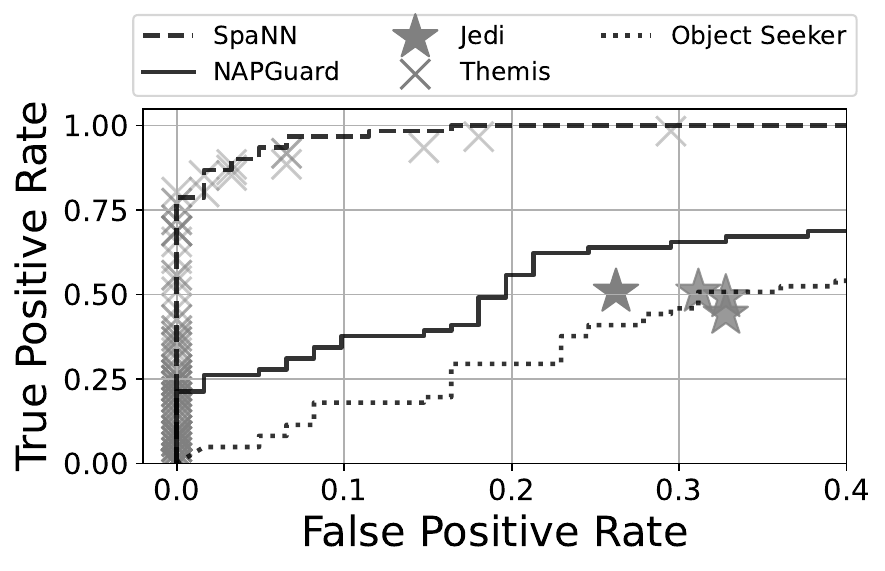}
\caption{Attack detection vs. false alarm rate for single (left) and double (right) adversarial patches for object detection (INRIA).}\label{roc-inria}
\end{figure*}
We start the evaluation by considering the receiver operating characteristic (ROC) curves, i.e., the true positive rate vs the false positive rate, obtained by varying the detection threshold. Since \method, \emph{NAPGuard}, and \emph{ObjectSeeker}-detect output a detection score, their ROC curves can be obtained easily. To obtain ROC curves for the other baselines, we use different values of their internal (by design fixed) detection thresholds. For \emph{Themis}-detect, we vary $\beta$ and $\theta$ from $0.05$ to $0.95$ in $0.05$ increments, and display results only for Pareto optimal configurations. For \emph{Jedi}-detect, we vary the threshold on the auto-encoder output, which determines the final mask applied to input images, changing it to $0.5$, $0.25$, and $0.125$ of its original value. Note that we use the default settings for natural feature suppression in \emph{NAPGuard} and the default number of splitting lines $k_x=k_y=30$ for \emph{ObjectSeeker}-detect.

\begin{figure*}[!t]
\centering
\includegraphics[width=0.32\textwidth]{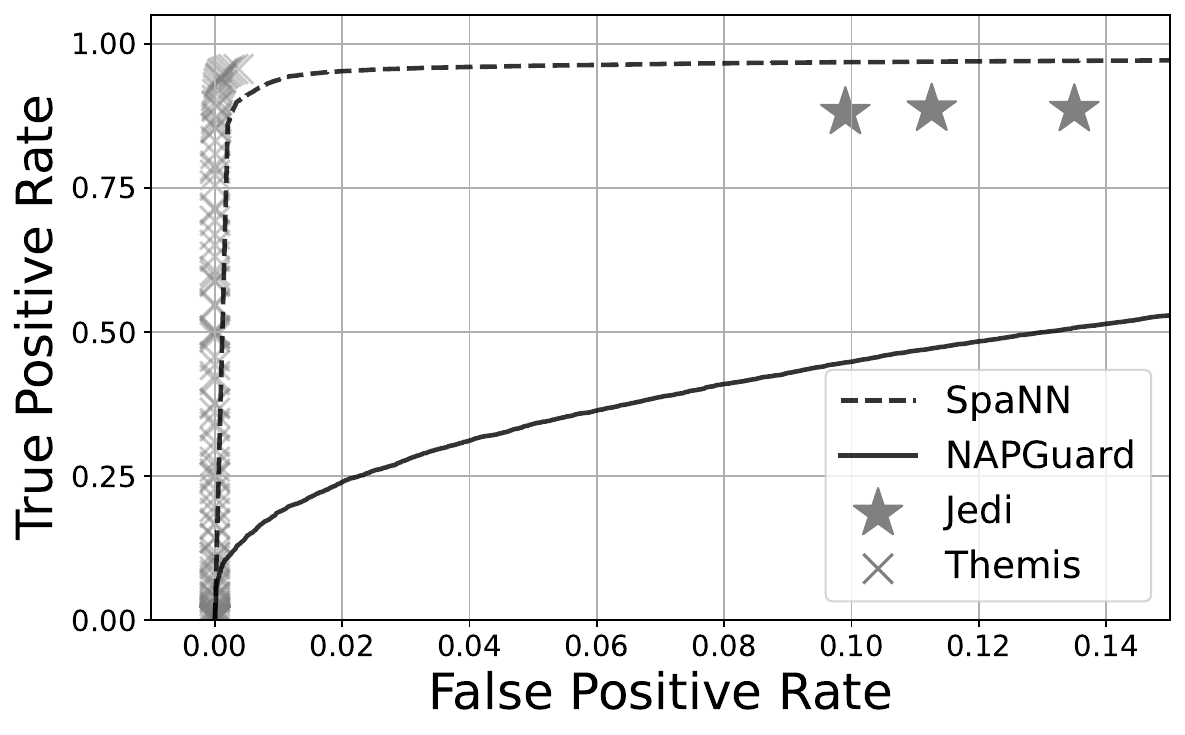}
\includegraphics[width=0.32\textwidth]{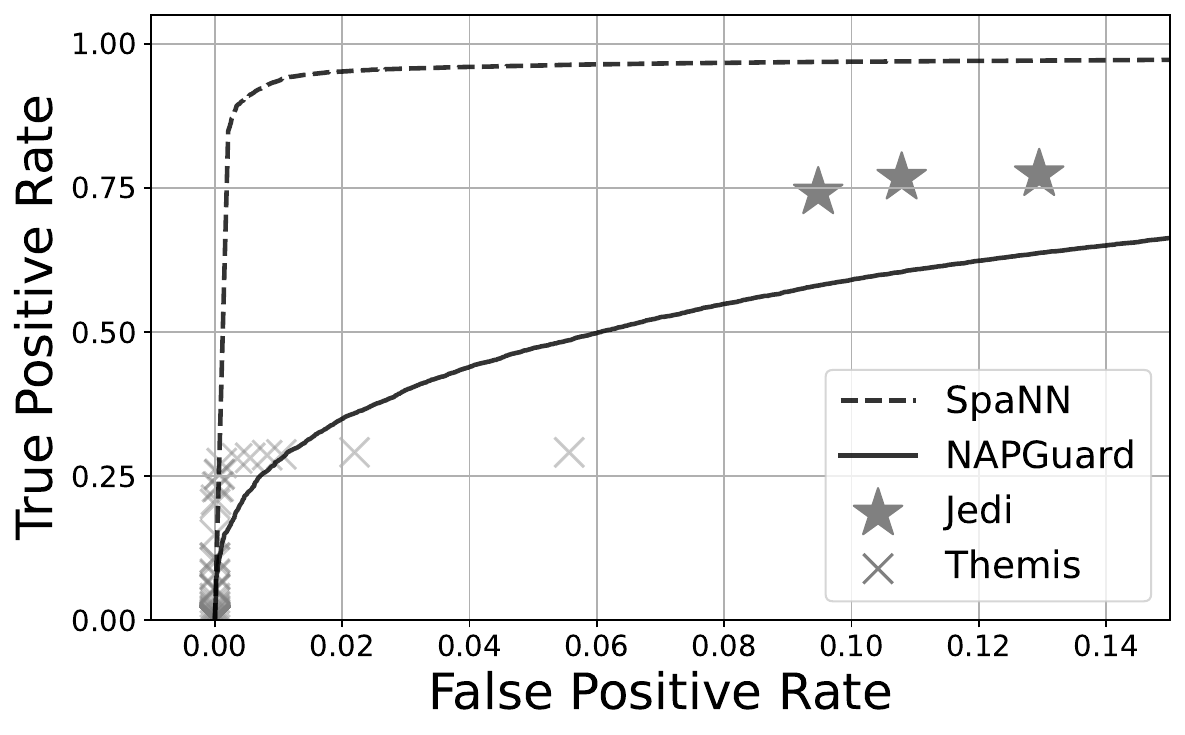}
\includegraphics[width=0.32\textwidth]{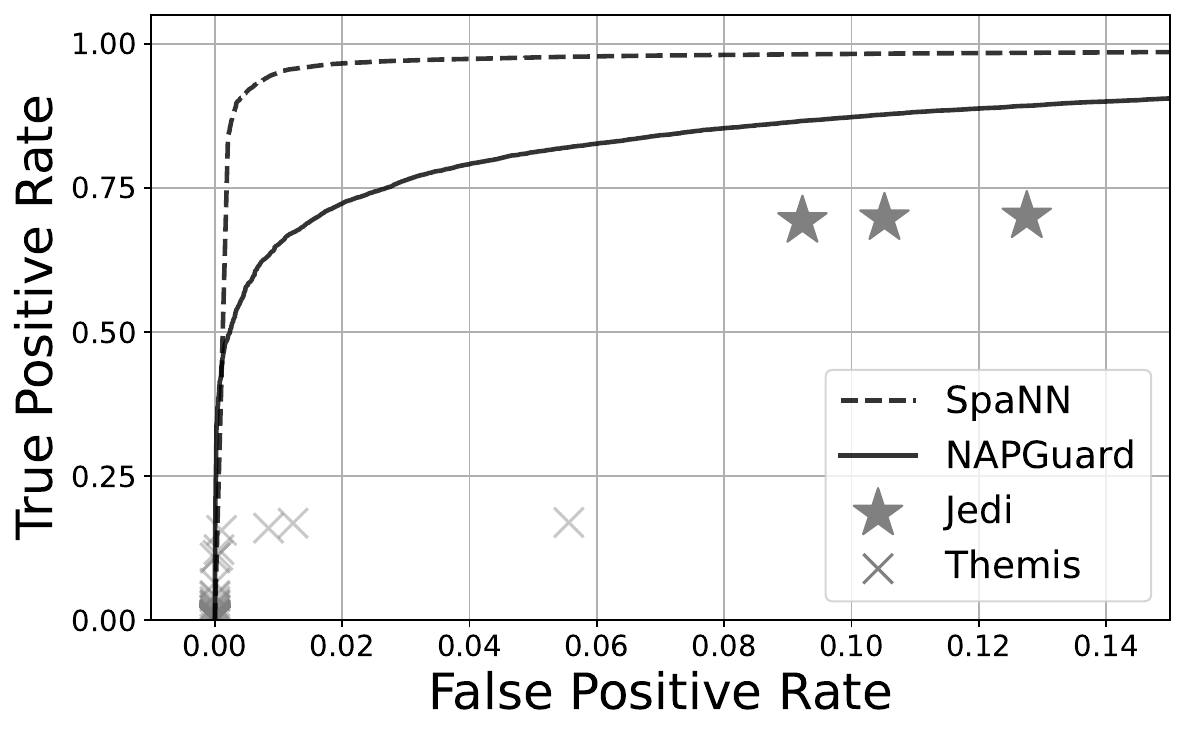}
\caption{Attack detection vs. false alarm rates for single (left), double (middle), and quadruple (right) adversarial patches for image classification (ImageNet).}\label{roc-imgnet}
\end{figure*}

\noindent
\textbf{Object Detection.} We first evaluate the attack detection performance on the INRIA test set defined in Section~\ref{nn-arch}. We consider effective attacks because some baselines are ill-equipped to detect ineffective attacks. Figure~\ref{roc-inria} shows the ROC curves obtained for \method, \emph{NAPGuard}, \emph{Jedi}-detect, \emph{Themis}-detect, and \emph{ObjectSeeker}-detect for single- and double-patch attacks. The figure shows that \method significantly outperforms all baselines regarding the true positive rate for both single and double-patch attacks, at the cost of a very low false alarm rate. The figure also shows that, aside from \emph{Themis}-detect for the double-patch case, the highest true positive rate achievable by the baseline methods is far below that of \method. Moreover, note that \emph{Jedi}-detect and \emph{ObjectSeeker}-detect experience a decrease in performance in the double-patch case. The corresponding results for Pascal VOC are available in Figure~\ref{roc-voc} in the appendix, where the superiority of \method over the baselines in terms of detected attacks and false alarms becomes emphasized. We report the attack detection accuracy achieved by each detector using their best-performing setting in the first eight rows of Table~\ref{detperf-tab}. The table confirms that \method significantly outperforms the baselines.

\begin{table*}
\caption{Attack detection accuracy on object detection (INRIA, Pascal VOC) and image classification (ImageNet, CIFAR-10).}\label{detperf-tab}
\centering
\footnotesize
\begin{tabular}{lcccccc} 
\toprule
   {\textbf{Attack}}& \textbf{\method} & \textbf{NAPGuard} & \textbf{Jedi} & \textbf{Themis} & \textbf{Object Seeker} \\
\midrule
    {Single-patch (INRIA, effective)} & \textbf{0.9015} & 0.6591 & 0.6894 & 0.8788 & 0.6742 \\
    {Single-patch (INRIA, non-effective)} & \textbf{0.9212} & 0.6081 & 0.5090 & 0.8176 & 0.6081 \\
    {Double-patch (INRIA, effective)} & \textbf{0.9508} & 0.7049 & 0.6230  & 0.9262 & 0.5984 \\
    {Double-patch (INRIA, non-effective)} & \textbf{0.9581} & 0.6916 & 0.5661  & 0.9053 & 0.5595 \\
    
    {Single-patch (VOC, effective)} & \textbf{0.8613}& 0.5668& 0.6515 & 0.8045 & 0.5761 \\
    {Single-patch (VOC, non-effective)} & \textbf{0.8478} & 0.5750&  0.5312 & 0.7497 & 0.5411  \\
    {Double-patch (VOC, effective)} & \textbf{0.9115} & 0.6507&  0.6226 & 0.8352 & 0.5329 \\
    {Double-patch (VOC, non-effective)} & \textbf{0.8900} & 0.6424&  0.5288 & 0.7787 & 0.5000\\
\midrule
    {Single-patch (ImageNet, effective)} & {0.9666} & 0.6982 & 0.8907 & \textbf{0.9766} &\textbf{-}\\
    {Single-patch (ImageNet, non-effective)} & \textbf{0.9635} & 0.6900 & 0.4981 & 0.5920 &\textbf{-}\\
    {Double-patch (ImageNet, effective)} & \textbf{0.9664} & 0.7584 & 0.8298 & 0.6390 &\textbf{-}\\
    {Double-patch (ImageNet, non-effective)} & \textbf{0.9664} & 0.7539 & 0.4995 & 0.5845 &\textbf{-}\\
    {Quadruple-patch (ImageNet, effective)} & \textbf{0.9733} & 0.8870 & 0.8001 & 0.5779 &\textbf{-}\\
    {Quadruple-patch (ImageNet, non-effective)} & \textbf{0.9765} & 0.8863 & 0.4885 & 0.6218 &\textbf{-}\\
    
    {Single-patch (CIFAR-10, effective)} & \textbf{0.9876} &  0.7593& 0.9020 & 0.9501 &\textbf{-}\\
    {Single-patch (CIFAR-10, non-effective)} & \textbf{0.9851} &  0.7563& 0.5030 & 0.5871 &\textbf{-}\\
    {Double-patch (CIFAR-10, effective)} & \textbf{0.9884} &  0.8281& 0.8100 & 0.7959 &\textbf{-}\\
    {Double-patch (CIFAR-10, non-effective)} & \textbf{0.9891} &  0.8282& 0.5116 & 0.5935 &\textbf{-}\\
    {Quadruple-patch (CIFAR-10, effective)} & \textbf{0.9976} &  0.9343& 0.6195 & 0.6916 &\textbf{-}\\
    {Quadruple-patch (CIFAR-10, non-effective)} & \textbf{0.9975} &  0.9325& 0.6527 & 0.6060 &\textbf{-}\\
   \bottomrule
\end{tabular}
\end{table*}

A significant advantage of \method compared to the recovery-based baselines is that it does not rely on the victim model's final output to detect attacks. This allows \method to detect patch attack attempts that fail to change the model's output, i.e., ineffective ones, and hence it becomes possible to detect an attack already before it becomes successful, providing improved situational awareness. Table~\ref{detperf-tab} shows that, for non-effective attacks, the accuracy of \method is close to that for effective attacks or even higher, and is significantly higher than that of the baselines, which either have significantly lower accuracy on non-effective attacks than on effective attacks, or perform poorly on both.

\noindent
\textbf{Image Classification.} We next report results for patch attack detection in the case of image classification. Figure~\ref{roc-imgnet} shows the ROC curves for \method, \emph{NAPGuard}, \emph{Jedi}-detect, and \emph{Themis}-detect for detecting effective single-, double-, and quadruple-patch attacks on ImageNet; note that \emph{ObjectSeeker}-detect is exclusively applicable to object detection. We can observe that, except for \emph{Themis}-detect in the single-patch case, \method dominates the baseline attack detectors, achieving a higher detection rate and a lower false alarm rate. Moreover, for the corresponding CIFAR-10 results in Figure~\ref{roc-cifar} in the appendix, \method dominates all baselines, including \emph{Themis}-detect in the single-patch scenario.

It is important to note that the results obtained using \method in Figures~\ref{roc-imgnet} and~\ref{roc-cifar} show once again that \method performs consistently well irrespective of the number of patches, i.e., attack detection is insensitive to the number of patches. In contrast, the ability of the baselines to detect attacks (true positive rate) varies depending on number of patches. Interestingly, \emph{NAPGuard} has a better performance as the number of patches increases, while the other baselines perform worse as the number of patches increases. We show the attack detection accuracy for effective and ineffective attacks in the last twelve rows of Table~\ref{detperf-tab}. With the exception of \emph{Themis}-detect for effective single-patch attacks, the table confirms the superior performance of \method, especially for ineffective attacks and for multiple patches.


\begin{figure*}[!t]
\centering
\subfloat[\footnotesize Attack detection accuracy.]{\includegraphics[width =0.25\textwidth]{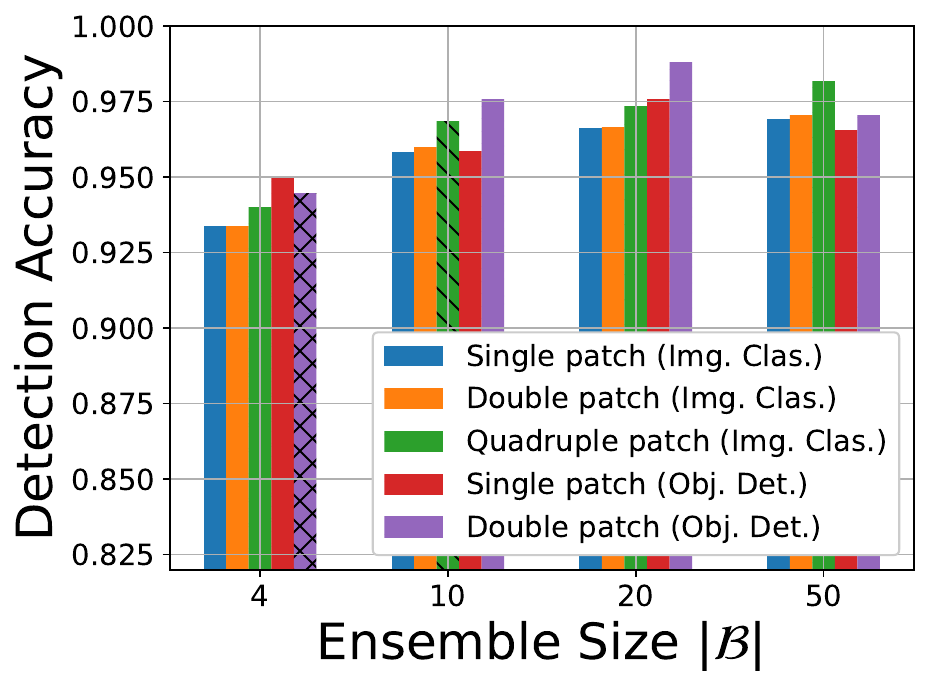}}\hspace{10mm}
\subfloat[\footnotesize Computation time (Obj.Det.).]{\includegraphics[width =0.28\textwidth]{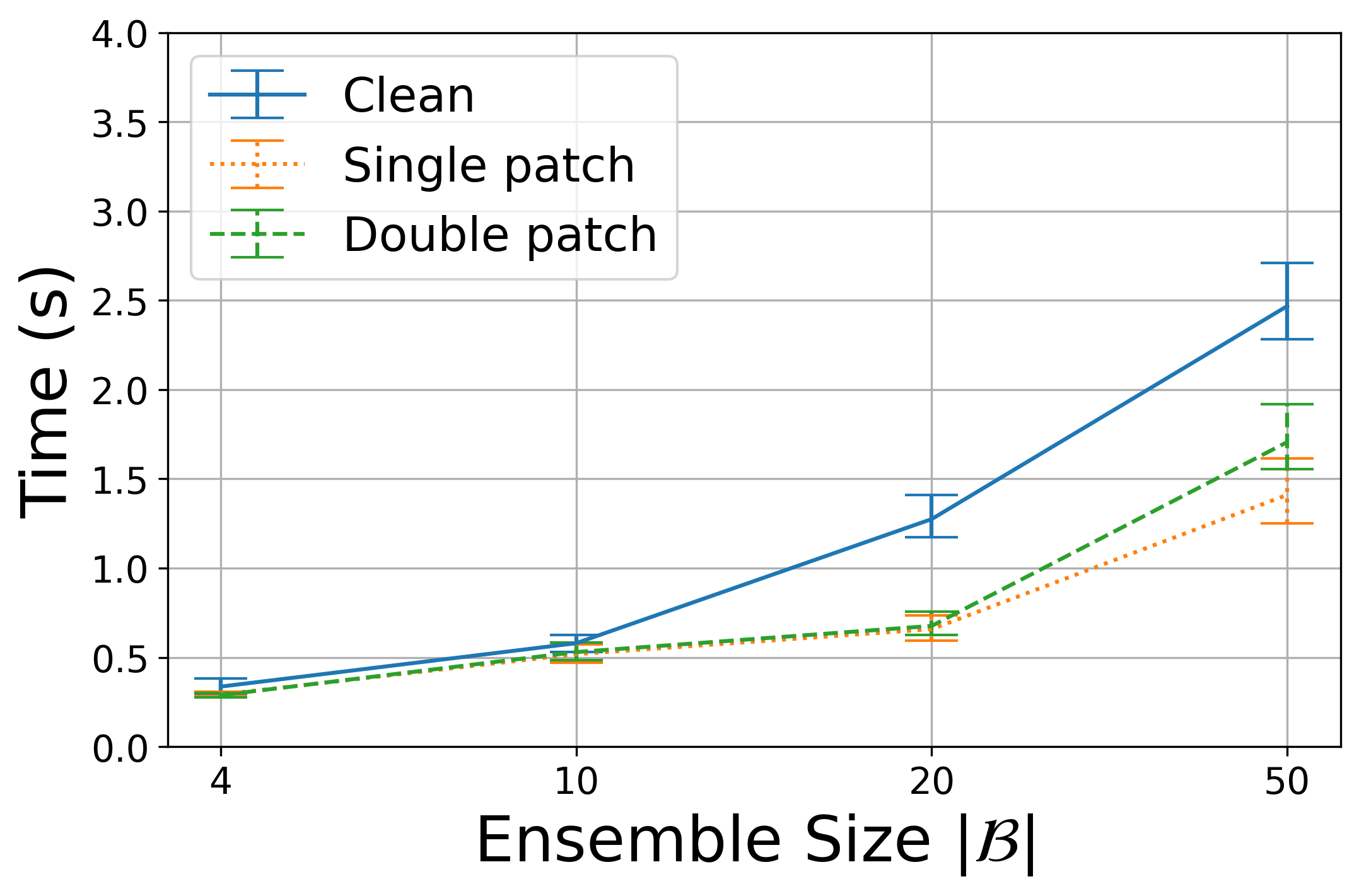}}\hspace{10mm}
\subfloat[\footnotesize Computation time (Img.Class).]{\includegraphics[width=0.28\textwidth]{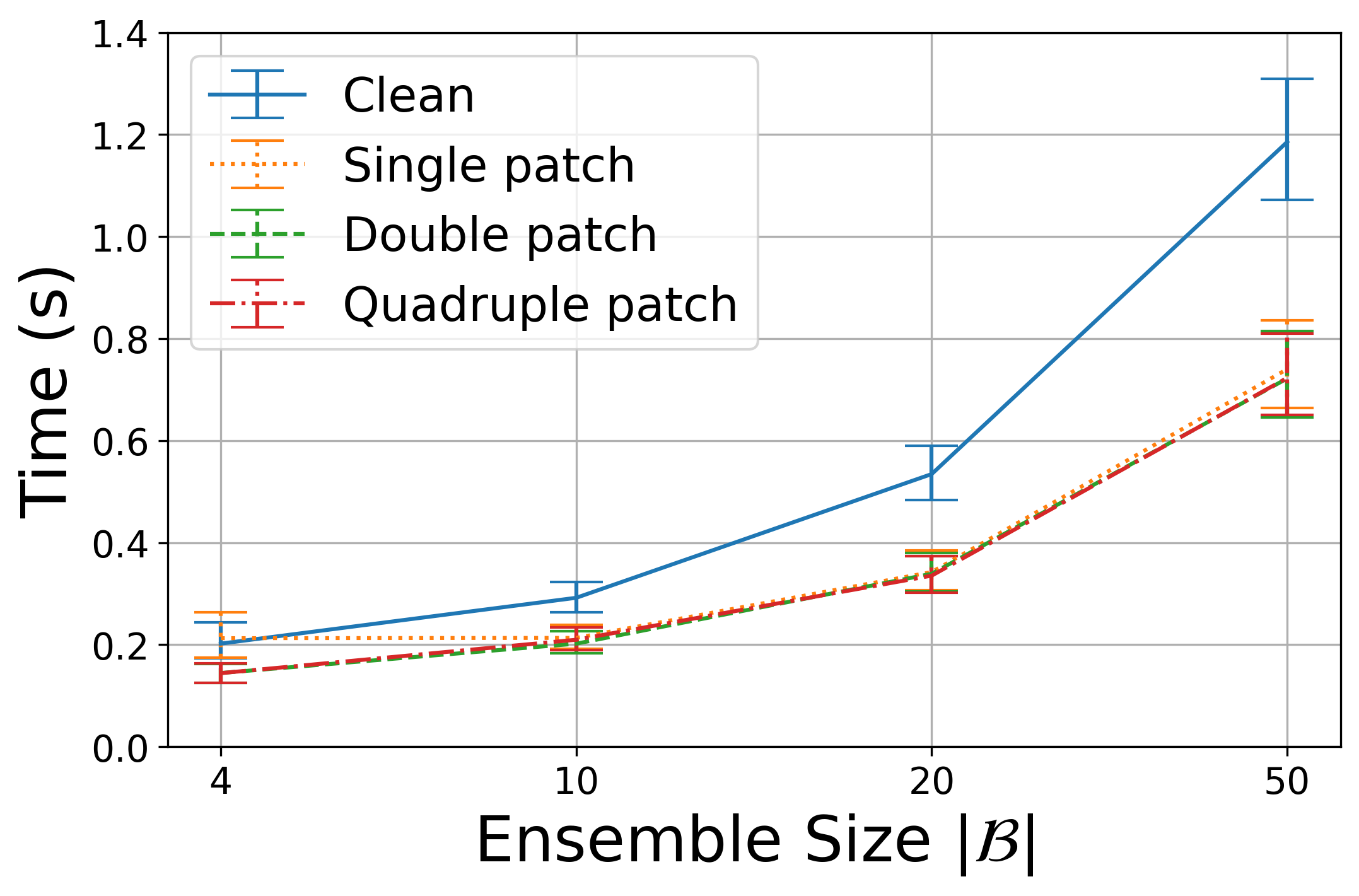}}
\caption{Accuracy (a) and computation time (b-c) vs. ensemble size $\vert\mathcal{B}\vert$ for \method. Error bars show first and third quartiles.}
\label{perf-complete}
\end{figure*}

\begin{figure*}[!t]
  \centering
\subfloat[\footnotesize \method.]{\includegraphics[width=0.35\textwidth, height=0.25\textwidth]{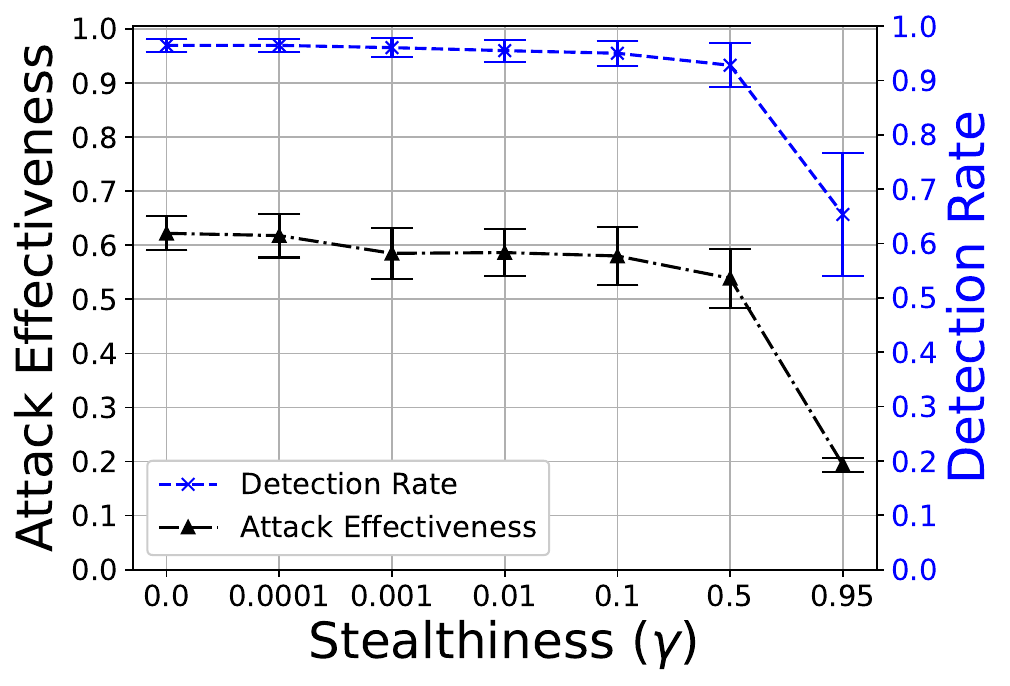}}\hspace{15mm}
\subfloat[\footnotesize Baselines.]{\includegraphics[width=0.35\textwidth, height=0.25\textwidth]{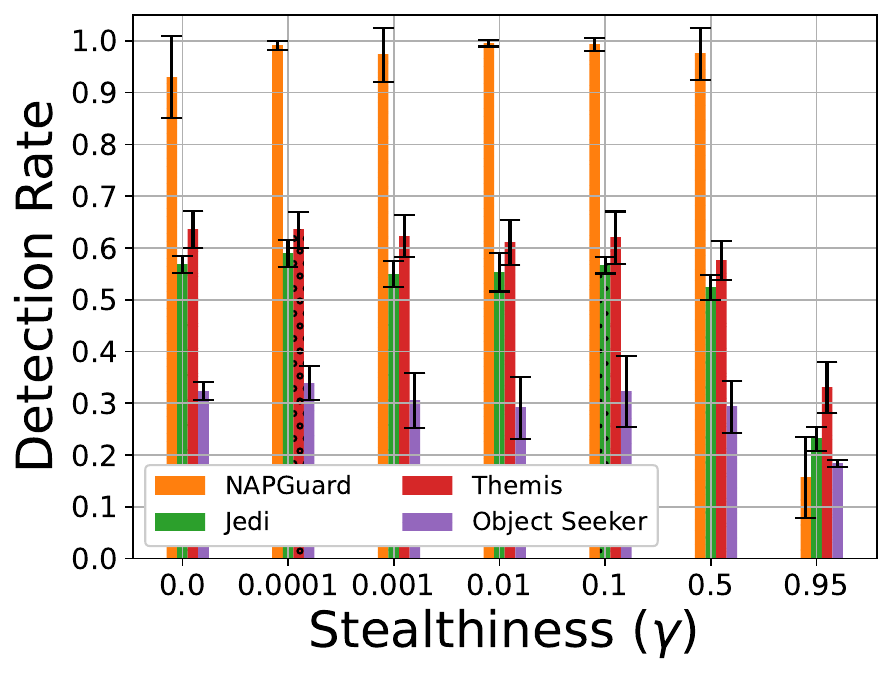}}
\caption{Attack detection (TP) and attack effectiveness vs. stealthiness of adaptive attack. Error bars show one standard deviation.}
\label{wb-complete}
\end{figure*}

\noindent
\textbf{Impact of the Ensemble Size.} Recall that the key tenet of the proposed detector is to use a set $\mathcal{B}$ of activity thresholds instead of a fixed threshold. Doing so avoids choosing a particular saliency threshold, making detection more efficient. At the same time, the cardinality of the set $\mathcal{B}$ has an impact on the computational burden of detecting an attack, as the cost of computing the clustering feature vector~$\mathbf{s}$ grows linearly in $|\mathcal{B}|$.
To characterize the tradeoff between attack detection performance and computational cost, we considered $4$ sets of saliency thresholds,
\begin{align*}
  \mathcal{B}_B:=\left\{\frac{b}{B}\right\}^{{B} - 1}_{b=0}, \quad B\in\{4,10,20,50\}.
\end{align*}

Figure~\ref{perf-complete}(a) shows  the attack detection accuracy on the INRIA (object detection) and ImageNet (classification) datasets as a function of the ensemble size $\vert\mathcal{B}\vert$.  Note that in this case the evaluation data for object detection is attacked using the adversarial patch by Thys et al.~\cite{foolAutoSurveillance}. The figure shows that, as one might expect, the attack detection accuracy increases as the ensemble size increases, yet the improvements become relatively small beyond $\vert\mathcal{B}\vert\geq 10$. Moreover, the accuracy is only slightly affected when decreasing the ensemble size to $4$, and in some scenarios increasing the ensemble size can even be deterimental (e.g., single- and double-patch attacks on object detection for $\vert\mathcal{B}\vert=50$). The results for Pascal VOC and CIFAR-10 in Figure~\ref{perf-complete2}(a) in the appendix are congruent with our analyses.


\noindent
\textbf{Computational Cost.} Next, we consider the computational cost of \method as a function of the ensemble size $\vert\mathcal{B}\vert$. Figures~\ref{perf-complete}(b)-(c) show the average computation time per image as a function of the ensemble size for the INRIA (object detection) and ImageNet (classification) datasets, respectively; we run our experiments on a system with 4 2x Intel Xeon Gold 6130 CPU cores and one NVIDIA T4 GPU. We can make three important observations from the figures. First, the computation time increases almost linearly with the ensemble size, implying that a small ensemble is preferred from a computational perspective. Second, the computational cost is slightly higher on clean images, as the neuron activations are more uniform, and thus, clustering is more computationally intensive. Finally, we observe that the computational cost of \method does not depend on the number of adversarial patches, unlike the computational cost of state-of-the-art methods~\cite{ObjSeek}. The figure also shows that attack detection takes longer for attacks against object detection, which is due to the larger image and feature map sizes used for object detection. Overall, the results show that an ensemble size of $\vert\mathcal{B}\vert=10$ provides a good tradeoff between attack detection accuracy and computation time. These observations are further supported by the corresponding results for Pascal VOC and CIFAR-10, available in Figures~\ref{perf-complete2}(b)-(c) in the appendix.

\subsection{Adaptive attacks}\label{sec:adpres}
Next, we evaluate the accuracy of the proposed detector against a powerful adversary that has access to our attack detection algorithm for creating effective patch attacks that can not be detected by \method. As a basis for the adaptive attack we use the attack model from Thys et al.~\cite{foolAutoSurveillance} but change the  loss function of the attacker to an adaptive loss $\mathcal{L}_a$:
\begin{align*}
    \mathcal{L}_{a}(\perturbation) &= (1-\gamma)\cdot\mathcal{L}(\perturbation) + \gamma\cdot\mathcal{L}_{\method}(\perturbation)\\
    \mathcal{L}_{\method}(\perturbation) &= \mathbb{E}_{\dataset}[\method(\transformation(\singlein, \perturbation))]
\end{align*}
Recall the general formulation for the original non-adaptive loss function $\mathcal{L}(\perturbation)$ in Section~\ref{sec:preliminaries}: $\singlein$ is in the dataset $\dataset$ over which the attack is optimized,  hence $\method(\transformation(\singlein, \perturbation))$ is the detection score \method assigns to an input $\singlein$ perturbed under the attack model with patch $\perturbation$. The parameter $\gamma$ controls the stealthiness of the attack, i.e., it determines how much an attacker prioritizes evading \method. 

For the evaluation we focus on single-patch attacks on object detection on the INRIA dataset. We trained adaptive attacks for $\gamma \in \{0.0, 0.0001, 0.001, 0.01, 0.1, 0.5, 0.95\}$. We train five separate patches for each value of $\gamma$. Figure~\ref{wb-complete}(a) shows the resulting attack effectiveness on the undefended model (i.e., the fraction of images in the INRIA test set which are successfully attacked) and the true positive rate achieved by \method (on both effective and ineffective attacks), as a function of the stealthiness weight $\gamma$. We do not show the false positive rate as the adaptive attack does not affect that. 
The figure shows that adapting the patch attack to be undetected by \method  results in decreased attack effectiveness. Comparing the curves for attack effectiveness and the true positive rate as a function of $\gamma$  we can observe that when the stealthy attack is able to compromise detection, its effectiveness decreases faster than \method's true positive rate, indicating that \method is robust to the adaptive attacker.

We also evaluate all baselines against the adaptive attack. Note that the adaptive attack was not optimized against the baseline defenses, it was optimized to bypass \method. We report the true positive rates achieved by the baselines as a function of $\gamma$ in Figure~\ref{wb-complete}(b). The figure shows that the adaptive attack is also able to bypass the baselines, and in the case of \emph{Jedi}-detect and \emph{Themis}-detect, our stealthy attack is able to reduce their detection capabilities well beyond what was reported in their original papers regarding adaptive or defense-aware attackers, even though the attack was not optimized against these defense schemes~\cite{themis,jedi}. The figure also shows that \emph{ObjectSeeker}-detect is affected by the adaptive attack despite its masking mechanism being oblivious to the patch attack's content; in particular, this result highlights how its detection rate depends on attack effectiveness. While \emph{NAPGuard} is not affected for most values of $\gamma$, it suffers a dramatic drop in detection rate at $\gamma=0.95$, noticeably beyond that of \method. These results confirm that \method is robust to adaptive attacks and maintains a higher detection rate than the baselines even for a stealthy attacker targeting \method.

\begin{figure*}[!t]
\centering
\subfloat[\footnotesize INRIA.]{\includegraphics[width =0.22\textwidth]{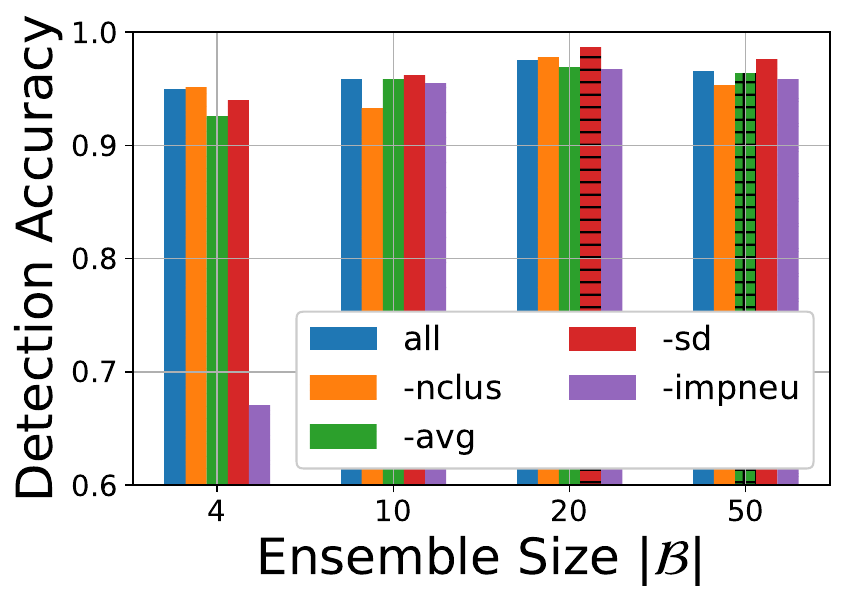}}\hfil
\subfloat[\footnotesize Pascal VOC.]{\includegraphics[width =0.22\textwidth]{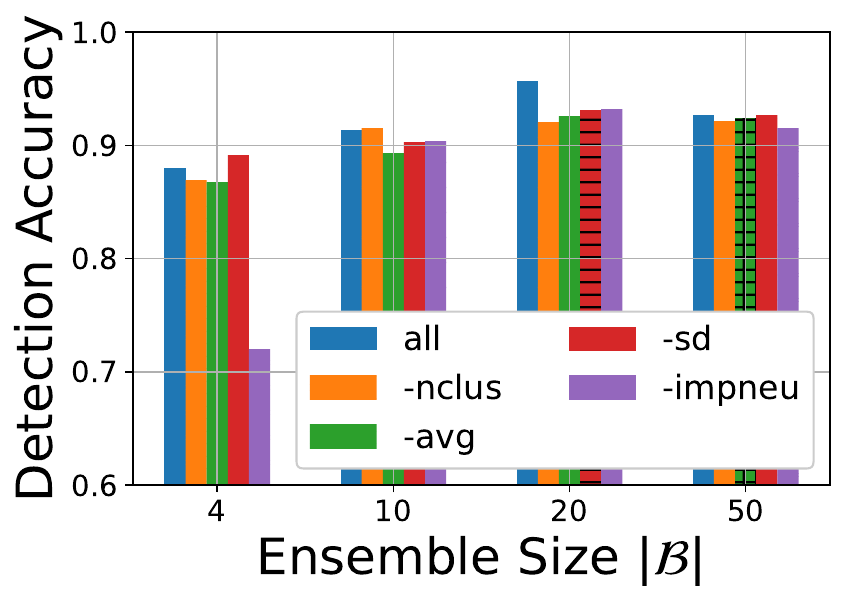}}\hfil
\subfloat[\footnotesize ImageNet.]{\includegraphics[width =0.22\textwidth]{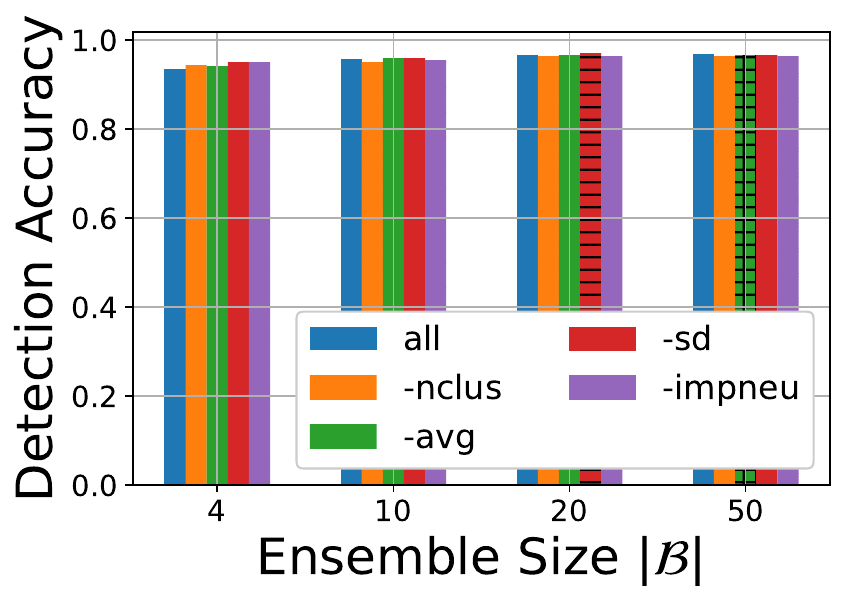}}\hfil
\subfloat[\footnotesize CIFAR-10.]{\includegraphics[width=0.22\textwidth]{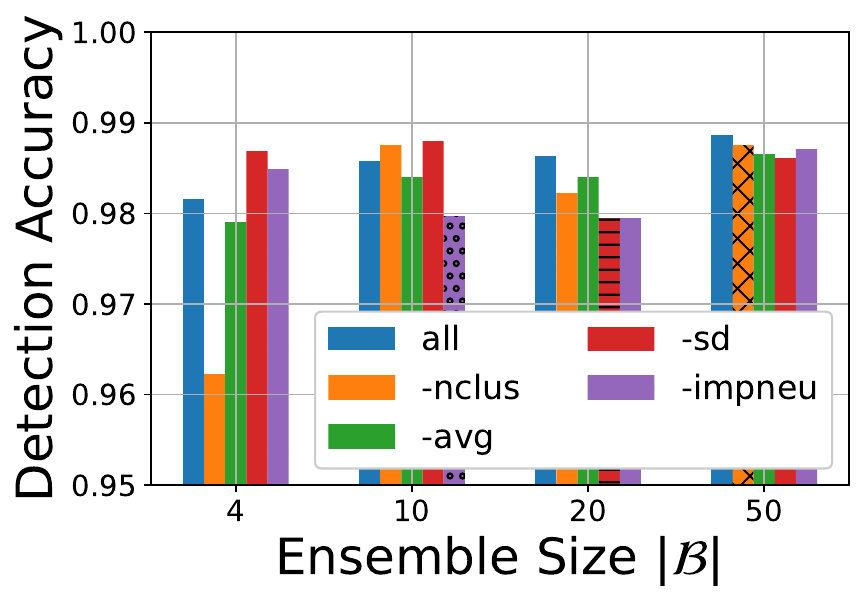}}
\caption{Impact on \method's attack detection accuracy vs. ensemble size $\vert\mathcal{B}\vert$ after dropping each of the four clustering features: number of clusters (\emph{nclus}), mean average intra-cluster distance (\emph{avg}), standard deviation of average intra-cluster distance (\emph{sd}), and number of important neurons (\emph{impneu}). The default case using all features is denoted by \emph{all}.}\label{ablation-plots}
\end{figure*}

\subsection{Choice of Clustering Features}
In Section~\ref{sec:preliminaries}, we provided the intuition behind the use of our proposed clustering features to detect patch attacks. In what follows we use SHAP values computed using the Kernel SHAP algorithm\footnote{The SHAP value is a commonly used measure of feature importance, and Kernel SHAP is an efficient method for approximating SHAP values~\cite{kernelshap}.} to quantify the importance each clustering feature fed into $AD$ has in accurate detection. Note that here we investigate feature importance for all four datasets introduced in Section~\ref{sec:setup}, not only INRIA and ImageNet. 

Recall that an attack detector \emph{AD} is trained for each dataset, and hence each dataset has been split into training, validation, and test sets. To obtain a measure of how important each clustering feature is, we use KernelSHAP to explain the difference between the detection score corresponding to an all-zero input and the score corresponding to each input in the validation set. We set Kernel SHAP to use 500 samples for the explanation of any single validation input. 

\begin{figure}[!t]
\centering
\includegraphics[scale=0.5]{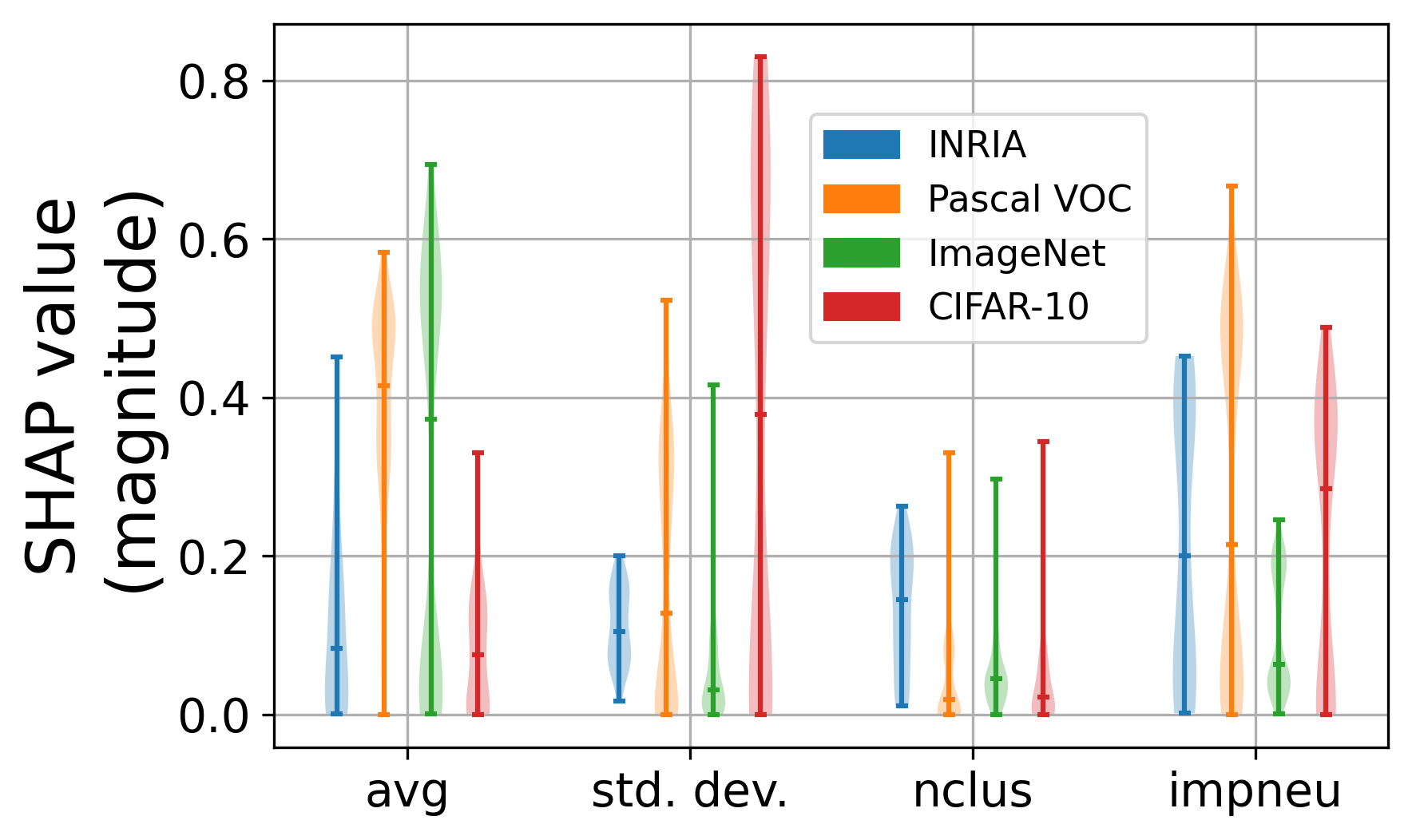}
\caption{Violin plots for feature importance calculated with Kernel SHAP for the proposed clustering features: average mean intra-cluster distance (\emph{avg}),  mean intra-cluster distance standard deviation (\emph{std. dev.}), number of clusters (\emph{nclus}), and number of important neurons (\emph{impneu}). Markers indicate minimum, median, and maximum over the validation set.} 
\label{shap}
\vspace{-3mm}
\end{figure}

Figure~\ref{shap} shows the results obtained for the four datasets: the vertical axis indicates the \emph{magnitude} of the estimated SHAP values, which are plotted over the validation set. Note that the magnitude is used to focus on the overall impact of each feature on the output (i.e., on the detection score), and not on whether it reduces or increases the value of said output. SHAP values are grouped by the feature they correspond to, in order to highlight the importance of each feature in determining the attack detection scores across the validation set. We make two main observations from these figures. First, which of the features are more important depends not only on the task, but also on the dataset. Second, despite such context dependence, there is no dataset for which any particular feature would be unimportant: considering that the detection scores are between 0 and 1, each feature is important for instances from all datasets. We thus conclude that each of the proposed clustering features contribute to accurate attack detection, and each feature may be more or less useful depending on the context under which an adversarial attack takes place.

We further perform an ablation of \method, where we drop each clustering feature, and then retrain and evaluate $AD$ using the same procedure described in Section~\ref{sec:adparams}. The architecture of $AD$ is unchanged with the exception of the input layer, which contains three channels instead of four for each of the different versions of $AD$ used in this ablation study; for each version, one of the clustering features is dropped, hence only three input channels are needed for the first layer. We focus on single-patch attacks and we use the attack proposed by Thys et al.~\cite{foolAutoSurveillance} for INRIA and Pascal VOC.

We present results for all datasets in Figure~\ref{ablation-plots}, which shows the detection accuracy for different ensemble sizes.  We observe that in three out of four datasets, the best performing configuration corresponds to the original model using all four clustering features. The exception is the INRIA dataset, where the model without the standard deviation of the average intra-cluster distance (the \emph{sd} feature in the figure) achieves a detection accuracy slightly above that of the other models. Hence in most cases, dropping any of the chosen clustering features limits the best detection accuracy achieved by \method. Moreover, the results for CIFAR-10 in Figure~\ref{ablation-plots}(d) show that dropping one of the proposed clustering features can lead to an unstable relation between ensemble size and detection accuracy, e.g., note how the model dropping \emph{sd} experiences a notable performance drop as $\vert\mathcal{B}\vert$ goes from 10 to 20, and then goes back up at $\vert\mathcal{B}\vert=50$, yet this final performance is still below that at $\vert\mathcal{B}\vert=10$; this behavior is rather counter-intuitive. From this ablation study we conclude that all features contribute to accurate detection and a stable relation between detection accuracy and ensemble size in \method. 

\subsection{Unsupervised Attack Detection}\label{sec:numres-occ}
\begin{table*}
\centering
\caption{Overall attack detection accuracy of \method and its OCC variant.}\label{tab-occ}
\begin{tabular}{lcccccccc}
\hline
 & \multicolumn{2}{c}{$\vert\mathcal{B}\vert = 4$} & \multicolumn{2}{c}{$\vert\mathcal{B}\vert = 10$} & \multicolumn{2}{c}{$\vert\mathcal{B}\vert = 20$} & \multicolumn{2}{c}{$\vert\mathcal{B}\vert = 50$} \\
\textbf{Attack} & \textit{Default} & \textit{OCC (DM-NAP)} & \textit{Default} & \textit{OCC (DM-NAP)} & \textit{Default} & \textit{OCC (DM-NAP)} & \textit{Default} & \textit{OCC (DM-NAP)} \\ \hline
Single-patch (INRIA) & 0.9497 & 0.8160 (0.6997) & 0.9583 & 0.9306 (0.8420) & 0.9757 & 0.9618 (0.8750) & 0.9653 & \textbf{0.9861} (0.9184) \\
Double-patch (INRIA) & 0.9444 & 0.8368 (0.7604) & 0.9757 & 0.9549 (0.8906) & 0.9878 & 0.9757 (0.9271)& 0.9705 & \textbf{0.9965} (0.9566)\\

Single-patch (VOC) & 0.8799 & 0.5423 (0.5147) & 0.9137 & 0.8081 (0.7642) & \textbf{0.9567} & 0.8417 (0.7174) & 0.9266 & 0.8491 (0.6240) \\
Double-patch (VOC) & 0.8706 & 0.5717 (0.5332) & 0.9281 & 0.8213 (0.7978) & \textbf{0.9645} & 0.8675 (0.7637) & 0.9318 & 0.8802 (0.7637)\\ \hline
Single-patch (ImageNet) & 0.9339 & 0.9364 & 0.9584 & 0.9192 & 0.9662 & 0.9378 & \textbf{0.9693} & 0.9478 \\
Double-patch (ImageNet) & 0.9338 & 0.9373 & 0.9597 & 0.9185 & 0.9664 & 0.9373 & \textbf{0.9703} & 0.9485 \\
Quadruple-patch (ImageNet) & 0.9399 & 0.9466 & 0.9684 & 0.9272 & 0.9733 & 0.9462 & \textbf{0.9816} & 0.9586 \\
Single-patch (CIFAR-10) & 0.9815 & 0.8735 & 0.9857 & 0.9829 & 0.9863 & 0.9717 & \textbf{0.9886} & 0.9738 \\
Double-patch (CIFAR-10) & 0.9800 & 0.7804 & 0.9890 & 0.9867 & 0.9886 & 0.9758 & \textbf{0.9914} & 0.9788 \\
Quadruple-patch (CIFAR-10) & 0.9637 & 0.7421 & 0.9992 & 0.9975 & 0.9975 & 0.9911 & \textbf{0.9996} & 0.9915 \\ \hline
\end{tabular}
\end{table*}

\method makes no assumptions on the shape, size, or number of patches, and the supervised training of the attack detector network \emph{AD} is quite sample efficient compared to, e.g., that of NAPGuard~\cite{napguard}. Moreover, the evaluations conducted so far, particularly in the context of object detection, show that \method can effectively detect unseen patch attacks (see the appendix for further results on the GAP dataset). However, relying on specific attack models is a common limitation of adversarial patch defenses that prior works have pointed out~\cite{ObjSeek, PAD, nutnet}. Therefore, to explore the potential of \method to perform unsupervised patch attack detection, we retrained \emph{AD} with the same architecture and training procedure, but using clean data samples only and providing, for each clean sample, a random input labelled as an adversarial example during training. Hence, the attack detection problem shifts from a binary classification setting into a one-class classification problem akin to anomaly detection. Following prior work~\cite{oc-cnn}, the random inputs labeled as adversarial (or anomalous) are sampled from a normal distribution, which we normalize between -1 and 1 before feeding them into \emph{AD}. 

Table~\ref{tab-occ} shows the best overall detection accuracy (i.e., over both effective and non-effective attacks) achieved by the default method and the proposed one-class classification (OCC) variant, for different ensemble sizes $\vert\mathcal{B}\vert$. Note that we used the attack by Thys et al.~\cite{foolAutoSurveillance} for object detection in this experiment, and for completeness we also show the performance of the OCC variant on the \emph{DM-NAP-Princess} patch.

The table shows that in general, using adversarial samples during training (i.e., the \emph{default} setting) leads to a higher attack detection accuracy. For object detection, the proposed OCC variant surprisingly performs on-par or even better than the default approach on the INRIA dataset, but is notably outperformed by the default on Pascal VOC. For image classification, the default once again outperforms the OCC variant, but the latter is still able to attain a relatively high accuracy on both ImageNet and CIFAR-10, and from Table~\ref{detperf-tab} we can observe that the OCC variant still outperforms all baselines in terms of overall accuracy. Moreover, the results for the OCC variant on the \emph{DM-NAP-Princess} patch show that it also outperforms all baselines for object detection attacks, except for single patches on VOC, where \emph{Themis} has a slightly higher overall accuracy (the attack effectiveness rates in Table~\ref{tab-atkeff} in the appendix enable our comparisons to the baselines in terms of overall accuracy).  We conclude that the clustering features used for classification by \method are a useful representation of the input data, and exploring more elaborate OCC approaches could further bridge the gap with our default supervised approach.

\section{Conclusion}
\label{sec:conc}
In this work, we propose \method, a patch attack detection method. \method needs no prior information about the number of patches, neither does it rely on a fixed saliency threshold to detect attacks, thereby overcoming shortcomings of existing defenses. Compared to state-of-the-art baselines, \method achieves superior patch-attack detection performance for object detection and image classification tasks, and its performance and computational costs are independent of the number of patches. Our results obtained using an adaptive attacker show that bypassing \method results in a large reduction of attack effectiveness, and our unsupervised attack detection results show that beyond detecting unseen patches effectively, \method can achieve a remarkable performance using only clean images during training. We conjecture that the clustering features introduced in \method could be leveraged for attack identification and recovery as well; we leave this to be the subject of future work.

\section*{Acknowledgement}
This work was partly funded by the KTH Railway Group. We acknowledge the National Academic Infrastructure for Supercomputing in Sweden (NAISS), partially funded by the Swedish Research Council through grant agreement no. 2022-06725, for computational and storage resources, and for awarding this project access to the LUMI supercomputer, owned by the EuroHPC Joint Undertaking and hosted by CSC (Finland) and the LUMI consortium.
{
    \small
    \bibliographystyle{ieeetr}
    \bibliography{main}
}


\clearpage
\appendix
\section{Appendix}
\subsection{Results on Pascal VOC and CIFAR-10}
\label{sec:add-res}
\begin{figure}[H]
\centering
\includegraphics[width = 0.34\textwidth]{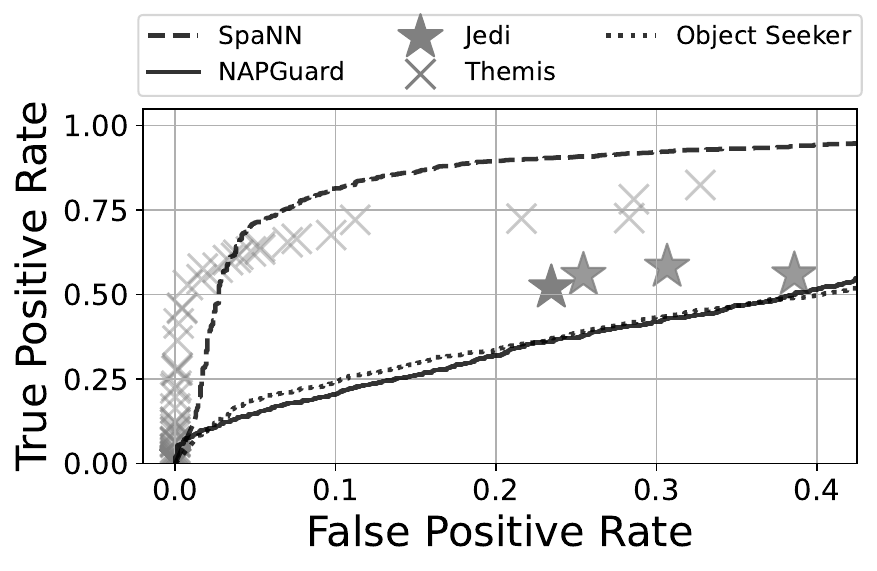}
\includegraphics[width=0.34\textwidth]{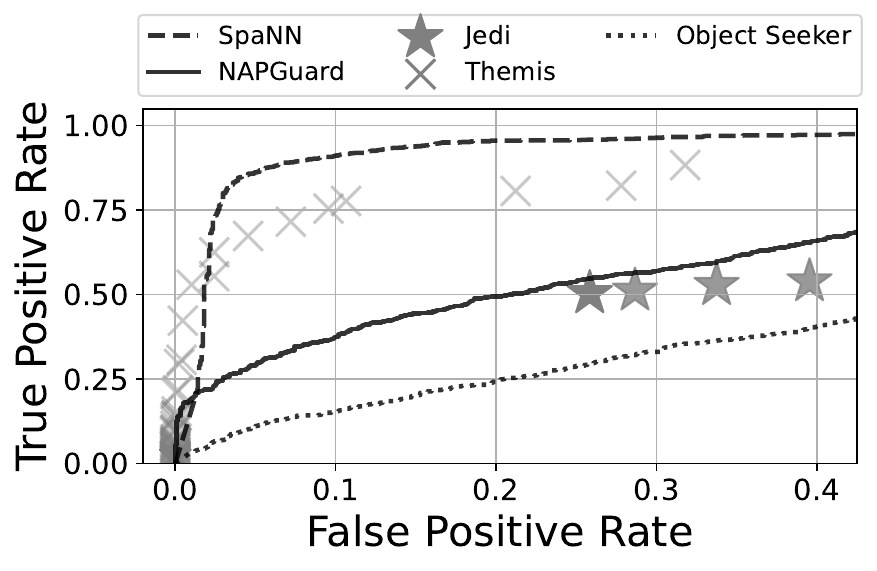}
\caption{Attack detection and false alarm rates for single (top) and double (bottom) adversarial patch detection for object detection (Pascal VOC).}\label{roc-voc}
\end{figure}

\begin{figure}[h]
\centering
\includegraphics[width=0.34\textwidth]{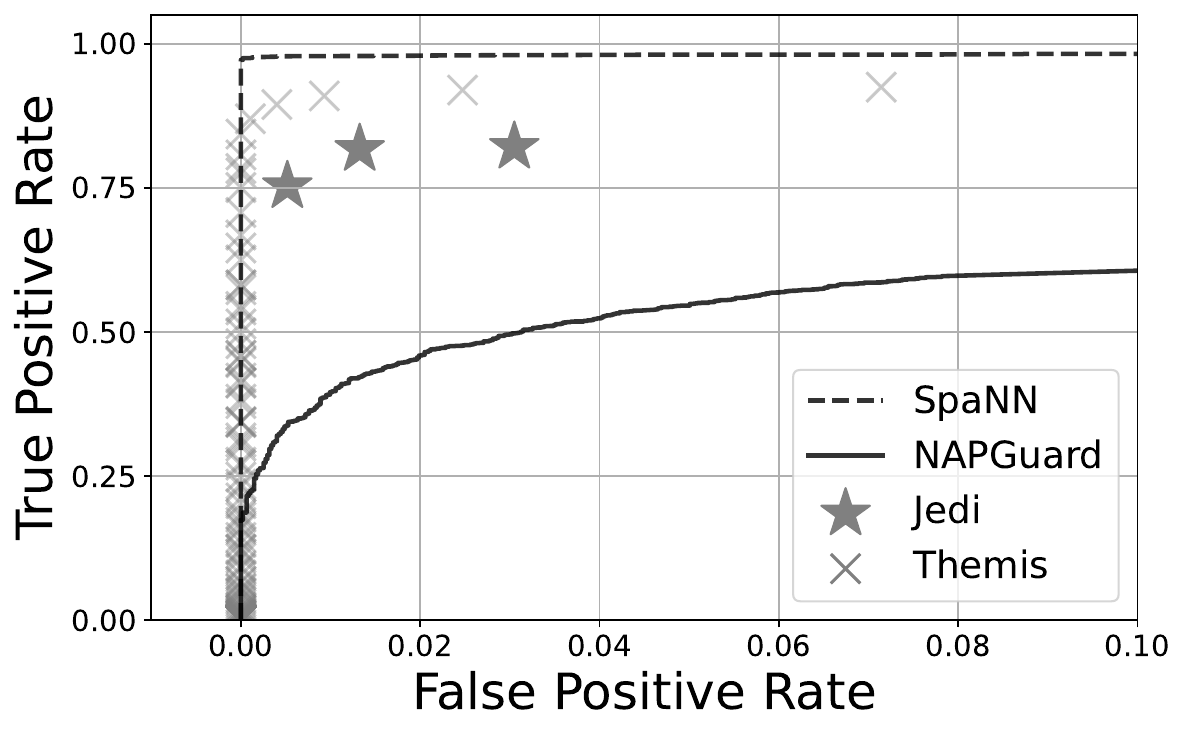}
\includegraphics[width=0.34\textwidth]{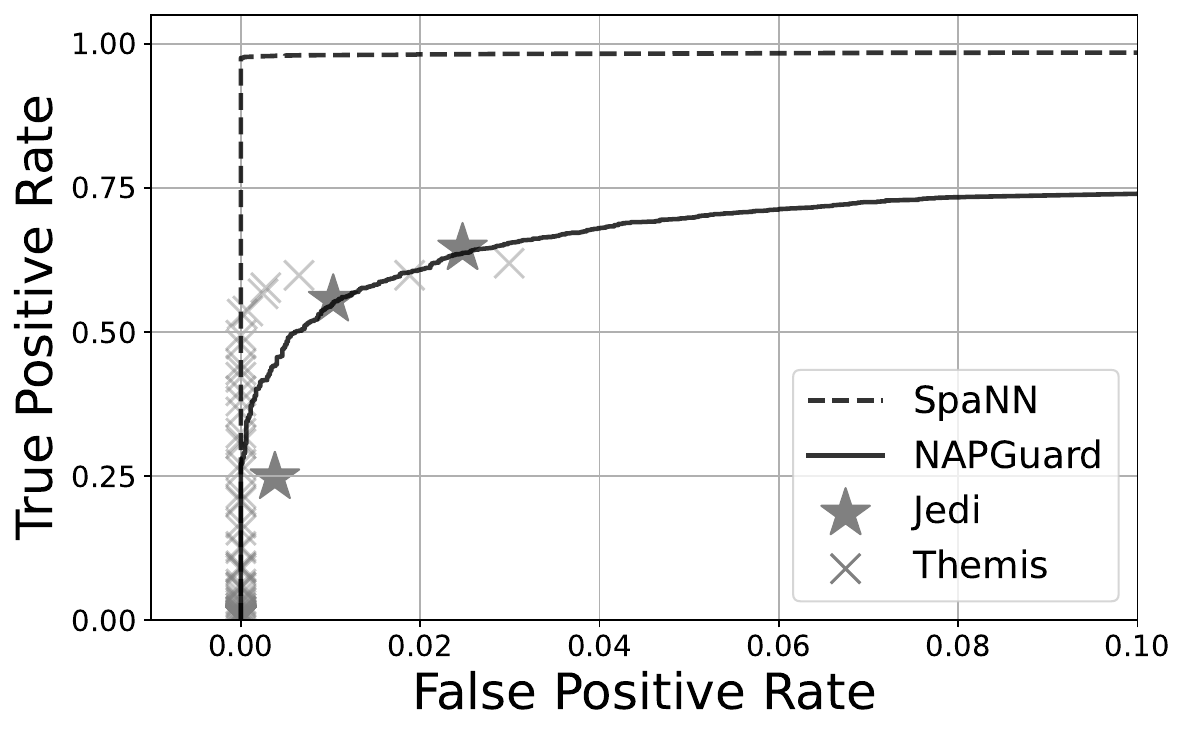}
\includegraphics[width=0.34\textwidth]{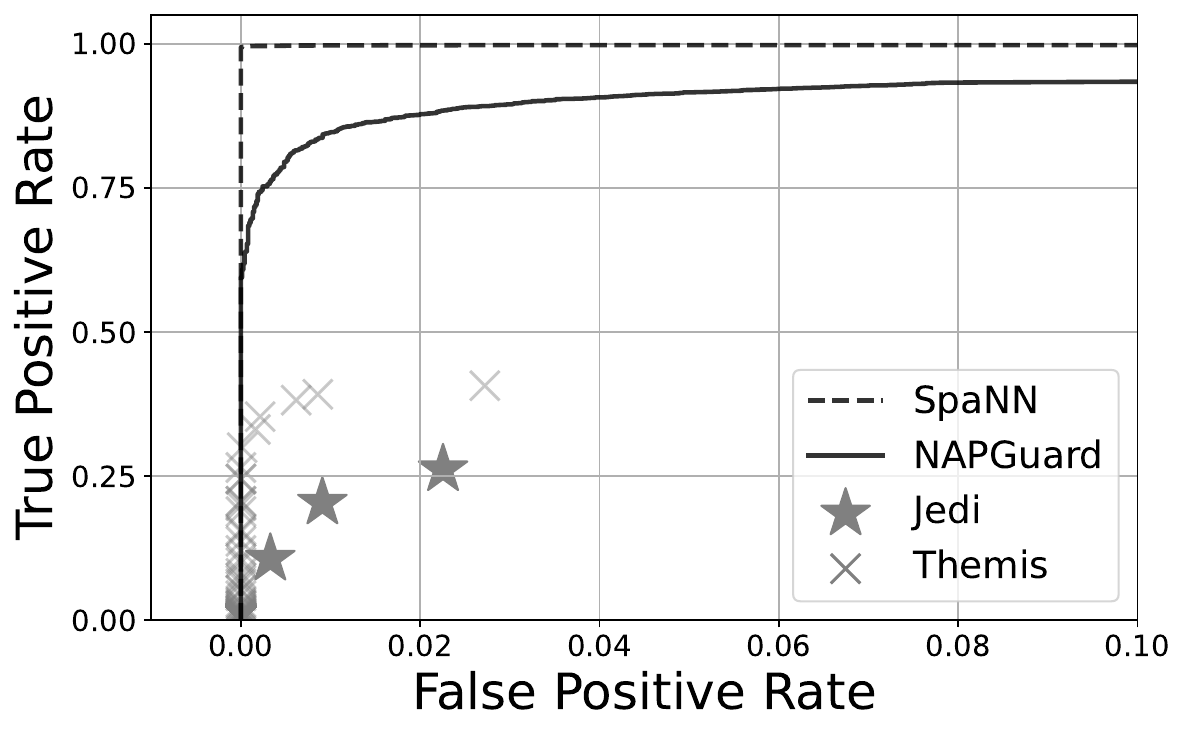}
\caption{Attack detection and false alarm rates for single (top), double (middle), and quadruple (bottom) adversarial patch detection for image classification (CIFAR-10).}\label{roc-cifar}
\end{figure}

In this appendix section we report the results on Pascal VOC and CIFAR-10 referenced in the paper. Note that the same DBSCAN parameters and training setup described in Section~\ref{sec:adparams} are used in all datasets. The ROC curves obtained using \method and using the baseline defenses for object detection (Pascal VOC) are shown in Figure~\ref{roc-voc}, which are consistent with the results for INRIA in Figure~\ref{roc-inria}. We also present the ROC curves for CIFAR-10 in Figure~\ref{roc-cifar}, which are consistent with the ImageNet results in Figure~\ref{roc-imgnet}.

\begin{figure*}[h]
\centering
\subfloat[\footnotesize Attack detection accuracy.]{\includegraphics[width=0.25\textwidth]{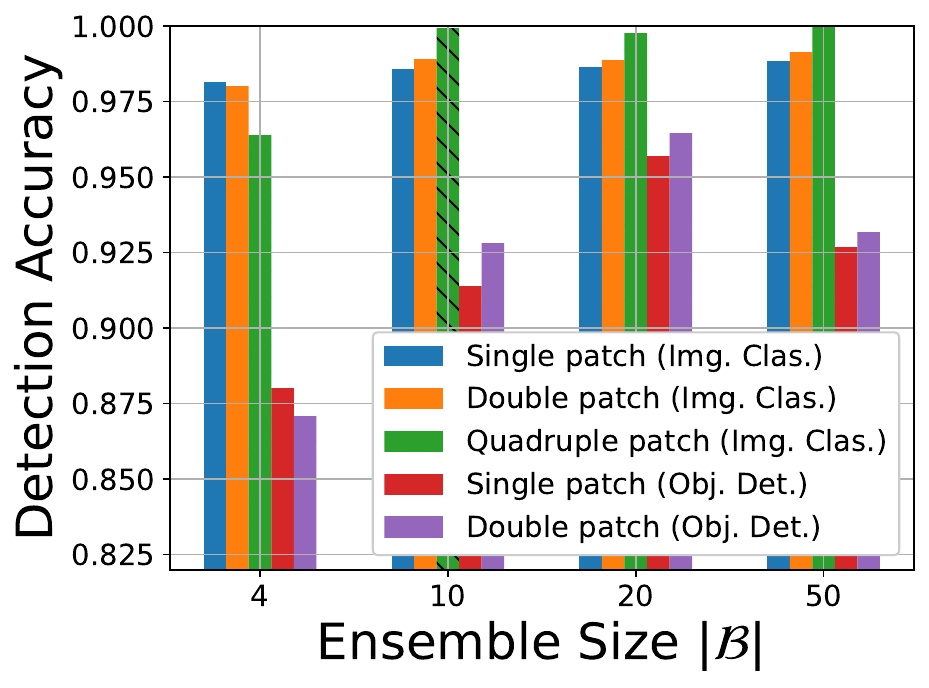}}\hspace{5mm}
\subfloat[\footnotesize Computation time (Obj.Det.).]{\includegraphics[width =0.25\textwidth]{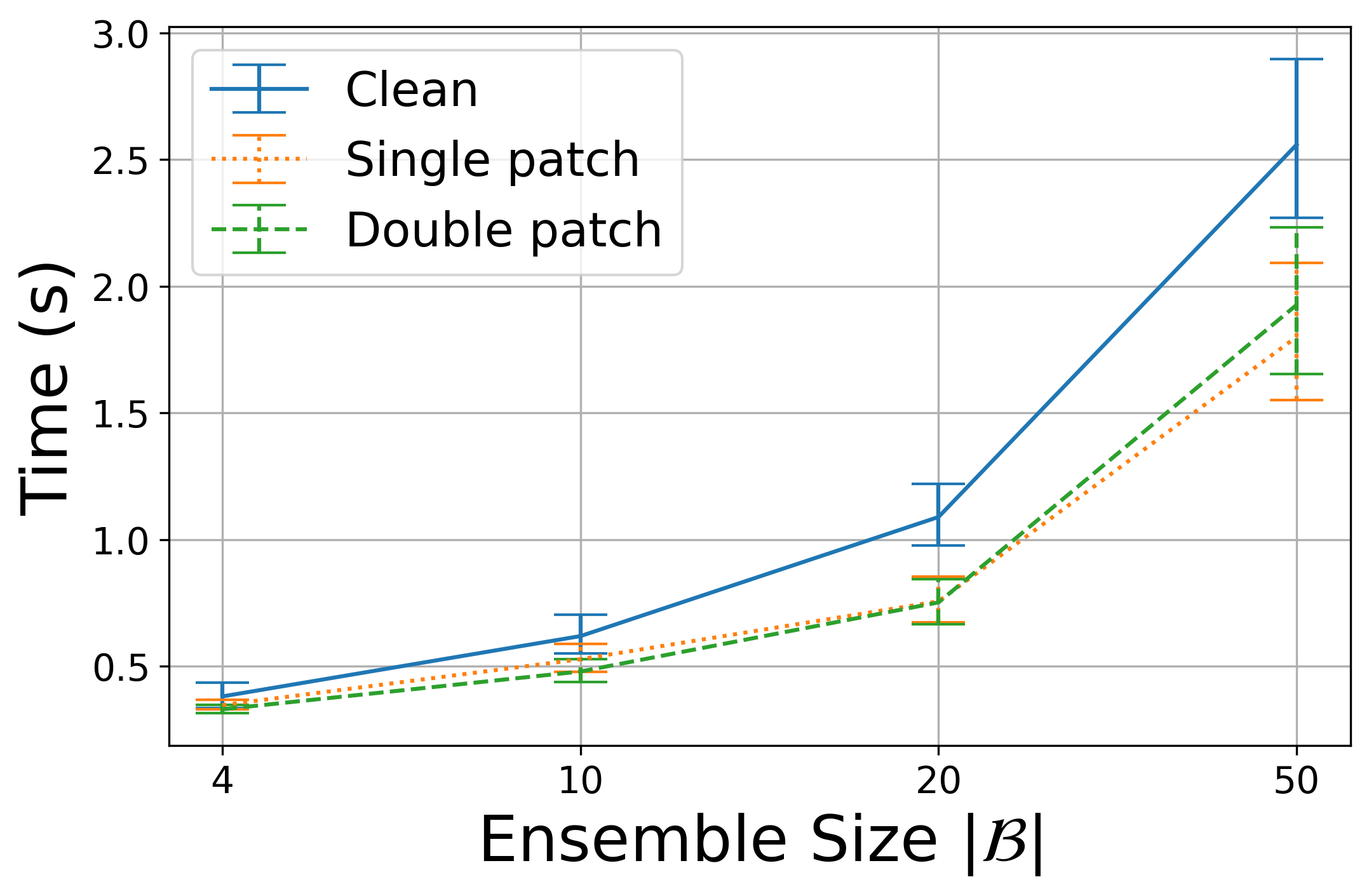}}\hspace{5mm}
\subfloat[\footnotesize Computation time (Img.Class).]{\includegraphics[width=0.25\textwidth]{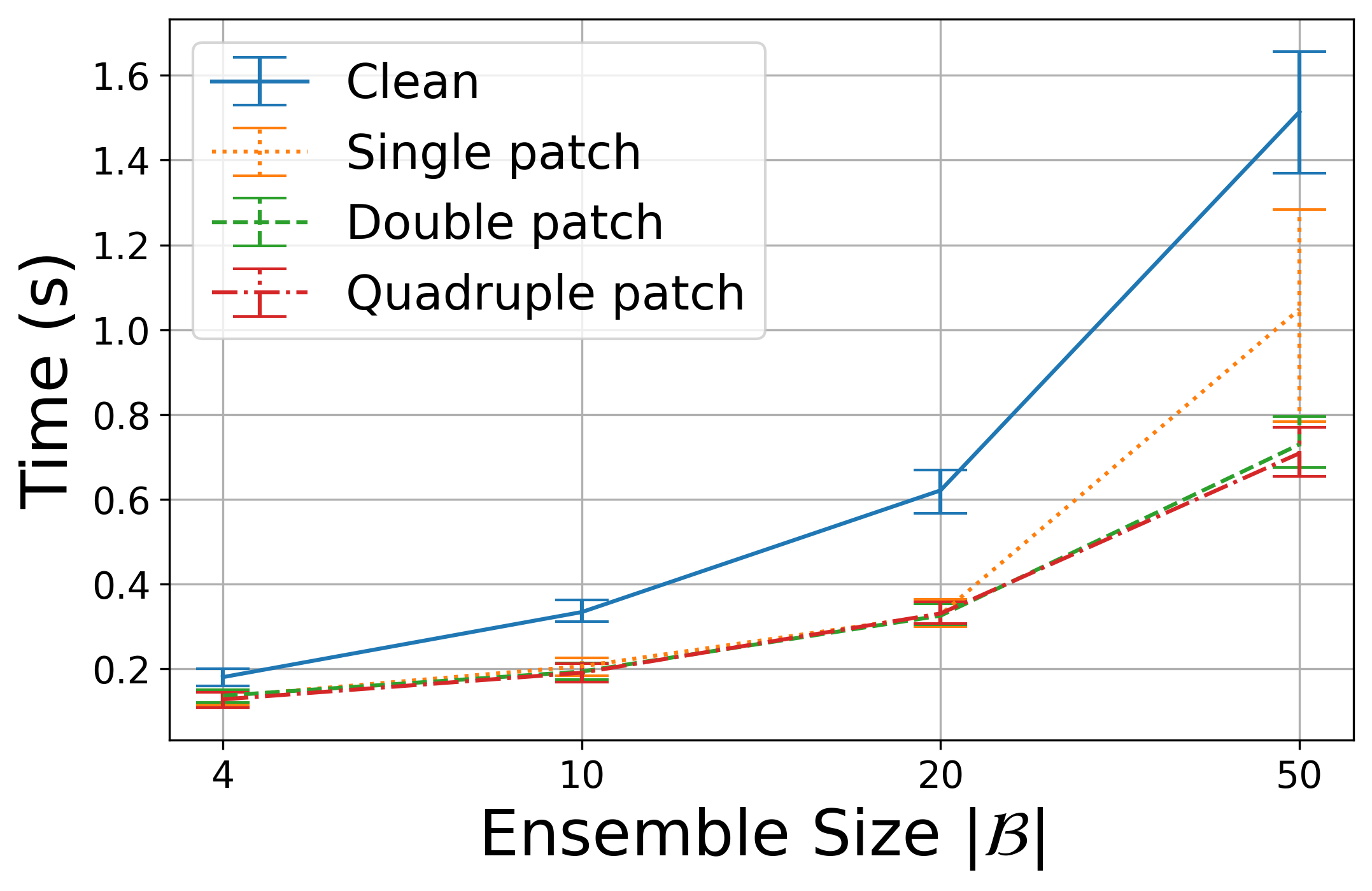}}
\caption{\method's attack detection accuracy (a) and computation time (b-c) vs. ensemble size $\vert\mathcal{B}\vert$, using Pascal VOC and CIFAR-10.}
\label{perf-complete2}
\end{figure*}

To complete our evaluation of the impact of the ensemble size and the resulting computational cost, we report results obtained with Pascal VOC (object detection) and CIFAR-10 (image classification) in Figure~\ref{perf-complete2}.  Figure~\ref{perf-complete2}(a) shows the attack detection accuracy, and confirms that increasing the ensemble can boost performance, but and increase beyond $\vert \mathcal{B}\vert = 10$ yields only relatively small gains in some scenarios, which is consistent with the results shown in Figure~\ref{perf-complete}(a). Moreover, the object detection results in Figure~\ref{perf-complete2}(a) confirm that further increasing $\vert \mathcal{B}\vert$ might even be detrimental; we conjecture the counterintuitive drop in performance for object detection for $\vert\mathcal{B}\vert = 50$ indicates that \method may overfit after a certain granularity for a fixed amount of training data. Regarding the computational cost as a function of $\vert \mathcal{B} \vert$, in accordance with the results shown in Figures~\ref{perf-complete}(b) and~\ref{perf-complete}(c), Figures~\ref{perf-complete2}(b) and~\ref{perf-complete2}(c) show that for object detection and for image classification, \method's running time increases as $\vert \mathcal{B} \vert$ increases with an approximately linear rate, and the computational cost of \method does not depend on the number of patches, as long as there are patches. At the same time, the computational time is higher for clean images. These results are also consistent with our observations made based on Figures~\ref{perf-complete}(b)-(c), i.e., the results are consistent across multiple datasets.

\subsection{Results on GAP Dataset}
The GAP dataset was released as a benchmark to evaluate \emph{NAPGuard} along other baseline detection methods, and contains 25 different types of patch attacks applied to data from the INRIA and COCO datasets~\cite{napguard}. All the attacks in the dataset are single-patch attacks applied to one or more objects, and they are split into three levels, GL1, GL2, and GL3, depending on how difficult it is for an attack detector to generalize to each type of patch (GL1 being the least difficult and GL3 the most difficult). To further assess \method's performance on unseen patches (beyond those used in our main evaluations), we compare to \emph{NAPGuard} on the GL2 and GL3 partitions; since \emph{NAPGuard} has access to attacks from GL1 during training, we do not consider that partition. Unlike the datasets used in our previous evaluations, which are balanced in terms of clean and attacked images, the GAP dataset contains only a few images without adversarial patches, in particular, the GL2 partition contains 828 images attacked with 8 different patches and 22 clean images, while the GL3 partition contains 584 images attacked with 6 different patches (including \emph{DM-NAP-Princess}) and 16 clean images. Note that in the GAP dataset only one type of patch attack is used per attacked image~\cite{napguard}.

\begin{figure}
\centering
\includegraphics[width = 0.34\textwidth]{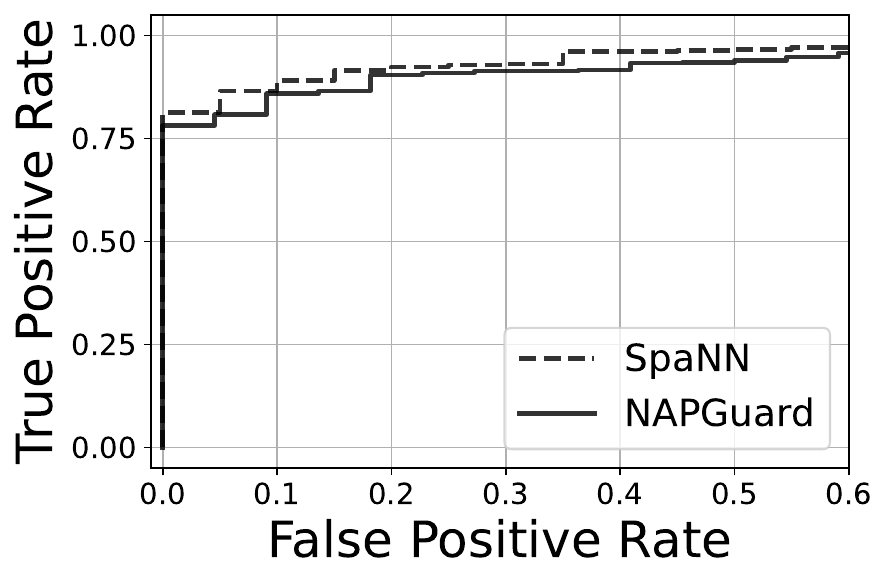}
\includegraphics[width=0.34\textwidth]{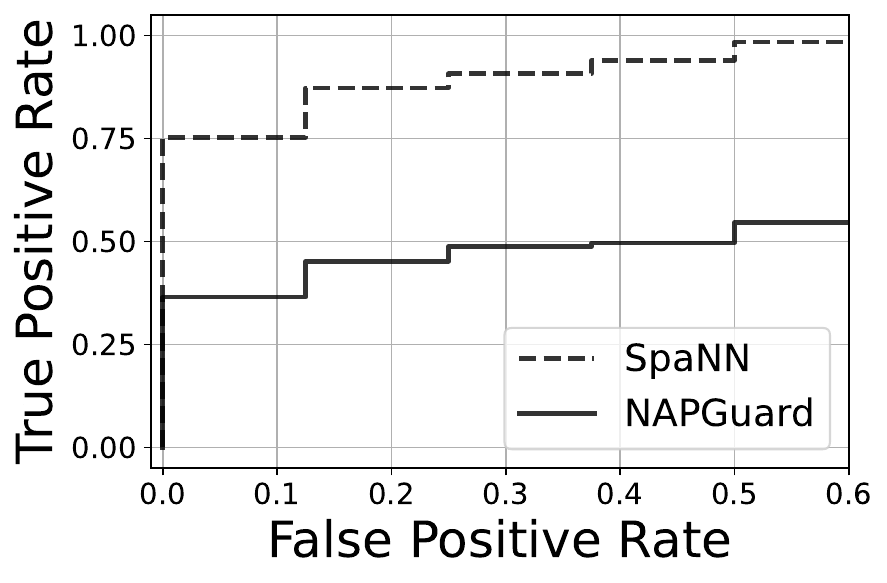}
\caption{Attack detection and false alarm rates for the GL2 (top) and GL3 (bottom) partitions from the GAP dataset.}\label{gl-results}
\end{figure}

In Figure~\ref{gl-results} we show the attack detection performance of \method and \emph{NAPGuard} on GL2 and GL3. We observe that the superior performance of \method asserts its effectiveness in detecting different types of unseen attacks. Notably, while \emph{NAPGuard} experiences a clear performance drop when going from GL2 to the more challenging GL3, \method's detection performance remains largely unaffected. Note that the false positives increase abruptly in Figure~\ref{gl-results} due to the scarcity of clean images in GL2 and GL3. 

\subsection{Results on universal adversarial perturbation (UAP) dataset for image classification}
\begin{figure*}[!t]
\centering
\subfloat[\footnotesize Single patch.]{\includegraphics[width = 0.16\textwidth]{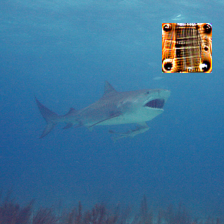}}\hfil
\subfloat[\footnotesize Double patch.]{\includegraphics[width = 0.16\textwidth]{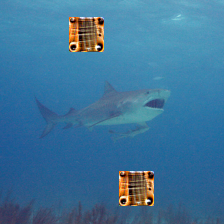}}\hfil
\subfloat[\footnotesize Quad patch.]{\includegraphics[width = 0.16\textwidth]{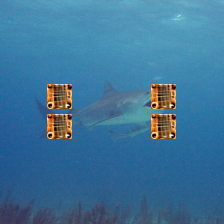}}
\caption{Single and multiple patches for image classification using the UAP attack~\cite{imgnt-patch}.} \label{patch-examples2}
\end{figure*}

Our results for the image classification task in Section~\ref{sec:numres} involve an input-specific attack for image classification~\cite{patchguard++}. Since this attack changes for every image, the attacks used for evaluation are not seen by \method during training, however, for completeness, we now evaluate \method on the targeted universal adversarial perturbation (UAP) attack~\cite{imgnt-patch}. In particular we use the \emph{Electric Guitar} patch and generate single and multiple patch attacks using the same attack model described for image classification in Section~\ref{sec:setup}. Note that for multiple patches we rescale the patch and apply it in separate regions, as illustrated in Figure~\ref{patch-examples2}. We use the default size of $50\times 50$ pixels for this attack. Moreover, to decouple the effect of training \method on an image-specific attack during training, we train \method using the TSEA-YOLOv3~\cite{tsea} patch instead, which is one of the patches used by \emph{NAPGuard} during training~\cite{napguard}. As before, for training we use only single-patch attacks with a fixed size of $32\times 32$ pixels; we also present results for our OCC variant introduced in Section~\ref{sec:numres-occ}, which does not use patch attacks during training.

\begin{figure}[h]
\centering
\includegraphics[width=0.34\textwidth]{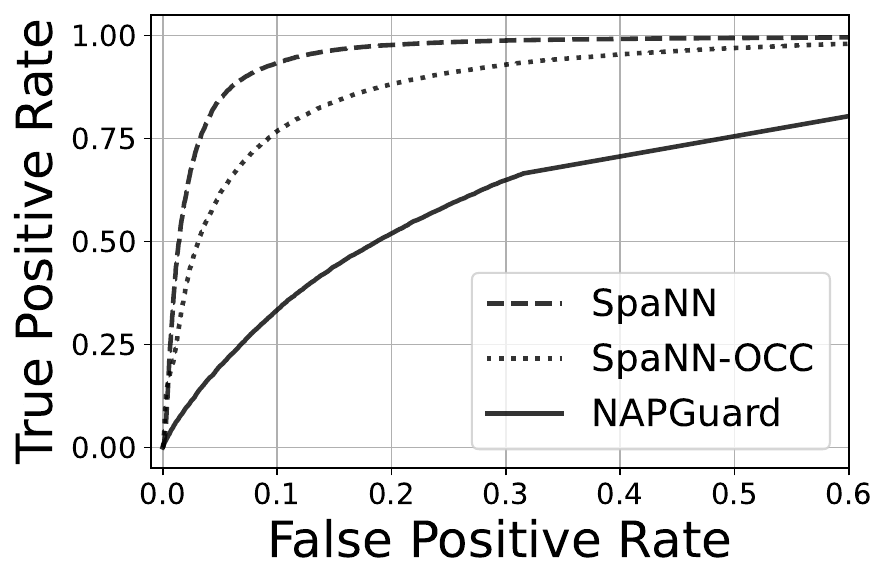}
\includegraphics[width=0.34\textwidth]{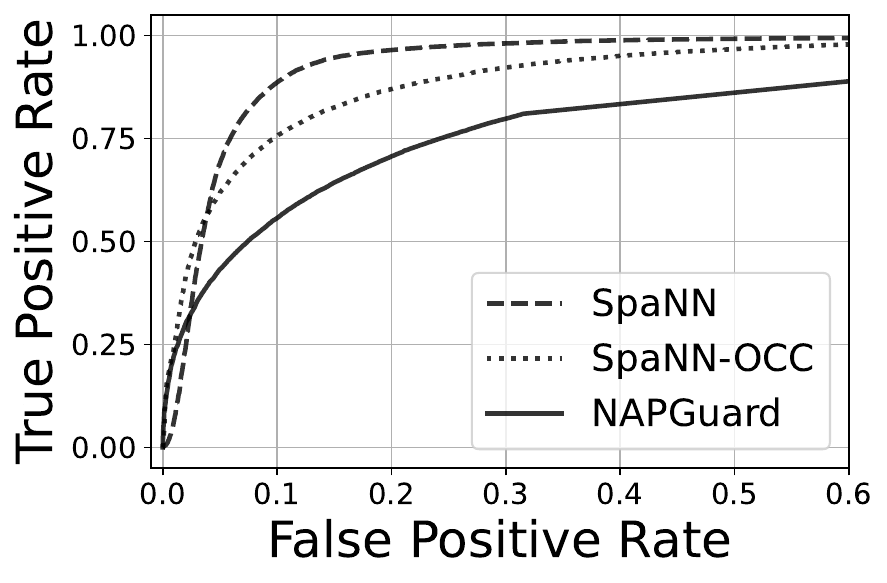}
\includegraphics[width=0.34\textwidth]{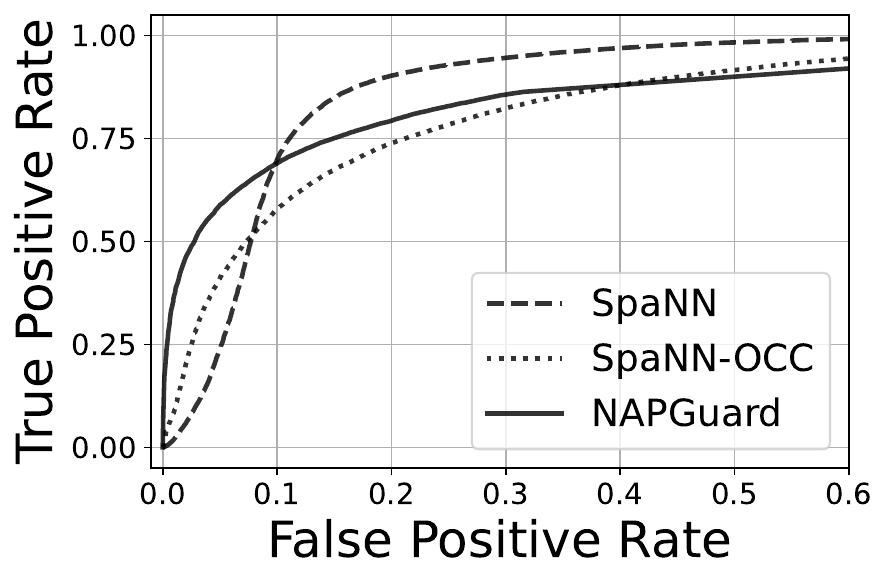}
\caption{Attack detection and false alarm rates for adversarial single (top), double (middle), and quadruple (bottom) patch detection for image classification (ImageNet), using the UAP \emph{Electric Guitar} patch~\cite{imgnt-patch}.}\label{roc-uap-in}
\end{figure}

\begin{figure}[h]
\centering
\includegraphics[width=0.34\textwidth]{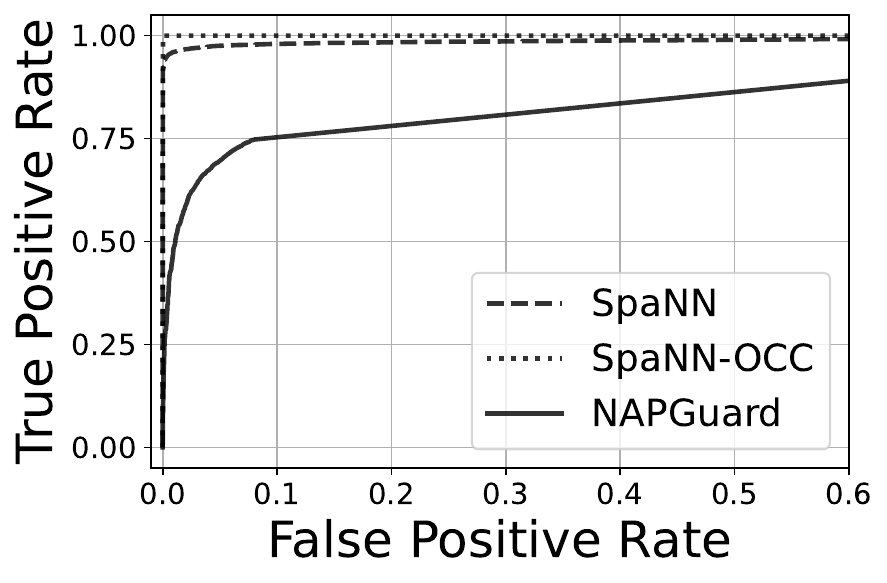}
\includegraphics[width=0.34\textwidth]{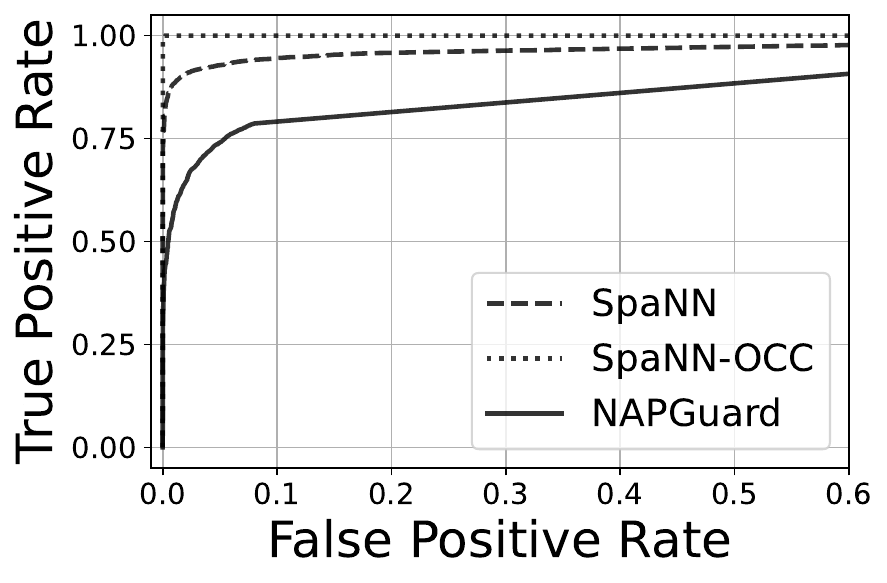}
\includegraphics[width=0.34\textwidth]{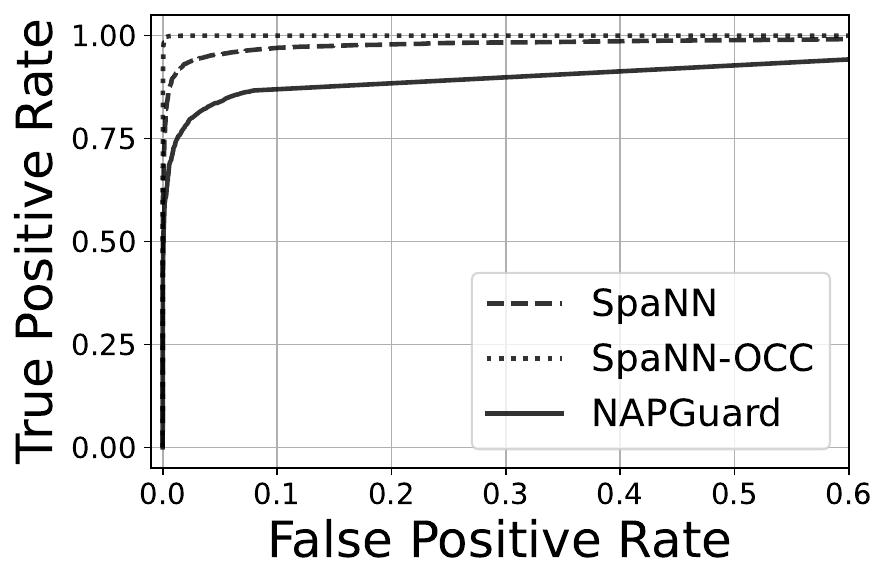}
\caption{Attack detection and false alarm rates for adversarial single (top), double (middle), and quadruple (bottom) patch detection for image classification (CIFAR-10), using the UAP \emph{Electric Guitar} patch~\cite{imgnt-patch}.}\label{roc-uap-cifar}
\end{figure}

Figures~\ref{roc-uap-in} and~\ref{roc-uap-cifar} show the attack detection performance of \method, its OCC variant (denoted \method-OCC), and \emph{NAPGuard} for ImageNet and CIFAR-10, respectively, using the UAP attack model. The figures show results over all attacks, regardless of effectiveness. The figures show that \method still enjoys a good detection performance for any number of patches with the UAP attack on image classification. In most cases, both \method and \method-OCC are able to outperform \emph{NAPGuard}, with the exception of quadruple patches on ImageNet, where \emph{NAPGuard} performs close to \method and outperforms \method-OCC.

\subsection{Effectiveness of Patch Attacks}
In Table~\ref{detperf-tab} we report attack detection accuracy for effective and ineffective attacks, while in Table~\ref{tab-occ} we report detection accuracy on all attacks, moreover, different attack models are used for object detection in both tables. To facilitate comparisons between the baselines and our OCC variant, and to provide further details on the attacks used in our evaluation, we show the attack effectiveness of the different attacks on our victim models in Table~\ref{tab-atkeff}, in other words, we report what fraction of the attacked versions of each dataset is considered effective. While enhancing attack effectiveness is outside of the scope of our work, we note that our multiple-patch attacks on image classification are more effective than their single patch counter parts (recall the attacked region is fixed regardless of the number of patches). The drop in effectiveness for double patches on object detection follows from the fact that we do not optimize these attacks and instead rescale, reshape, and translate attacks optimized under the single-patch scenario to generate our double patch attacks. While we do not optimize our multiple-patch versions of the UAP attack either, this attack was optimized using random locations, hence it is directly applicable for our double- and quadruple-patch attacks on image classification~\cite{imgnt-patch}.

\begin{table*}
\centering
\caption{Effectiveness of different patch attacks}\label{tab-atkeff}
\begin{tabular}{lcccccccc}
\toprule
\multicolumn{1}{c}{\multirow{2}{*}{\textbf{Num. Patches}}} & \multicolumn{8}{c}{\textbf{Dataset}} \\ \cline{2-9} 
\multicolumn{1}{c}{} & \multicolumn{2}{c}{INRIA} & \multicolumn{2}{c}{VOC} & \multicolumn{2}{c}{ImageNet} & \multicolumn{2}{c}{CIFAR-10} \\
\multicolumn{1}{c}{} & Thys et al.~\cite{foolAutoSurveillance} & DM-NAP~\cite{diffpatch} & Thys et al.~\cite{foolAutoSurveillance} & DM-NAP~\cite{diffpatch}  & \multicolumn{1}{c}{PG++~\cite{patchguard++}} & \multicolumn{1}{c}{UAP~\cite{imgnt-patch}} & \multicolumn{1}{c}{PG++~\cite{patchguard++}} & \multicolumn{1}{c}{UAP~\cite{imgnt-patch}} \\
\hline
Single & 0.8090 & 0.2292 & 0.7364 & 0.2852 & \multicolumn{1}{c}{0.8959} & \multicolumn{1}{c}{0.2153} & \multicolumn{1}{c}{0.5022} & \multicolumn{1}{c}{0.0880}\\
Double & 0.5139 & 0.2153 & 0.5908 & 0.2601 & \multicolumn{1}{c}{0.9433} & \multicolumn{1}{c}{0.3354} & \multicolumn{1}{c}{0.6880} & \multicolumn{1}{c}{0.1459}\\
Quadruple & - & - & - & - & \multicolumn{1}{c}{0.9851} & \multicolumn{1}{c}{0.2924} & \multicolumn{1}{c}{0.7918} & \multicolumn{1}{c}{0.1666}\\
\bottomrule
\end{tabular}
\end{table*}

\subsection{Computational Cost Comparison}
\begin{figure*}[!t]
\vspace{-2mm}
\centering
\subfloat[\footnotesize INRIA.]{\includegraphics[width =0.22\textwidth]{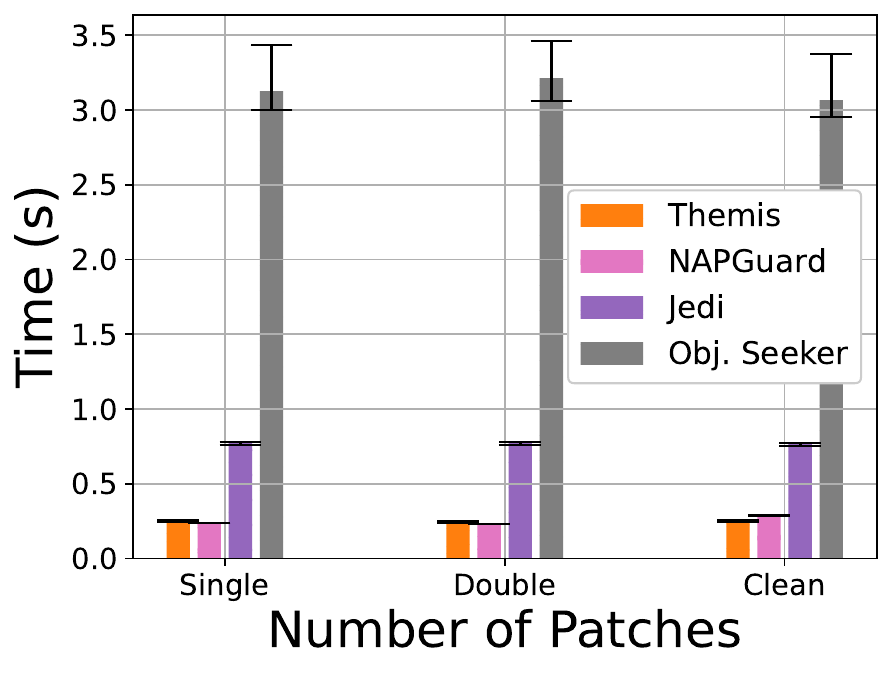}}\hfil
\subfloat[\footnotesize Pascal VOC.]{\includegraphics[width =0.22\textwidth]{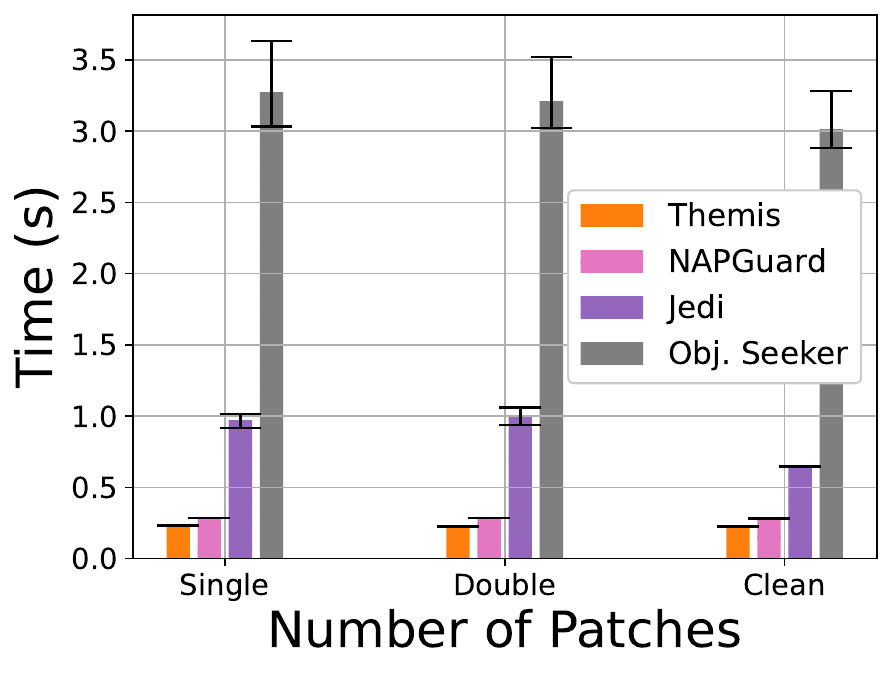}}\hfil
\subfloat[\footnotesize ImageNet.]{\includegraphics[width =0.22\textwidth]{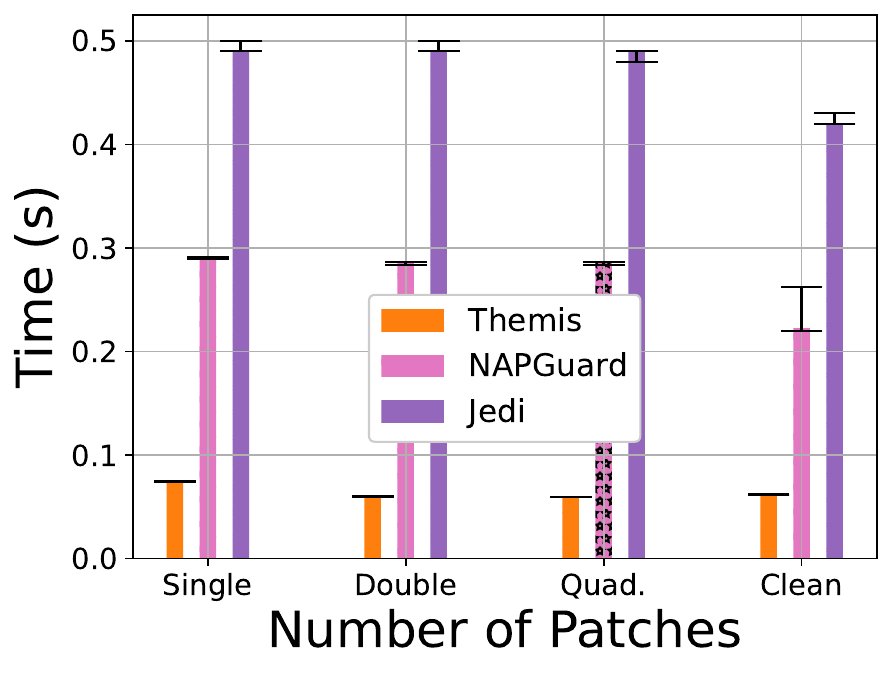}}\hfil
\subfloat[\footnotesize CIFAR-10.]{\includegraphics[width=0.22\textwidth]{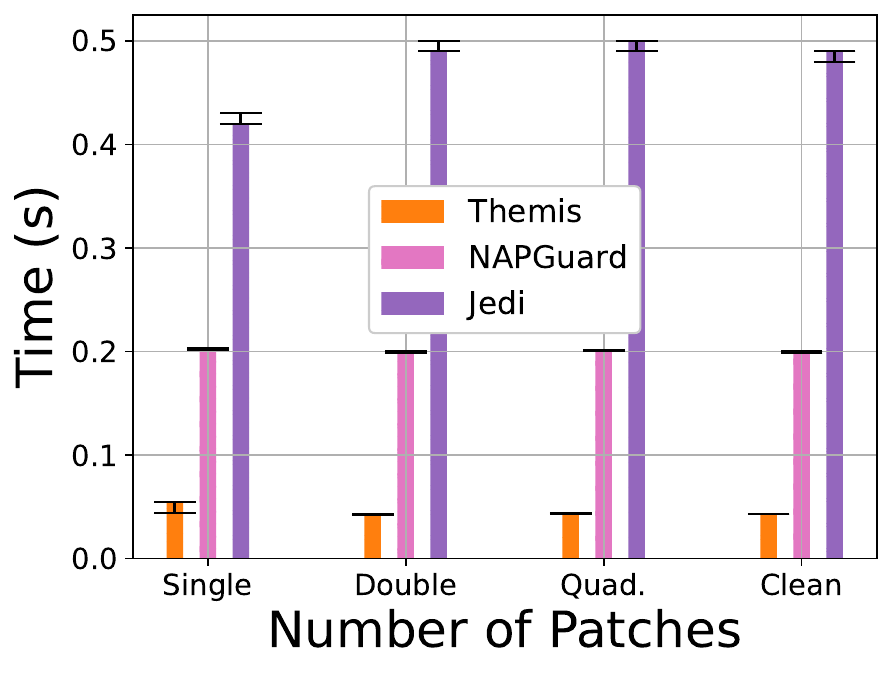}}
\caption{Computational cost of existing defenses against patch attacks. Error bars represent the first and third quartiles across each dataset.}\label{costcomp}
\end{figure*}

In Section~\ref{subsec:results} we showed how \method can tradeoff computational cost and detection accuracy. To provide further insight on how \method compares to existing approaches, we present the cost of \emph{NAPGuard}, \emph{Themis}, \emph{Jedi}, and \emph{Object Seeker} in Figure~\ref{costcomp}, using the same hardware we used for the execution times of \method reported in Figures~\ref{perf-complete} and~\ref{perf-complete2}. For all the methods in the figure, we report the time corresponding to their default parameters.

We note that \method can outperform certifiable defenses (i.e., \emph{Object Seeker}) in terms of computational cost even for a relatively large ensemble size $\vert\mathcal{B}\vert$ (c.f. Figures~\ref{perf-complete} and~\ref{perf-complete2}). Moreover, by reducing the ensemble size, \method can outperform \emph{Jedi} in terms of both detection accuracy and computational cost. At lower ensemble sizes ($\vert\mathcal{B}\vert \leq 10$), \method can even approximate the low cost of \emph{NAPGuard} and \emph{Themis} (which are computationally efficient by design~\cite{themis,napguard}) and still maintain a competitive detection accuracy. Moreover, we point out that unlike \method, existing computationally efficient methods such as \emph{Themis} and \emph{NAPGuard} lack the mechanisms to tradeoff a reduction in their speed for a higher detection accuracy.

\begin{figure*}[t]
\centering
\vspace{-2mm}
\subfloat[\footnotesize Number of important neurons.]{\includegraphics[width=0.19\linewidth, height=0.2\linewidth]{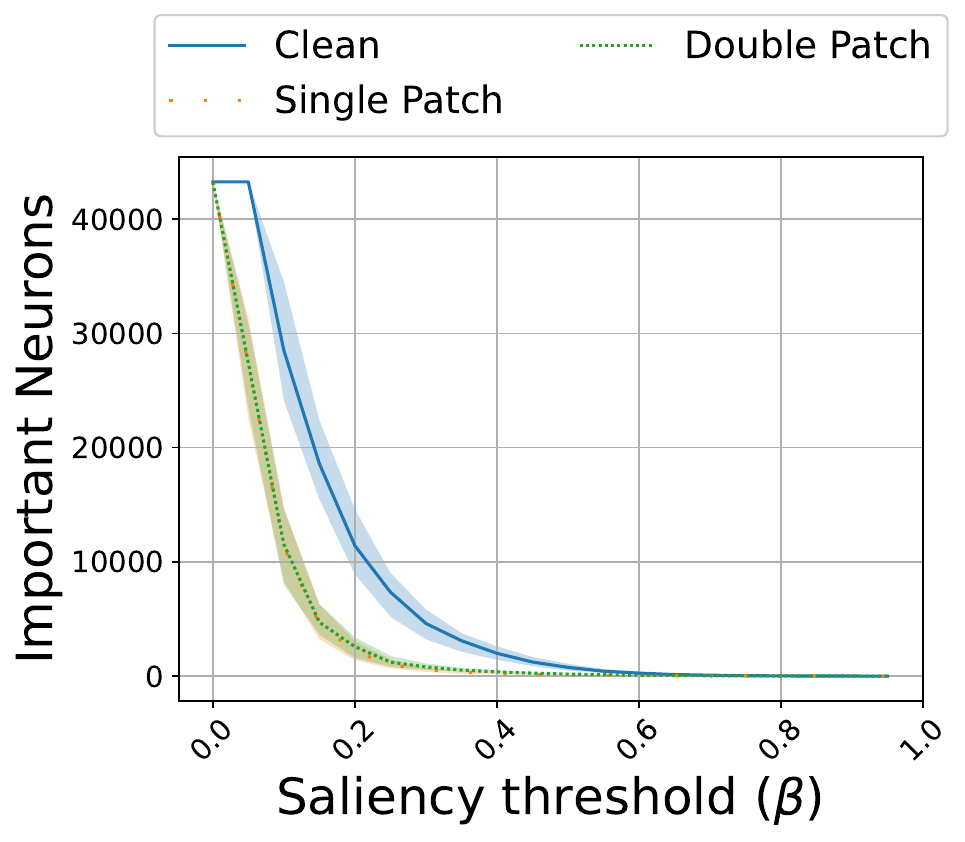}%
}
\hfil
\subfloat[\footnotesize Number of high entropy regions.]{\includegraphics[width=0.19\linewidth, height=0.2\linewidth]{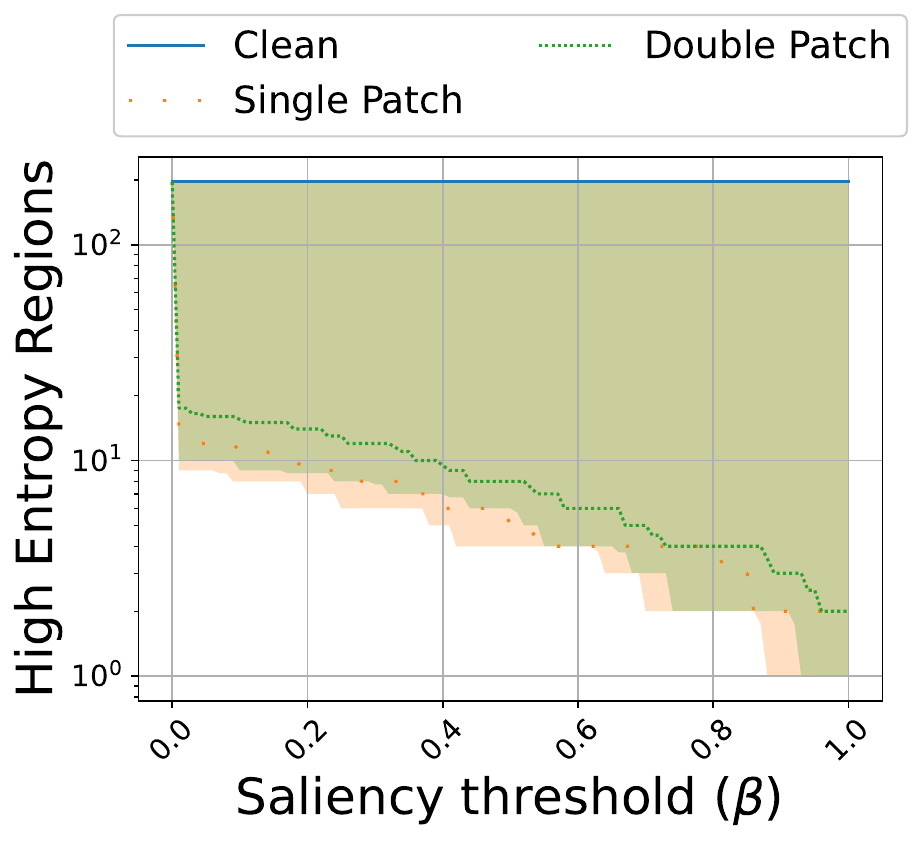}%
}
\hfil
\subfloat[\centering \footnotesize Number of clusters.]{\includegraphics[width=0.19\linewidth, height=0.2\linewidth]{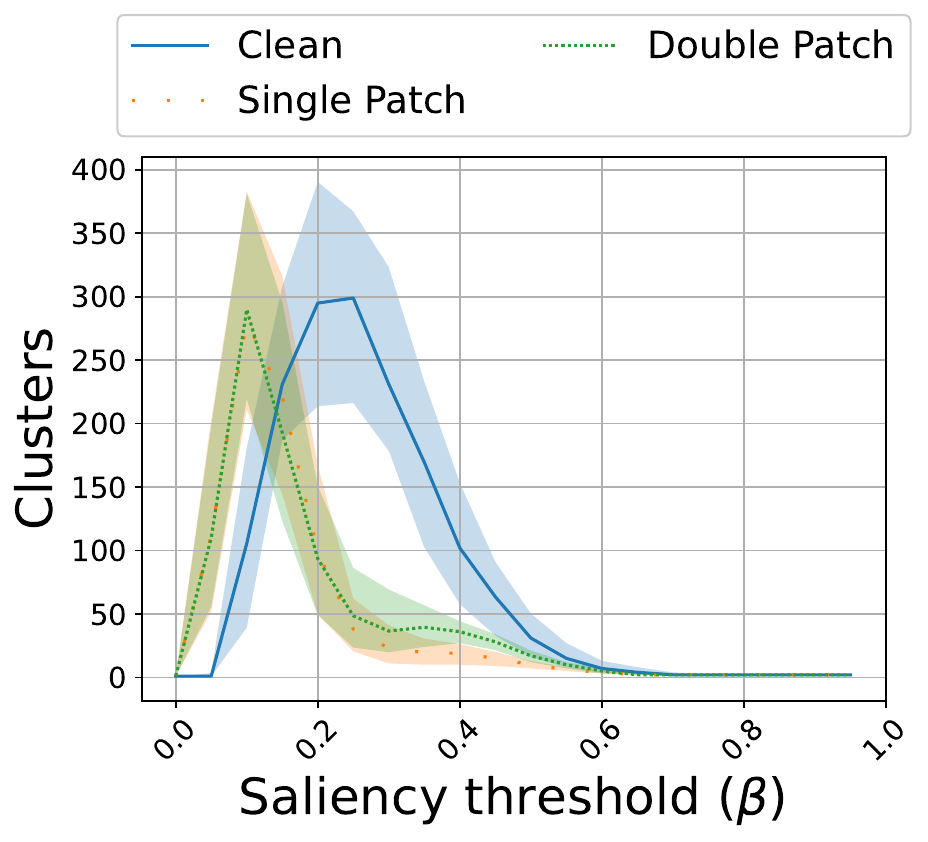}%
}
\hfil
\subfloat[\footnotesize Average mean intra-cluster distance.]{\includegraphics[width=0.19\linewidth, height=0.2\linewidth]{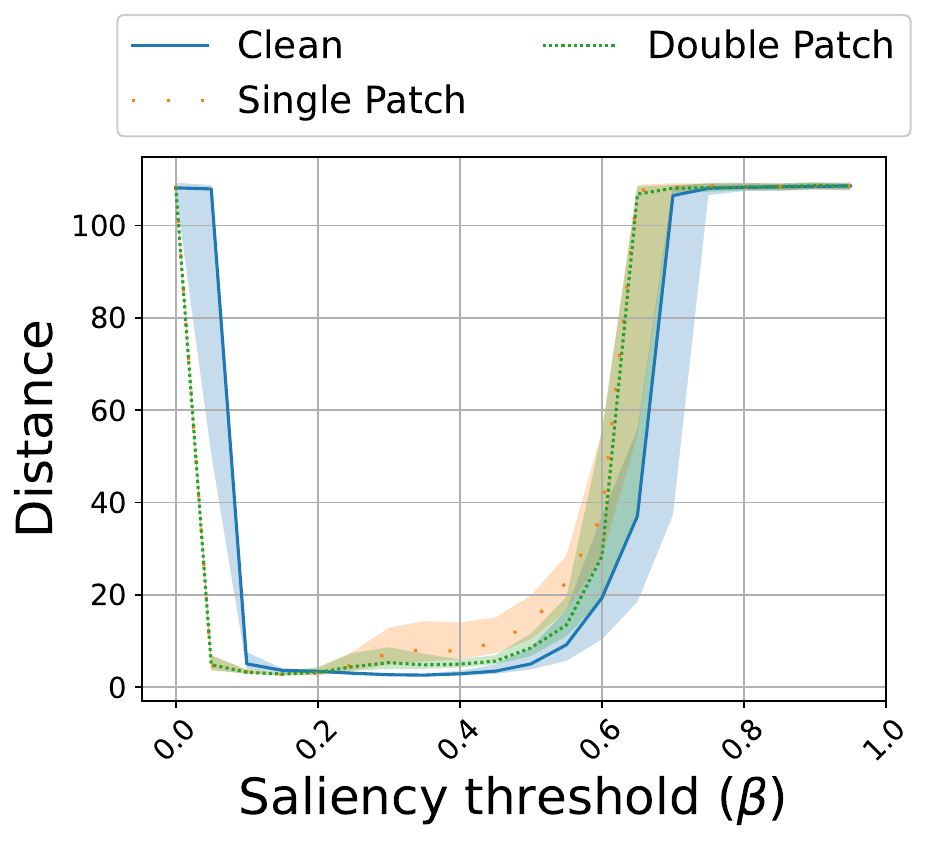}%
}
\hfil
\subfloat[\footnotesize Standard deviation of mean intra-cluster distance.]{\includegraphics[width=0.19\linewidth, height=0.2\linewidth]{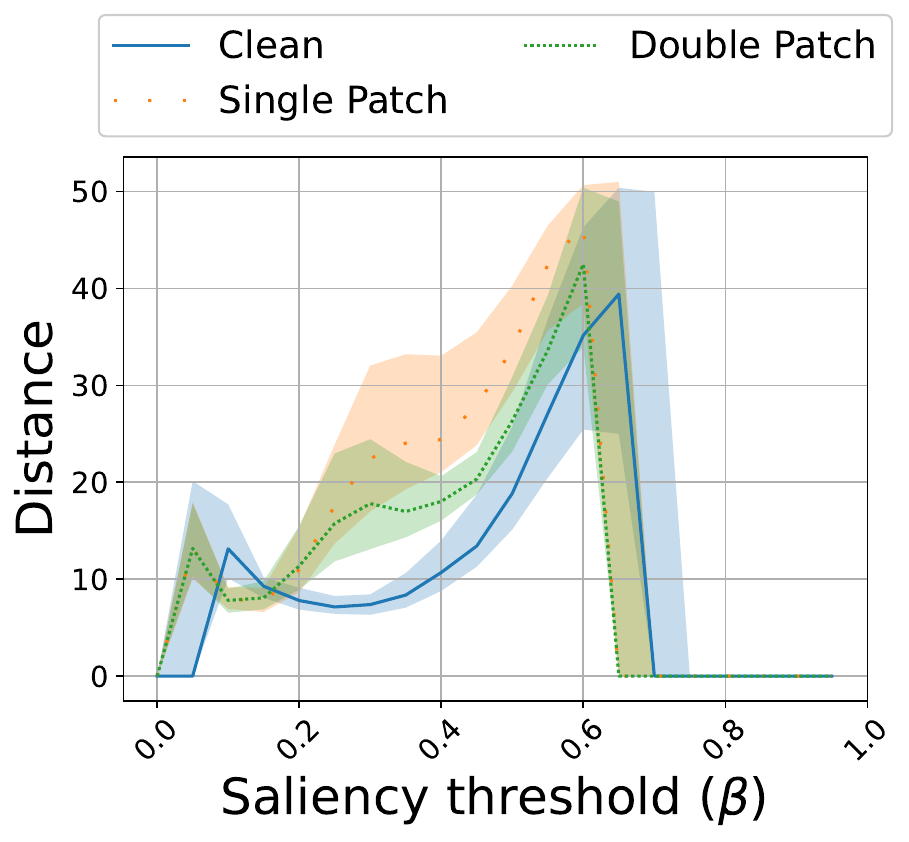}%
}
\caption{Input characteristics vs. saliency threshold $\beta$ (INRIA). Lines represent the median for each quantity, and shaded regions show the first and third quartiles.}
\label{preliminary-inria}
\end{figure*}

\begin{figure*}[t]
\centering
\subfloat[\footnotesize Number of important neurons.]{\includegraphics[width=0.19\linewidth, height=0.2\linewidth]{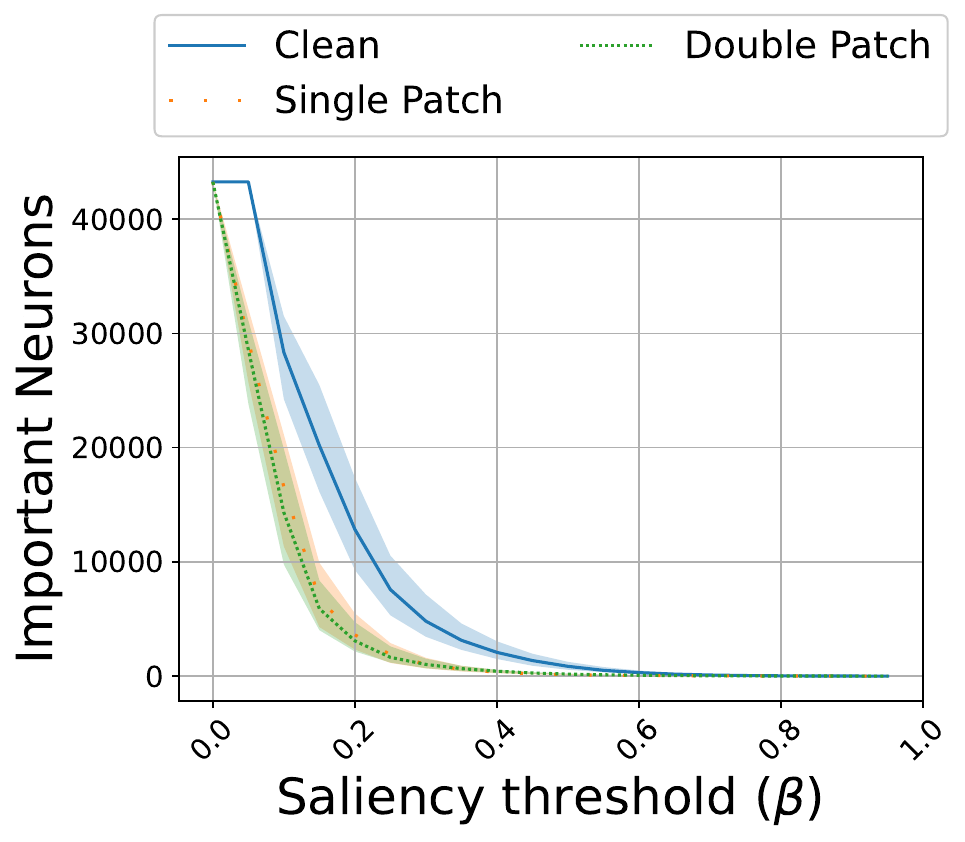}%
}
\hfil
\subfloat[\footnotesize Number of high entropy regions.]{\includegraphics[width=0.19\linewidth, height=0.2\linewidth]{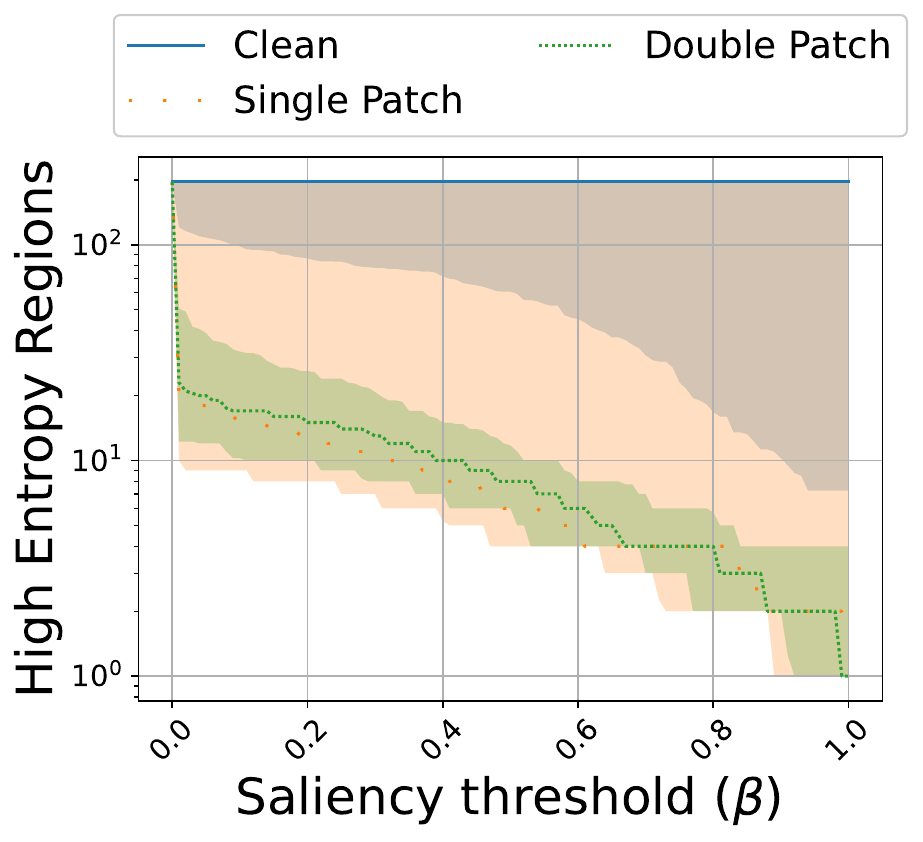}%
}
\hfil
\subfloat[\centering \footnotesize Number of clusters.]{\includegraphics[width=0.19\linewidth, height=0.2\linewidth]{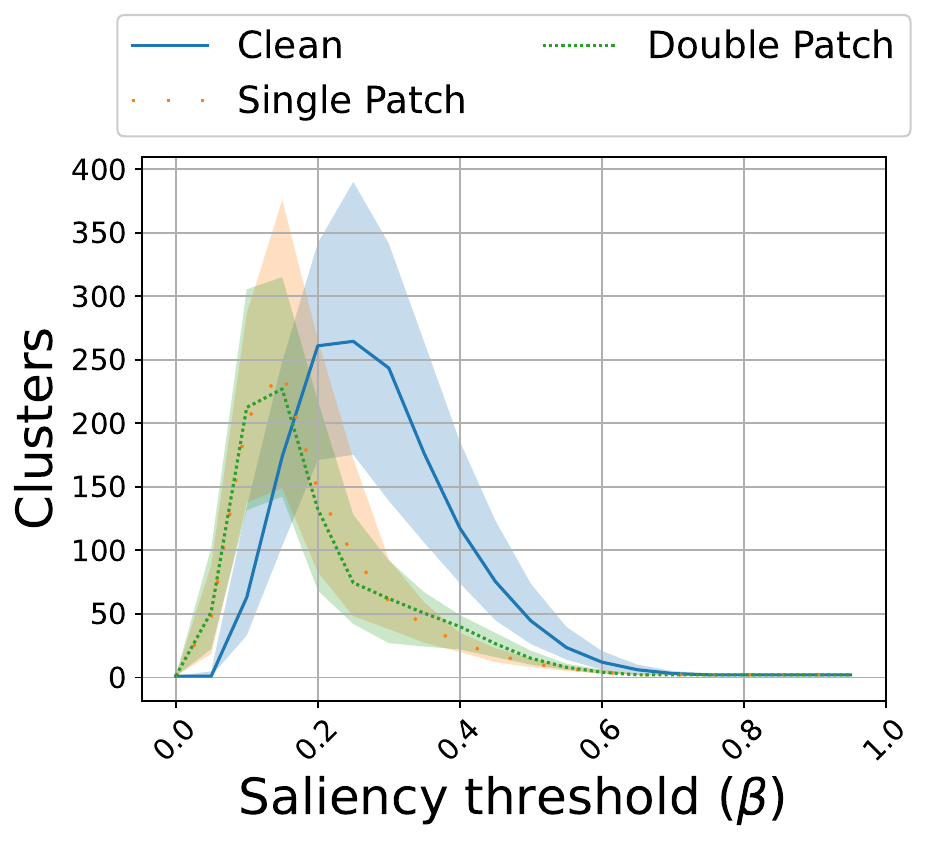}%
}
\hfil
\subfloat[\footnotesize Average mean intra-cluster distance.]{\includegraphics[width=0.19\linewidth, height=0.2\linewidth]{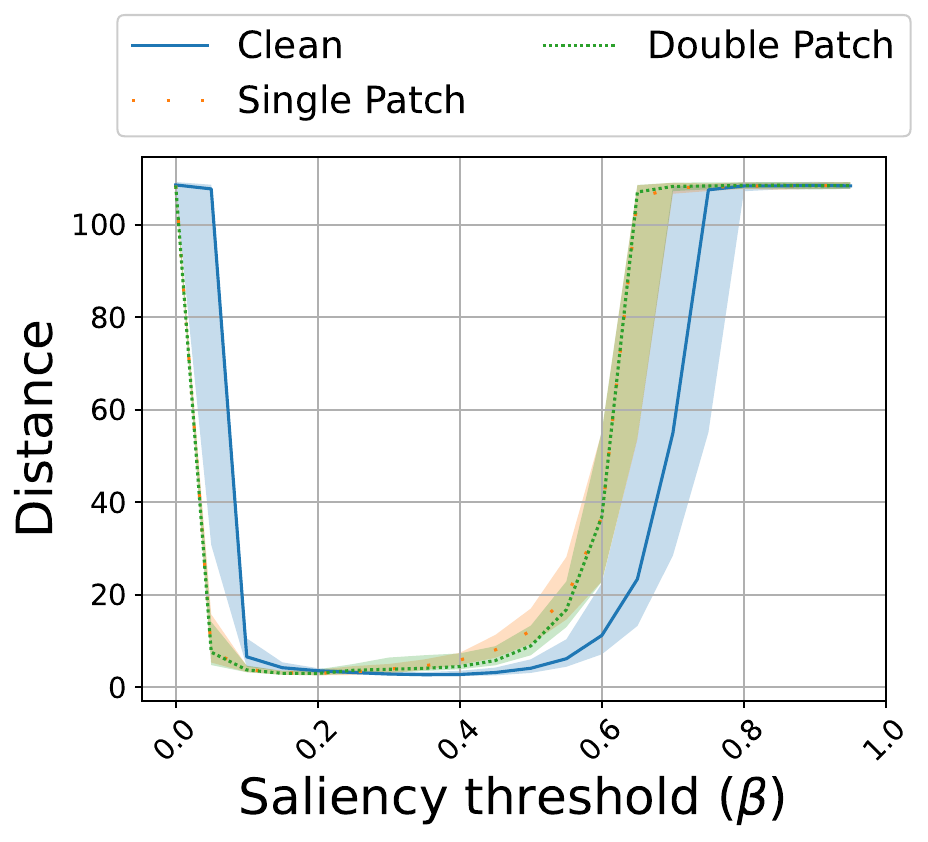}%
}
\hfil
\subfloat[\footnotesize Standard deviation of mean intra-cluster distance.]{\includegraphics[width=0.19\linewidth, height=0.2\linewidth]{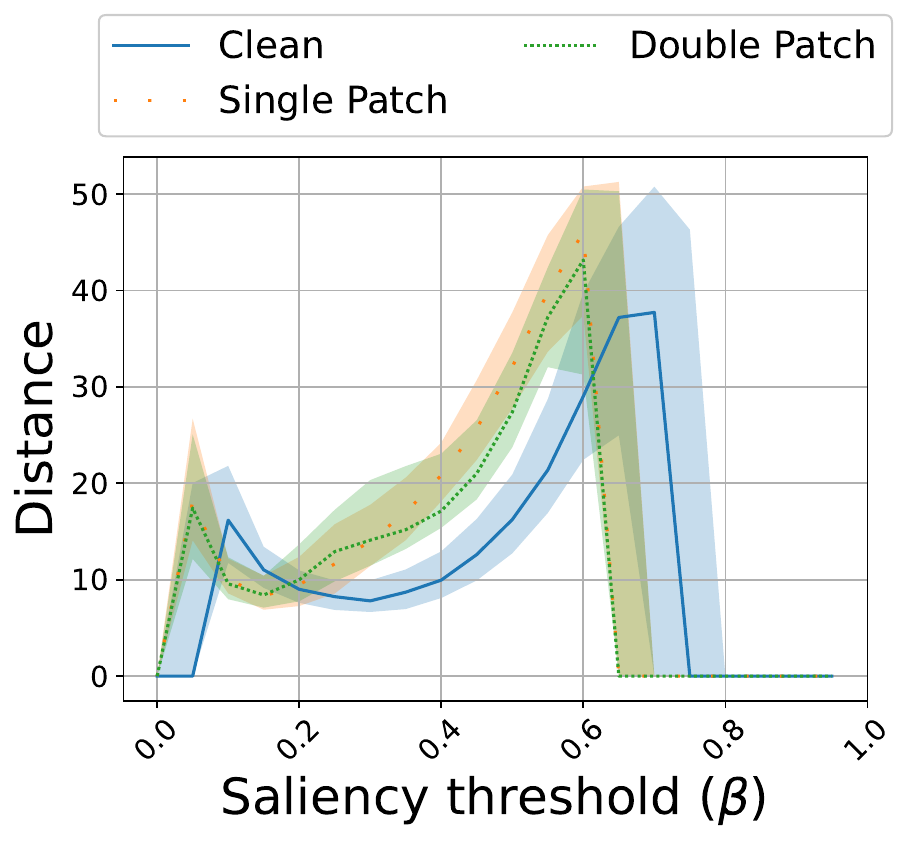}%
}
\caption{Input characteristics vs. saliency threshold $\beta$ (Pascal VOC). Lines represent the median for each quantity, and shaded regions show the first and third quartiles.}
\label{preliminary-voc}
\end{figure*}

\begin{figure*}[t]
\centering
\subfloat[\footnotesize Number of important neurons.]{\includegraphics[width=0.19\linewidth, height=0.2\linewidth]{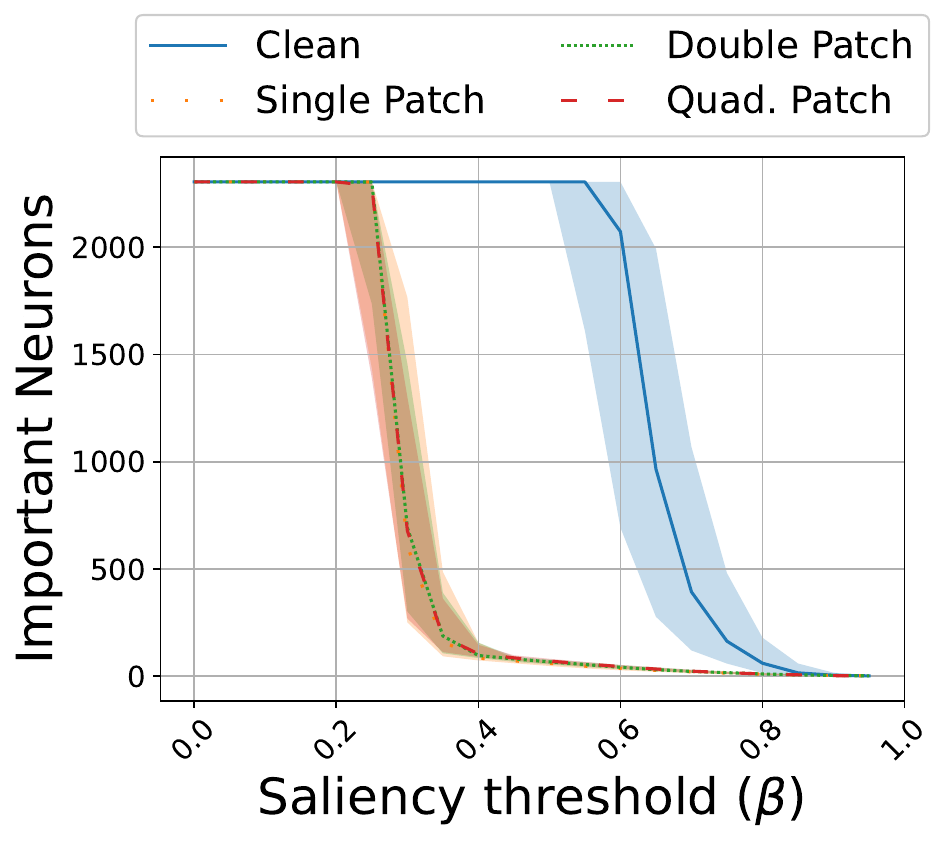}%
}
\hfil
\subfloat[\footnotesize Number of high entropy regions.]{\includegraphics[width=0.19\linewidth, height=0.2\linewidth]{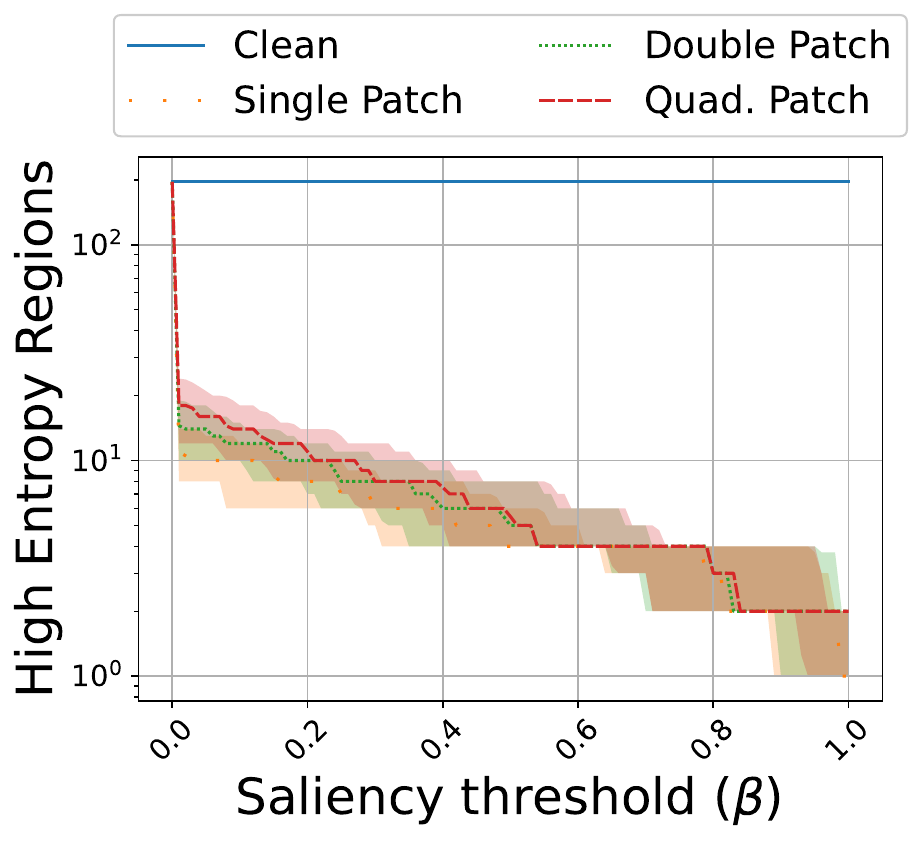}%
}
\hfil
\subfloat[\centering \footnotesize Number of clusters.]{\includegraphics[width=0.19\linewidth, height=0.2\linewidth]{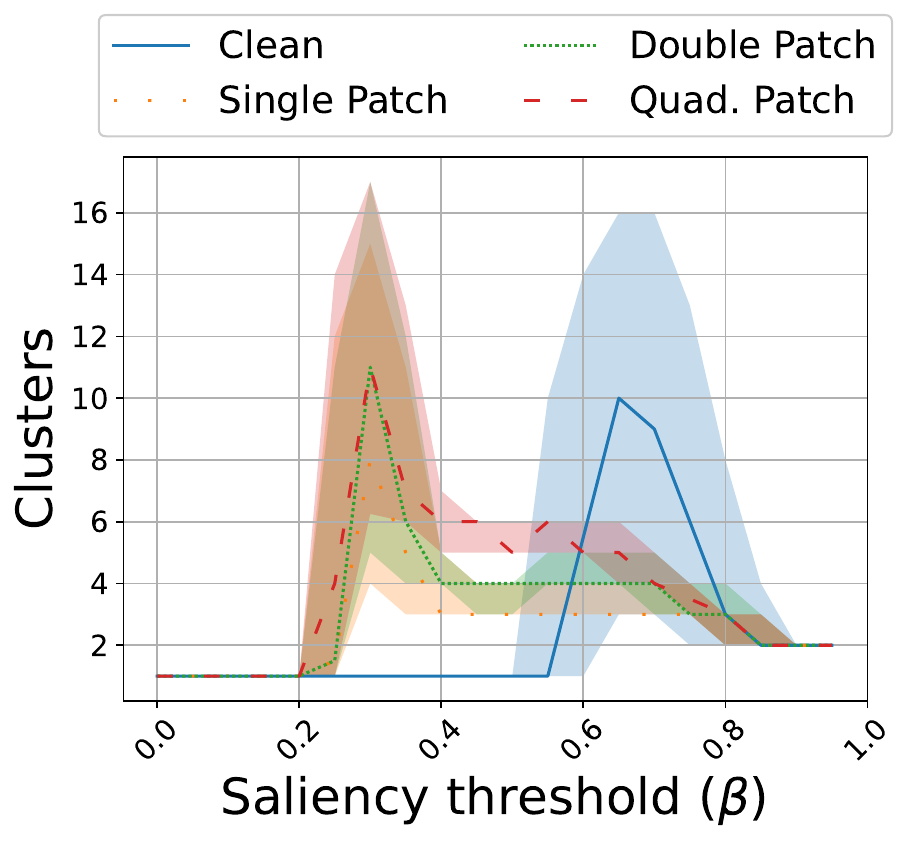}%
}
\hfil
\subfloat[\footnotesize Average mean intra-cluster distance.]{\includegraphics[width=0.19\linewidth, height=0.2\linewidth]{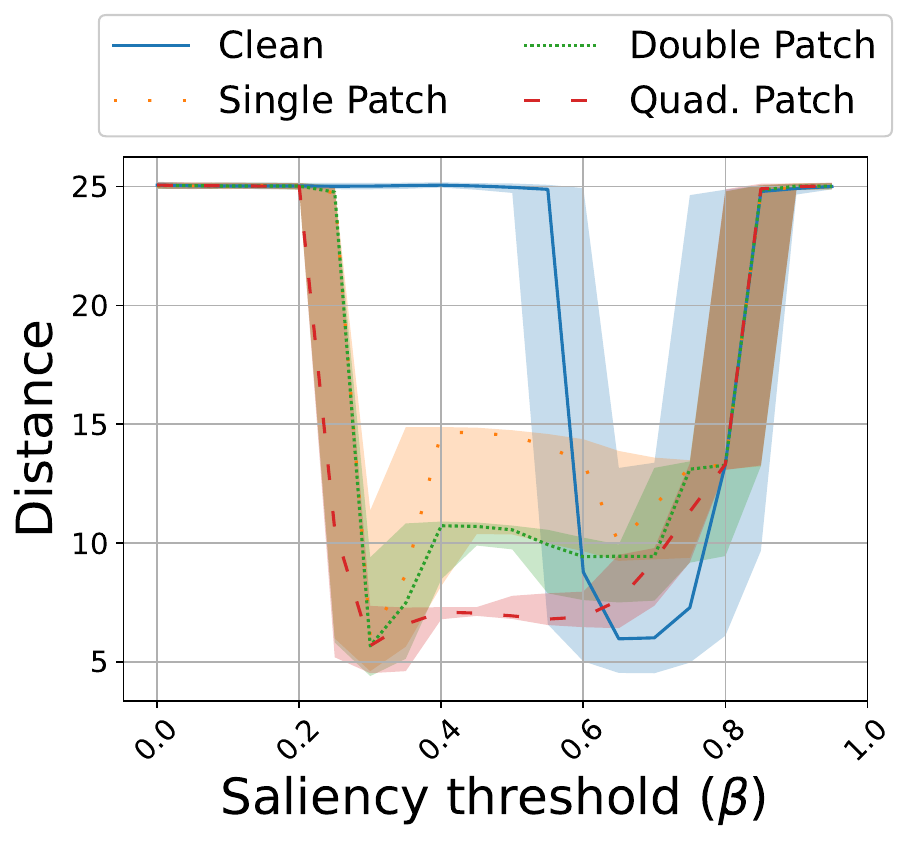}%
}
\hfil
\subfloat[\footnotesize Standard deviation of mean intra-cluster distance.]{\includegraphics[width=0.19\linewidth, height=0.2\linewidth]{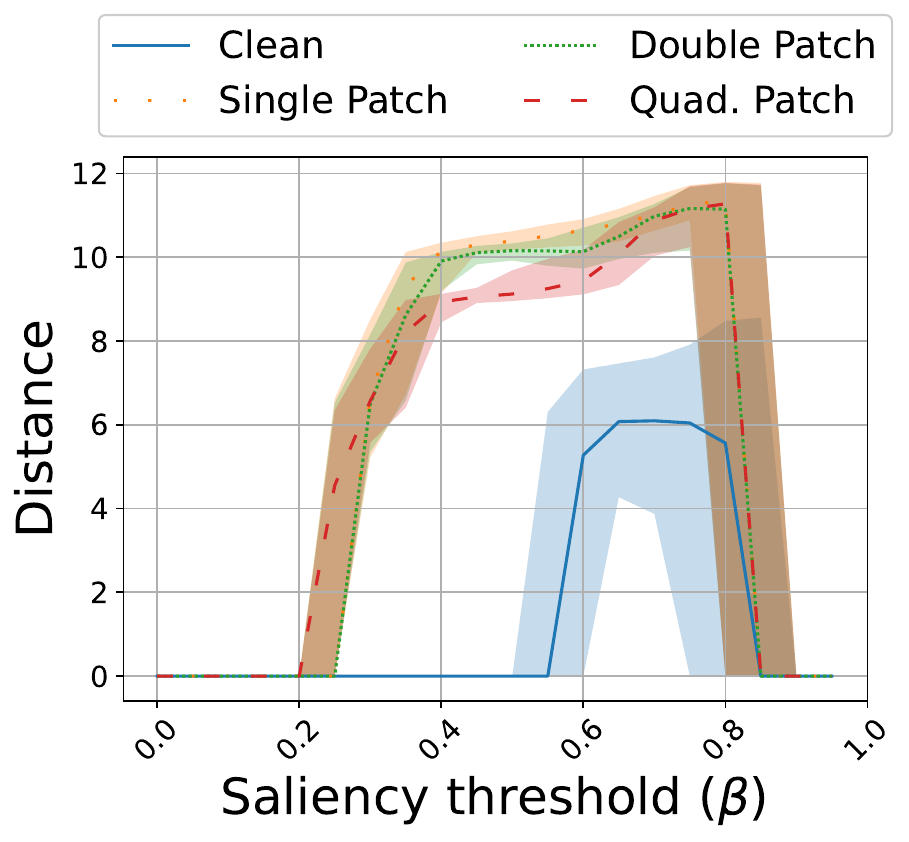}%
}
\caption{Input characteristics vs. saliency threshold $\beta$ (CIFAR-10). Lines represent the median for each quantity, and shaded regions show the first and third quartiles.}
\label{preliminary-cifar}
\end{figure*}

\subsection{Clustering Features Across Attack Models and Datasets}\label{sec:clust-feats}
In the main body of the paper, we show the dependency of attack detection results on the choice of the saliency threshold, and then motivate \method by showing how the change of our proposed clustering features across saliency thresholds can be used to discriminate between clean and attacked images. In particular, Figure~\ref{preliminary} shows this for a random subset of images from the ImageNet dataset. For completeness, we now present similar figures for all datasets, to confirm that (i) the dependence on the saliency threshold, and (ii) the ability to detect patch attacks from the curves generated by our proposed features across a set of thresholds, are not specific to a particular dataset or a particular attack model. We present clustering features as a function of the saliency threshold $\beta$ for random subsets from the INRIA, Pascal VOC, and CIFAR-10 datasets in Figures~\ref{preliminary-inria}, \ref{preliminary-voc}, and \ref{preliminary-cifar}, respectively. The subsets for Pascal VOC and CIFAR-10 contain 250 images each, and the subset for INRIA contains 100 images. These figures allow us to make the following observations:
\begin{itemize}
    \item In all datasets, the ability to discriminate between clean images and attacked images, based on the number of important neurons or high-entropy regions, depends on the choice of the saliency threshold.
    \item For all datasets, the shape of the curves generated by our proposed features (i.e., the number of important neurons, the number of clusters, the average intra-cluster distance, and the standard deviation of the latter) across different thresholds can be used to discriminate clean images from patched images with any number of patches. This shows they can be used for accurate attack detection across patch attack models, victim models, and tasks, as confirmed by our numerical results.
    \item The differences between clean and patched images in terms of our proposed clustering features show a consistent pattern across datasets, although the precise shapes of the curves differ across datasets.
\end{itemize}

The first observation supports the concerns we raise regarding the use of a single threshold for attack detection, as they are not constrained to a particular task or attack model. The second observation gives an intuition as to why \method is able to perform well on different datasets, regardless of the attack model, and explains why the curves fed into the attack detector network \emph{AD} enable detecting attacks with any number of patches. Our approach works under the assumption that \emph{AD} can be trained for each context, and the requirement to train \emph{AD} for each dataset is explained by our third observation: even though similar patterns are observed for all datasets, the precise patterns corresponding to clean and attacked images change slightly between datasets. Note, for example, how the number of neurons tends to drop for smaller values of $\beta$ in attacked images for all datasets, but comparing Figure~\ref{preliminary-inria}(a) with Figure~\ref{preliminary-cifar}(a), it is evident that the curve for clean images in Figure~\ref{preliminary-inria}(a) is further away from to the curve for clean images in Figure~\ref{preliminary-cifar}(a) than it is to the curves for attacked images in both figures. We also note that the victim model is of particular importance: the similarity between Figures~\ref{preliminary-inria} and~\ref{preliminary-voc} indicates that using the same victim model on a different dataset has only a slight impact on the curves for clean and attacked images. Note that for image classification, the weights, input, and output layers of the victim model depend on the particular dataset used, hence the differences between Figures~\ref{preliminary} and~\ref{preliminary-cifar} are more noticeable. Finally, we observe that while the curves corresponding to object detection are notably distinct from those corresponding to image classification, there are certain consistencies across contexts, such as the number of important neurons decreasing at a larger saliency threshold for clean images or the number of clusters peaking at a higher saliency threshold for clean images.

The retraining requirement to adjust \method to a particular context is its main limitation, however, we argue that \method's training overhead is a significant improvement over prior methods, which demand intricate procedures to determine adequate context-dependent parameter settings \cite{jedi,z-mask,ObjSeek,themis}, whereas \method can learn automatically and efficiently to detect attacks in a given context. Moreover, our results regarding patches that are not seen during training and unsupervised detection (i.e., our OCC variant) further alleviate the impact of this drawback. From the high level similarities between the curves in Figures~\ref{preliminary}, \ref{preliminary-inria}, \ref{preliminary-voc}, and \ref{preliminary-cifar}, we conjecture that \method has the potential to generalize across contexts to some extent without retraining the attack detector architecture, even when the particular task, victim model, dataset, or attack model are not seen during training. We leave the exploration of such an approach to transferability between contexts to be the subject of future work.

\subsection{Baseline Attack Detection Algorithms}
\begin{algorithm}[b!]
\caption{\emph{Jedi}-detect:}\label{jedi-detect}
\begin{algorithmic}
\Require Model~$\model$, Auto-Encoder~$AE$, Auto-Encoder output threshold $t_{AE}$, input data~$\mathcal{X}$, entropy statistics for clean images~$E_{clean}$
\For{$\singlein \in \mathcal{X}$}
\State $E \gets \text{EntropyHeatMap}(\singlein)$
\Comment{$E \in \mathbb{R}^{H\times W}$}
\State $t := \text{ComputeThreshold}(E, E_{clean})$
\State ${E} := ({E \geq t})\odot E$\Comment{$E_{ij}:=\mathbbm{1}({E_{ij} \geq t})\cdot E_{ij}$}\vspace{2mm}
\State $E \gets \text{PreProcessing}(E)$
\State $E \gets \text{AE}(E)$
\State ${E} := ({E \geq t_{AE}})\odot E$\Comment{${E}_{ij} := \mathbbm{1}({E_{ij} \geq t_{AE}})\cdot E_{ij}$}\vspace{2mm}
\State $\singlein^m \gets \text{MaskInpainting}(\singlein, E)$ \Comment{Tarchoun et al.~\cite{jedi}}
\State $\hat{\mathbf{y}} \gets \model(\singlein)$
\State $\hat{\mathbf{y}}_J \gets \model(\singlein^m)$
\If{$\hat{\mathbf{y}}_J \neq \hat{\mathbf{y}}$}
\State \Return Detected Attack. \Comment{$E$ covered a patch}
\EndIf
\State \Return $\hat{\mathbf{y}}$ \Comment{$\singlein$ is a clean image}
\EndFor
\end{algorithmic}
\end{algorithm}
\begin{algorithm}[!t]
\caption{\emph{Themis}-detect:}\label{themis-detect}
\begin{algorithmic}
\Require Model~$\model$, window threshold $\theta \in [0,1]$, importance threshold $\beta \in [0,1]$, window size $n_w\in\mathbb{Z}$, input data~$\mathcal{X}$
\For{$\singlein \in \mathcal{X}$}
\State $M, \hat{\mathbf{y}} \gets \model(\singlein)$ \Comment{$M \in \mathbb{R}^{m_x\times m_y}$ is a feature map}
\State $t := \beta\cdot\max(M)$ \Comment{Importance threshold}
\State$B := M\geq t$\Comment{${B}_{ij} = \mathbbm{1}({M_{ij} \geq t})$}\vspace{2mm}
\For{$W\in B$} \Comment{$W \in \mathbb{R}^{n_w\times n_w}$ is a window of $B$}
\vspace{1mm}
\If{$\sum_{a_{ij} \in W} a_{ij} \geq \theta\cdot n_w^2$}
\vspace{1mm}
\State $\singlein^m = \text{Mask}(\singlein, W)$ \Comment{$W$ may be a patch}
\State $\hat{\mathbf{y}}_W \gets M(\singlein^m)$
\If{$\hat{\mathbf{y}}_W \neq \hat{\mathbf{y}}$}
\State \Return Detected Attack. \Comment{$W$ \emph{is} a patch}
\EndIf
\EndIf
\EndFor
\State \Return $\hat{\mathbf{y}}$ \Comment{$\singlein$ is a clean image}
\EndFor
\end{algorithmic}
\end{algorithm}
\begin{algorithm}[t]
\caption{\emph{ObjectSeeker}-detect:}\label{objskr-detect}
\begin{algorithmic}
\Require Model~$\model$, number of horizontal lines $k_x$, number of vertical lines $k_y$, input data~$\mathcal{X}$
\For{$\singlein \in \mathcal{X}$}
\State $\hat{\mathbf{y}} \gets \model(\singlein)$ \Comment{Original inference}
\State $\alpha:=1$ \Comment{Initialize min. intersection over area (IoA)}
\For{$l_x \in \{1,...,k_x\}$} \Comment{Horizontal lines}
\State $\singlein^{a}, \singlein^{b} \gets \text{HorizontalSplit}(\singlein, l_x)$
\State $\hat{\mathbf{y}}^a
\gets M(\singlein^{a})$
\State $\hat{\mathbf{y}}^b \gets M(\singlein^{b})$
\State $\alpha_{1}\gets\underset{q}{\text{min}}\hspace{2mm}\underset{r}{\text{max}}\hspace{2mm}\text{IoA}(\hat{\mathbf{y}}^a_{q}, \hat{\mathbf{y}}_r)$
\State $\alpha_{2}\gets\underset{q}{\text{min}}\hspace{2mm}\underset{r}{\text{max}}\hspace{2mm}\text{IoA}(\hat{\mathbf{y}}^b_{q}, \hat{\mathbf{y}}_r)$
\State $\alpha \gets \text{min}\{\alpha, \alpha_{1}, \alpha_{2}\}$ 
\EndFor
\For{$l_y \in \{1,...,k_y\}$} \Comment{Vertical lines}
\State $\singlein^{a}, \singlein^{b} \gets \text{VerticalSplit}(\singlein, l_y)$
\State $\hat{\mathbf{y}}^a \gets M(\singlein^{a})$
\State $\hat{\mathbf{y}}^b \gets M(\singlein^{b})$
\State $\alpha_{1}\gets\underset{q}{\text{min}}\hspace{2mm}\underset{r}{\text{max}}\hspace{2mm}\text{IoA}(\hat{\mathbf{y}}^a_{q}, \hat{\mathbf{y}}_r)$
\State $\alpha_{2}\gets\underset{q}{\text{min}}\hspace{2mm}\underset{r}{\text{max}}\hspace{2mm}\text{IoA}(\hat{\mathbf{y}}^b_{q}, \hat{\mathbf{y}}_r)$
\State $\alpha \gets \text{min}\{\alpha, \alpha_{1}, \alpha_{2}\}$ 
\EndFor
\State \Return $1 - \alpha$ \Comment{Attack detection score}
\EndFor
\end{algorithmic}
\end{algorithm}
We described our baseline defenses in Section~\ref{sec:method}. For completeness, here we present the pseudocode for \emph{Jedi}-detect, \emph{Themis}-detect, and \emph{ObjectSeeker}-detect in Algorithms~\ref{jedi-detect}, \ref{themis-detect}, and~\ref{objskr-detect}, respectively. For clarity, we also discuss a few details of these algorithms. We exclude \emph{NAPGuard} from this section since we did no relevant modifications to its original formulation.

For \emph{Jedi}-detect and \emph{Themis}-detect (Algorithms~\ref{jedi-detect} and \ref{themis-detect}), we abuse notation and denote non-equivalent model inferences as $\hat{\mathbf{y}}^a \neq \hat{\mathbf{y}}^b$ for any task and attack model. That is, for image classification this means that $\hat{\mathbf{y}}^a$ and $\hat{\mathbf{y}}^b$ are different labels, while for object detection it means that $\hat{\mathbf{y}}^a$ and $\hat{\mathbf{y}}^b$ are sets of bounding boxes such that some object detected in $\hat{\mathbf{y}}^b$ has an intersection over union (IoU) below 50\% for every object in $\hat{\mathbf{y}}^a$.

For \emph{ObjectSeeker-detect} (Algorithm~\ref{objskr-detect}), $\alpha_1$ and $\alpha_2$ denote variables used to update $\alpha$, the least maximum intersection over area (IoA) for all objects detected in masked images  for a given input $\singlein$. $\alpha_1$ and $\alpha_2$ are obtained as follows: (i) the victim model is used to detect objects in a masked image; (ii) for each detected object, the IoA between that object and each object detected in the original non-masked image is calculated; the maximum IoA of each object in the masked image across the objects in the original is computed, and the minimum of these maximum IoAs across objects then yields $\alpha_1$ and $\alpha_2$ (each associated with masking one half of the original image). Then $\alpha$ is updated using the minimum between itself, $\alpha_1$, and $\alpha_2$. Thus $\alpha$ represents the lowest overlap that an object in a masked image has with the original objects. Note that in Algorithm~\ref{objskr-detect}, $q$ and $r$ are used to index each bounding box (object) present in the model's output. Since a lower overlap is indicative of a patch attack being suppressed, the detection score is computed as $1-\alpha$.

In addition to our description of baseline parameters in Section~\ref{subsec:results}, it is important to point out that the official implementation of $\emph{Jedi}$ offers two different autoencoder models, one for the ImageNet and Pascal VOC datasets, and one for the CASIA dataset. Since the distinction between the two is not based on the datasets, but on the patch attacks used on each dataset, we follow the original work and use the ImageNet/Pascal VOC autoencoder for attacks on image classification (i.e., CIFAR-10 and ImageNet) and the CASIA autoencoder for attacks on object detection (INRIA and Pascal VOC)~\cite{jedi}.

\subsection*{References}
All our citations in the appendix are made with respect to the references of the main paper.

\end{document}